\def\year{2021}\relax
\documentclass[letterpaper]{article} % DO NOT CHANGE THIS
\usepackage{aaai21}  % DO NOT CHANGE THIS
\usepackage{times}  % DO NOT CHANGE THIS
\usepackage{helvet} % DO NOT CHANGE THIS
\usepackage{courier}  % DO NOT CHANGE THIS
\usepackage[hyphens]{url}  % DO NOT CHANGE THIS
\usepackage{graphicx} % DO NOT CHANGE THIS
\usepackage{natbib}  % DO NOT CHANGE THIS AND DO NOT ADD ANY OPTIONS TO IT
\usepackage{caption} % DO NOT CHANGE THIS AND DO NOT ADD ANY OPTIONS TO IT
\usepackage{graphicx,amsmath,amsfonts,amssymb,cite,bm,url,breakurl,epsfig,epsf,mathbbol,fmtcount,semtrans,caption,subcaption,multirow,comment}
\usepackage[noend]{algpseudocode}
\usepackage{thm-restate}

\usepackage{amssymb}

\usepackage[utf8]{inputenc} % allow utf-8 input
\usepackage[T1]{fontenc}    % use 8-bit T1 fonts
\usepackage{url}            % simple URL typesetting
\usepackage{booktabs}       % professional-quality tables
\usepackage{amsfonts}       % blackboard math symbols
\usepackage{nicefrac}       % compact symbols for 1/2, etc.

\makeatother

  \makeatletter
\def\BState{\State\hskip-\ALG@thistlm}
\makeatother

  \usepackage{mathtools}

\usepackage{tikz}
\usepackage{pgfplots}
\usetikzlibrary{pgfplots.groupplots}

\usepackage{cite}
\usepackage{caption}
\usepackage[bottom,hang,flushmargin]{footmisc} 

\newcommand{\tsn}[1]{{\left\vert\kern-0.25ex\left\vert\kern-0.25ex\left\vert #1 
    \right\vert\kern-0.25ex\right\vert\kern-0.25ex\right\vert}}
\newtheorem{theorem}{Theorem}

\newtheorem{assumption}{Assumption}

\newtheorem{lemma}{Lemma}

\newtheorem{definition}{Definition}

\newcommand{\som}[1]{}%\marginpar{\tiny\ttfamily {\color{red}{SO: #1}}}}

\newcommand{\so}[1]{{\color{black}{#1}}}
\newcommand{\fx}[1]{{\color{black}{#1}}}

\newcommand{\eps}{\varepsilon}
\newcommand{\Fl}{{\cal{F}}_\Lc}
\newcommand{\Plt}{{\cal{F}}}
\newcommand{\Zc}{{\cal{Z}}}
\newcommand{\fl}{f_\Lc}
\newcommand{\vs}{\vspace{-2pt}}
\newcommand{\vp}{\vspace{2pt}}

\newcommand{\distas}{\overset{\text{i.i.d.}}{\sim}}

\newcommand{\pbar}{{\bar{p}}}
\newcommand{\sbar}{{\bar{s}}}

\newcommand{\beq}{\begin{equation}}

\newcommand{\eeq}{\end{equation}}

\newcommand{\nn}{\nonumber}
\newcommand{\la}{\lambda}
\newcommand{\bla}{\boldsymbol{\lambda}}
\newcommand{\blad}{\boldsymbol{\lambda}_\Delta}

\newcommand{\Ub}{{\mtx{U}}}
\newcommand{\cmt}[1]{}

\newcommand{\diag}[1]{\text{diag}(#1)}

\newcommand{\Lc}{{\cal{L}}}
\newcommand{\Lch}{{\hat{\cal{L}}}}

\newcommand{\bSi}{{\boldsymbol{{\Sigma}}}}

\newcommand{\bmu}{{\boldsymbol{{\mu}}}}

\newcommand{\onebb}{{\mathbf{1}}}
\newcommand{\Iden}{{\mtx{I}}}

\newcommand{\z}{{\vct{z}}}

\newcommand{\tn}[1]{\|{#1}\|_{\ell_2}}

\newcommand{\Dc}{\mathcal{D}}

\newcommand{\Rc}{\mathcal{R}}

\newcommand{\tb}{\boldsymbol{\theta}}
\newcommand{\tbh}{\boldsymbol{\hat{\theta}}}

\newcommand{\bt}{{\boldsymbol{\beta}}}

\newcommand{\bts}{\boldsymbol{\beta}^\st}
\newcommand{\bth}{{\boldsymbol{\hat{\beta}}}}
\newcommand{\btha}{{\boldsymbol{\hat{\beta}}^{\rm{AO}}}}
\newcommand{\btbh}{{\hat{\bar{\boldsymbol{\beta}}}}}

\newcommand{\Bc}{\mathcal{B}}
\newcommand{\Sc}{\mathcal{S}}

\newcommand{\Mc}{\mathcal{M}}

\newcommand{\Nn}{\mathcal{N}}

\newcommand{\vb}{\vct{v}}

\newcommand{\Ic}{{\mathcal{I}}}

\newcommand{\w}{\vct{w}}
\newcommand{\wo}{\boldsymbol{\omega}}
\newcommand{\wh}{\hat{\vct{w}}}
\newcommand{\wha}{\hat{\vct{w}}^{\rm{AO}}}

\newcommand{\ab}{\vct{a}}
\newcommand{\bb}{\vct{b}}
\newcommand{\ub}{{\vct{u}}}

\newcommand{\h}{\vct{h}}

\newcommand{\g}{{\vct{g}}}

\newcommand{\Tc}{\mathcal{T}}

\newcommand{\Fb}{\mathbb{F}}
\newcommand{\Tb}{\mathbb{T}}

\newcommand{\bz}{\boldsymbol{\zeta}}

\newcommand{\bg}{\boldsymbol{\phi}}

\newcommand{\st}{\star}

\newcommand{\x}{\vct{x}}

\newcommand{\y}{\vct{y}}

\newcommand{\bgl}{{~\big |~}}

\definecolor{emmanuel}{RGB}{255,127,0}

\newcommand{\R}{\mathbb{R}}
\newcommand{\Pro}{\mathbb{P}}

\renewcommand{\P}{\operatorname{\mathbb{P}}}
\newcommand{\E}{\operatorname{\mathbb{E}}}

\newcommand{\vct}[1]{\bm{#1}}
\newcommand{\mtx}[1]{\bm{#1}}

\newcommand{\Pc}{{\cal{P}}}
\newcommand{\X}{{\mtx{X}}}
\newcommand{\Xb}{{\mtx{\bar{X}}}}

\newcommand{\sSi}{\sqrt{\bSi}}

\newcommand{\Rb}{{\mtx{R}}}

\def \endprf{\hfill {\vrule height6pt width6pt depth0pt}\medskip}
\newenvironment{proof}{\noindent {\bf Proof} }{\endprf\par}

\def\wh{\widehat{\vct{w}}}

\newcommand{\bea}{\begin{align}}
\newcommand{\eea}{\end{align}}
\newcommand{\rP}{\stackrel{{P}}{\longrightarrow}}

\def\nn{\notag}

\def\tb{\vct{t}}
\newcommand{\simiid}{\stackrel{iid}{\sim}}

\newcommand{\Sigmab}{\boldsymbol{\Sigma}}

\newcommand{\betab}{\boldsymbol{\beta}}
\newcommand{\lai}{\bSi_{i,i}}

\newcommand{\gba}{\bar{\g}}
\newcommand{\hba}{\bar{\h}}

\newcommand{\ct}[1]{}%\marginpar{\color{blue}\tiny\ttfamily CT: #1}}
\newcommand{\cts}[1]{{\color{black}{#1}}}

\newcommand{\betas}{\betab^\star}
\newcommand{\ksi}{\xi}

\newcommand{\Ec}{\mathcal{E}}

\newcommand{\ggt}{\widetilde{g}}

\newcommand{\blat}{\bla^\st}
\usepackage[switch]{lineno}
\usepackage{color}
\newcommand{\fy}[1]{{\color{black}{#1}}}
\title{Provable Benefits of Overparameterization in Model Compression:\\From Double Descent to Neural Net Pruning}
\title{Provable Benefits of Overparameterization in Model Compression:\\From Double Descent to Pruning Neural Networks}

\author {
            Xiangyu Chang\textsuperscript{\rm 1}\quad
        Yingcong Li\textsuperscript{\rm 1}\quad
        Samet Oymak\textsuperscript{\rm 1} \quad
        Christos Thrampoulidis\textsuperscript{\rm 2}\footnote{The author names are in alphabetical order.}\\ 
}
\affiliations{
    \textsuperscript{\rm 1} University of California, Riverside \\
    \textsuperscript{\rm 2} University of British Columbia, Vancouver\\
    \{cxian008,~yli692,~soymak\}@ucr.edu,~cthrampo@ece.ubc.ca
}
\begin{document}

\maketitle
%\linenumbers
\begin{abstract}
Deep networks are typically trained with many more parameters than the size of the training dataset. Recent empirical evidence indicates that the practice of overparameterization not only benefits training large models, but also assists – perhaps counterintuitively – building lightweight models. Specifically, it suggests that overparameterization benefits model pruning / sparsification. This paper sheds light on these empirical findings by theoretically characterizing the high-dimensional asymptotics of model pruning in the overparameterized regime. The theory presented addresses the following core question: ``should one train a small model from the beginning, or first train a large model and then prune?''. We analytically identify regimes in which, even if the location of the most informative features is known, we are better off fitting a large model and then pruning rather than simply training with the known informative features. This leads to a new double descent in the training of sparse models: growing the original model, while preserving the target sparsity, improves the test accuracy as one moves beyond the overparameterization threshold. Our analysis further reveals the benefit of retraining by relating it to feature correlations. We find that the above phenomena are already present in linear and random-features models. Our technical approach advances the toolset of high-dimensional analysis and precisely characterizes the asymptotic distribution of over-parameterized least-squares. The intuition gained by analytically studying simpler models is numerically verified on neural networks.
\end{abstract}
%\begin{Keywords}
%\end{Keywords}

%We made minor modifications in the main text (first 10 pages), added a discussion section \ref{sec discuss}, and also clarified the exact locations when referencing the SM.
\section{Introduction}

Large model size and overparameterization in deep learning are known to improve generalization performance \cite{neyshabur2017geometry}, and, state-of-the-art deep neural networks (DNNs) can be outrageously large. However, such large models are not suitable for certain important application domains, such as mobile computing \cite{tan2019mnasnet,sandler2018mobilenetv2}. Pruning algorithms aim to address the challenge of building lightweight DNNs for such domains. While there are several pruning methods, their common goal is to compress large DNN models by removing weak connections/weights with minimal decline in accuracy. Here, a key empirical phenomenon is that {\em{it is often better to train and prune a large model rather than training a small model from scratch}}. Unfortunately, the mechanisms behind this phenomenon are poorly understood especially for practical gradient-based algorithms. This paper sheds light on this by answering: {\em{What are the optimization and generalization dynamics of pruning overparameterized models? Does gradient descent naturally select the good weights?}}

% for building good lightweight models 
%An important drawback of deep neural networks (DNN) is that they are computationally demanding 

%A core challenge in machine learning stems from {\em{limited hardware budgets}}. 
% with substantial computation cost e.g.~OpenAI's latest language model has 175 billion parameters and would cost \$4.6 million to train \cite{gpt3}
%State-of-the-art deep neural networks (DNN) can be outrageously large. At the other end of the spectrum, while mobile platforms such as smartphones and unmanned aerial vehicles provide key application domains for \ML, they require lightweight models that are energy and storage efficient during inference  
\vp
\noindent{\bf{Contributions:}} %Our core contribution is a thorough
%theoretical
We analytically study the performance of popular pruning strategies. First, we analyze linear models, and then, generalize the results to nonlinear feature maps. Through extensive simulations, we show that our analytical findings predict similar behaviors in more complex settings.  

%\textbf{(a)}
\noindent\textbf{(a)} {\bf{Distributional characterization (DC):}} The key innovation facilitating our results is a theoretical characterization of the distribution of the solution of overparameterized least-squares. This DC enables us to accurately answer {\em{``what happens to the accuracy if X\% of the weights are pruned?''}}.

\noindent\textbf{(b)} {\bf{Benefits of overparameterization:}} Using DC, we 
%exactly
 %and provably 
{obtain rigorous precise characterizations of  the pruning performance in linear problems. Furthermore, we use, so called ``linear gaussian equivalences", to obtain sharp analytic predictions for nonlinear maps, which we verify via extensive numerical simulations.}
%for pruning strategies and verify our predictions for nonlinear maps via establishing linear equivalences. 
 By training models of growing size and compressing them to fixed sparsity, we identify a novel double descent behavior, where the risk of the pruned model is 
 %often
{consistently} minimized in the overparameterized regime. Using our theory, we uncover rather surprising scenarios where pruning an overparameterized model is provably better than training a small model with the exact information of optimal nonzero locations. 
%\som{neural net does not keep the optimal locations}
%\ct{need to be careful here}

\noindent\textbf{(c)} {\bf{Benefits of retraining:}} An important aspect of pruning is retraining the model using the favorable nonzero locations identified during the initial training. We show that retraining can actually hurt the performance when features are uncorrelated. However, it becomes critical as correlations increase. Importantly, we devise the DC of the {\em{train$\rightarrow$prune$\rightarrow$retrain process}} (see Figs.~\ref{figRF} and \ref{figRF2} and the discussion around Def.~\ref{RT_def} for details), and, we demonstrate that it correctly captures the pruning performance of random features  that are known to be good proxies for understanding DNN behavior \cite{jacot2018neural}. 

{We anticipate that our techniques towards establishing the DC of the overparameterized problems might be useful, beyond the context of pruning, in other statistical inference tasks that require careful distributional studies.}
%Of general interest, we emphasize that our high-dimensional analysis of the complex procedures like pruning are entirely novel and we anticipate that our techniques can help address other interesting statistical learning tasks which require a careful distributional characterization.% This also reveals different pruning double-descents.
% are known to be a good proxy to understand the pruning in neural nets.

\begin{figure}[t!]
\centering
	\begin{tikzpicture}
	\node at (0,0) {\includegraphics[scale=0.33]{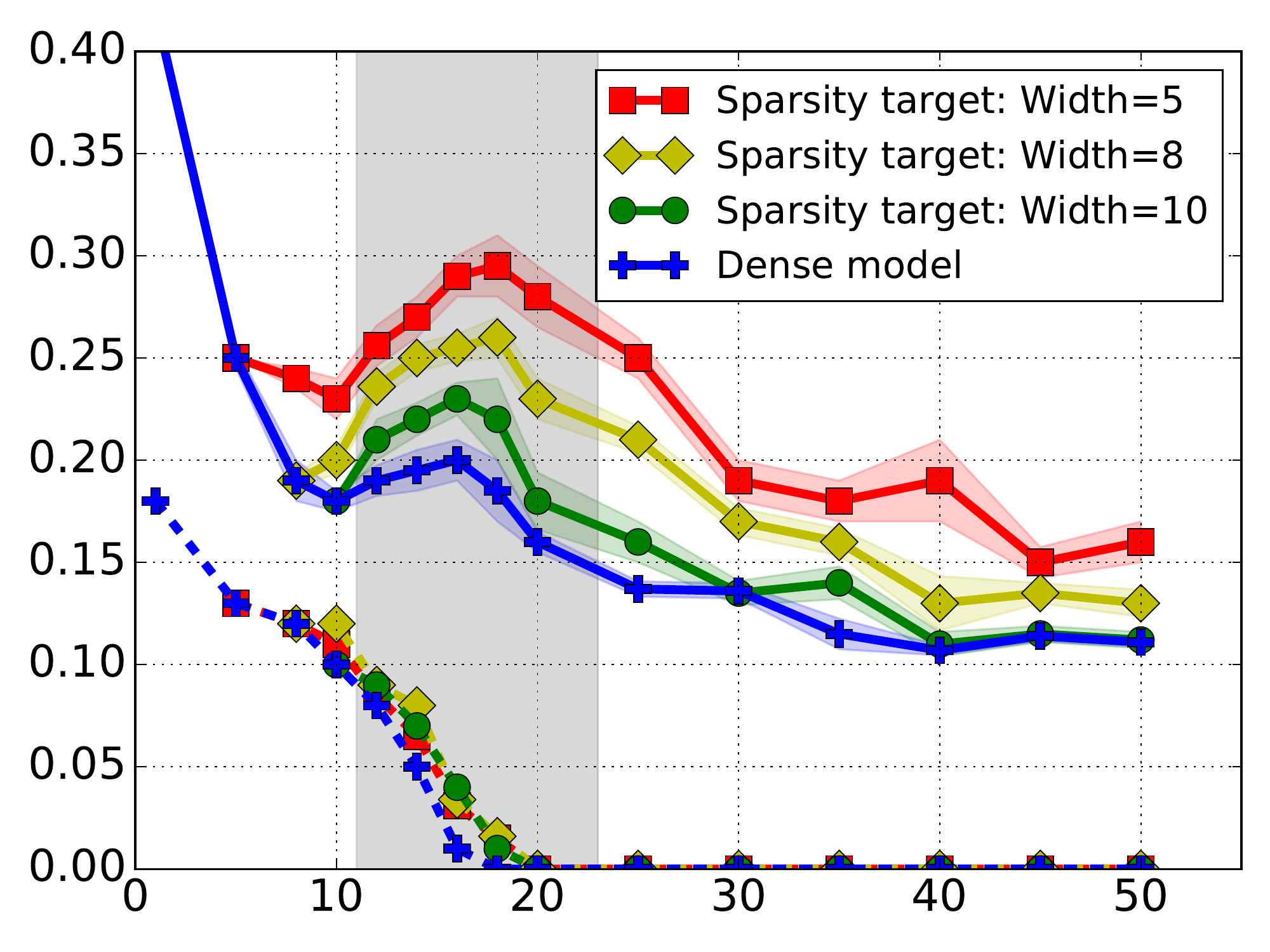}};
	\node at (-3.5,0) [rotate=90,scale=1.]{Training / Test Error};
	\node at (0,-2.6) [scale=.9]{Width Parameter (\# of filters $k$)};%Each bar in the figure shows the correlations' geometric mean of datasets in this group when training to corresponding non-zero level. 
%	\node (A) at (0, 0) {A};
%	\node (B) at (1, 1) {B};
%	\draw [->] (A) edge (B) 
	\end{tikzpicture}
	\vspace{-10pt}
	\caption{\small{We train sparse ResNet-20 models on the CIFAR-10 dataset with varying width (i.e. number of filters) and sparsity targets. The width parameter controls the overall model size. The solid (resp. dashed) lines are test (resp. training) errors. The blue line corresponds to training of a dense model with width-$k$. The other three curves correspond to sparsity targets $s\in\{5,8,10\}$, for which a dense model of width-$k$ is first pruned to achieve the exact same number of nonzeros as a dense model of width-$s$ and then retrained over the identified nonzero pattern. Surprisingly, all curves interpolate (achieve zero training error) around the same width parameter despite varying sparsity. The best test error is always achieved in the overparameterized regime (large width). Test error curves have two local minima which uncovers a novel double descent phenomena for pruning. {The shaded region highlights the transition to zero training error, where the test error peaks.}}}
	%We train and prune ResNet-20 models on the CIFAR-10 dataset with varying filter numbers (which modifies the total number of parameters) and sparsity targets. Solid and dotted lines are test and training error respectively. The dense line shows training results with $k$ filters. Correspondingly, the other two lines prune dense model to sparse ResNet-20 and make sure they have the same number of non-zero parameters as dense $s$-filter ResNet-20. From red and green curves, training and test error decrease with model size, implying that training and pruning a larger model preforms better than training small model.  Secondly, a novel double descent phenomena appears when training with sparse models .
	\label{figNN}\vspace{-15pt}
\end{figure}

%\somm{it might be good to move the sentence from appendix to the main body that this train error is actually the retrain error!}
\subsection{Prior Art}
%, as well as, optimization dynamics
This work relates to the literature on model compression and overparameterization in deep learning. 
%For analysis, we also use tools related to high-dimensional statistics \cite{thrampoulidis2015regularized,OymLAS,thrampoulidis2018precise,hastie2019surprises}. 
% by the problem structure
%Recent works use \emph{magnitude-based} \cite{han2015learning,frankle2019lottery,franklestabilizing} and \emph{Jacobian-based} \cite{shunshi2019one} pruning criteria, which are shown to achieve stellar performance. 

%There are various saliency-based approaches \cite{hassibi1993second,hassibi1994optimal,lecun1990optimal,dong2017learning} for neural net pruning to select .
\vs\noindent {\bf{Neural network pruning:}} Large model sizes in deep learning have led to a substantial interest in model pruning/quantization \cite{han2015deep,hassibi1993second,lecun1990optimal}. DNN pruning has a diverse literature with various architectural, algorithmic, and hardware considerations \cite{sze2017efficient,han2015learning}. The pruning algorithms can be applied before, during, or after training a dense model \cite{lee2018snip,wang2020picking,jin2016training,oymak2018learning} and in this work we focus on after training. Related to over-parameterizarion, \cite{frankle2018lottery} shows that a large DNN contains a small subset of favorable weights (for pruning), which can achieve similar performance to the original network when trained with the same initialization. \cite{zhou2019deconstructing,malach2020proving,pensia2020optimal} demonstrate that there are subsets with good test performance even without any training and provide theoretical guarantees. However, these works do not answer why practical gradient-based algorithms lead to good pruning outcomes. {Closer to us, \cite{li2020exploring} derives formulas for predicting the pruning performance of over-parameterized least-squares without proofs. In contrast, we provide provable guarantees, and, also obtain DC for more complex problems with general design matrices and nonlinearities.}

%For example, \cite{lee2018snip,wang2020picking} \ct{I don't understand this sentence}prune the network before training by the connection sensitivity and gradient flow, respectively. Furthermore, \cite{aghasi2017net,oymak2018learning,jin2016training} use $\ell_1$-penalization pruning strategies, for which they provide provable performance guarantees. 

\begin{figure}[t!]
\centering
	\begin{tikzpicture}
	\node at (0,0) {\includegraphics[scale=0.33]{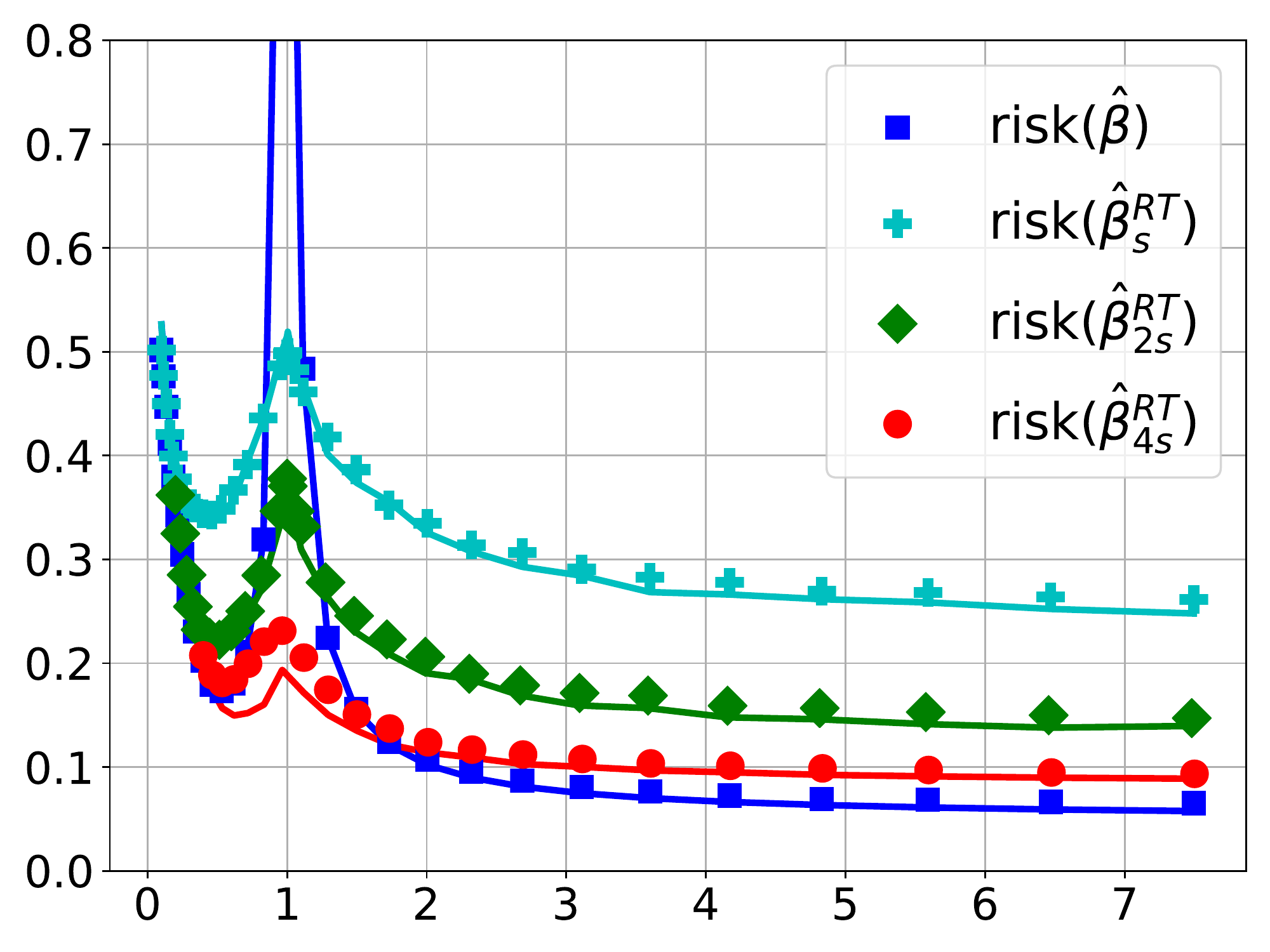}};
	\node at (-3.5,0) [rotate=90,scale=1.]{Test Risk};
	\node at (0,-2.6) [scale=.9]{Overparameterization ($p/n$)};%Each bar in the figure shows the correlations' geometric mean of datasets in this group when training to corresponding non-zero level. 
	\end{tikzpicture}
	\vspace{-10pt}
	\caption{\small{Random feature regression (RFR) with ReLU feature-map $\phi(\ab)=\text{ReLU}(\Rb\ab)$. Here $\Rb$ has i.i.d.~standard normal entries corresponding to the input layer of a shallow neural net and we regress the output layer. Solid lines follow from our distributional characterization and the markers are obtained by solving random feature regression, which exhibit a good match. The blue line is the performance of usual RFR with growing number of features $p$. The other lines are obtained by solving RFR with $p$ features and pruning and retraining the solution to fixed sparsity levels ($s,2s,4s$) with $s/n=0.1$. Importantly, the risks of the retrained models exhibit double descent and are minimized when $p\gg n$ despite fixed model size / sparsity. The slight mismatch of the red curve/markers is explained in Fig.~\ref{figRF2}.}}%,mei2019generalization
	\cmt{We emphasize that RFR is a well-recognized proxy for the DNN regression due to the equivalence between kernels and wide DNNs \cite{jacot2018neural}.}
	\label{figRF}\vspace{-15pt}
\end{figure}
\cmt{The green line prunes the resulting coefficients to find a sparse model with fixed sparsity $s/n=0.1$. The red line resolves RFR using the features identified after pruning. All lines exhibit double descent at $n=p$.}
%oymak2019overparameterized
\noindent {\bf{Benefits of overparameterization:}} Studies on the optimization and generalization properties of DNNs demonstrate that overparameterization acts as a catalyst for learning. \cite{arora2018optimization,neyshabur2014search,gunasekar2017implicit,ji2018risk} argue that gradient-based algorithms are implicitly biased towards certain favorable solutions (even without  explicit regularization) to explain benign overfitting \cite{bartlett2020benign,oymak2020towards,du2019gradient,chizat2019lazy,belkin2018understand,belkin2019does,tsigler2020benign,liang2018just,mei2019generalization,ju2020overfitting}. More recently, these studies have led to interesting connections between kernels and DNNs and a flurry of theoretical developments. Closest to us, \cite{nakkiran2019deep,belkin2019two,belkin2019reconciling} uncover a double-descent phenomenon: the test risk has two minima as a function of model size. One minimum occurs in the classical underparameterized regime whereas the other minimum occurs when the model is overparameterized and the latter risk can in fact be better than former. Closer to our theory, \cite{derezinski2019exact,hastie2019surprises,montanari2019generalization,deng2019model,kini2020analytic,liang2020precise,salehi2020performance,ju2020overfitting} characterize the asymptotic performance of overparameterized learning problems. However these works are limited to characterizing the test error of regular (dense) training. In contrast, we use distributional characterization (DC) to capture the performance of more challenging pruning strategies and we uncover novel double descent phenomena (see Fig.~\ref{figNN}).

%A related line of work connects the benefits of overparameterization to the double descent phenomena \cite{nakkiran2019deep,belkin2019two,belkin2019reconciling,hastie2019surprises,deng2019model}. 

\cmt{For linear models, implicit bias phenomena have been studied for various loss functions and algorithms (e.g.~logistic loss converging to max-margin solution on separable data) \cite{ji2018risk,soudry2018implicit,nacson2019stochastic}. Follow-up works show that such results continue to hold for nonlinear problems \cite{gunasekar2017implicit,oymak2019overparameterized,azizan2018stochastic}. More recently, this line of works has further motivated the study of generalization/optimization guarantees for deep networks and their connections to random features \cite{du2019gradient,allen2019convergence,chizat2019lazy,belkin2018understand,belkin2019does,liang2018just,mei2019generalization}.}

%\som{Circle and plus markers use the same samples and fresh samples for retraining which lead to the same performance.}

\section{Problem Setup}\label{sec:setup}
%$\Fb_s$ be the operator that returns the first $s$ entries of a vector and 
Let us fix the notation. Let $[p]=\{1,2,\dots,p\}$. Given $\bt\in \R^p$, let $\Tb_s(\bt)$ be the pruning operator that sets the smallest $p-s$ entries in absolute value of $\bt$ to zero. Let $\Ic(\bt)\subset[p]$ return the index of the nonzero entries of $\bt$. $\Iden_n$ denotes the $n\times n$ identity matrix and $\Nn(\bmu,\bSi)$ denotes the normal distribution with mean $\bmu$ and covariance $\bSi$. $\X^\dagger$ denotes the pseudoinverse of matrix $\X$. %Throughout, we will assume that the covariance matrices are diagonal. \dots
%\distas\Dc
%et $\Dc$ be a data distribution over $\R^d\times \R$ and l

\vp
\noindent \textbf{Data:} Let $(\ab_i,y_i)_{i=1}^n\subset \R^d\times \R$ with i.i.d.~input-label pairs. Let $\phi(\cdot):\R^d\rightarrow\R^p$ be a (nonlinear) feature map. We generate $\x_i=\phi(\ab_i)$ and work with the dataset $\Sc=(\x_i,y_i)_{i=1}^n$ coming i.i.d. from some distribution $\Dc$. As an example, of special interest to the rest of the paper, consider random feature regression, where $\x_i=\psi(\Rb\ab_i)$ for a nonlinear activation function $\psi$ that acts entry-wise and a random matrix $\Rb\in \R^{p\times d}$ with i.i.d.~$\Nn(0,1)$ entries; see Fig.~\ref{figRF}. In matrix notation, we let $\y=[y_1~\dots~y_n]^T\in\R^n$ and $\X=[\x_1~\dots~\x_n]^T\in\R^{n\times p}$ denote the vector of labels and the feature matrix, respectively.
%Gather the labels $\y=[y_1~\dots~y_n]^T\in\R^n$ and features $\X=[\x_1~\dots~\x_n]^T\in\R^{n\times p}$ in matrix notation. 
Throughout, we focus on regression tasks, in which the training and the test risks of a model $\bt$ are defined as% on the feature map 
\begin{align}
&\text{Population risk:}~~~\Lc(\bt)=\E_{\Dc}[(y-\x^T\bt)^2].\label{test formula}\\
&\text{Empirical risk:}~~~~\Lch(\bt)=\frac{1}{n}\tn{\y-\X\bt}^2.\label{erm}
\end{align}
 %As illustrated in Fig.~\ref{figRF}, a specific feature map we shall consider is the random feature regression where $\x_i=\psi(\Rb\ab_i)$ for some nonlinear activation $\psi$ which applies entrywise and a random matrix $\Rb\in \R^{p\times d}$ with i.i.d.~$\Nn(0,1)$ entries. 
 \cmt{Let $\phi(\cdot):\R^d\rightarrow \R^k$ be a nonlinear map for feature extraction. $\phi$ will be important for random feature analysis.} 
% For regression tasks using feature map $\phi$, the training and test risks of a model $\bt$ is defined as% on the feature map 
%\begin{align}
%&\text{Population risk:}~~~~~\Lc(\bt)=\E_{\Dc}[(y-\x^T\bt)^2].\label{test formula}\\
%&\text{Empirical risk:}~\Lch(\bt)=\frac{1}{n}\sum_{i=1}^n\tn{\y-\X\bt}^2.\label{erm}
%\end{align}
During training, we will solve the empirical risk minimization (ERM) problem over a set of selected features $\Delta\subset[p]$, from which we obtain the least-squares solution
\begin{align}
\label{eq:ERM}
\bth(\Delta)=\arg\min_{\bt\,:\,\Ic(\bt)=\Delta} \Lch(\bt).
\end{align}
For example, regular ERM corresponds to $\Delta=[p]$, and we simply use $\bth=\bth([p])$ to denote its solution above. Let $\bSi=\E[\x\x^T]$ be the covariance matrix and $\bb=\E[y\x]$ be the cross-covariance. The parameter minimizing the test error is given by $\bt^\st=\bSi^{-1}\bb$. We are interested in training a model over the training set $\Sc$ that not only achieves small test error, but also, it is sparse. We do this as follows. First, we run stochastic gradient descent (SGD) to minimize the empirical risk (starting from zero initialization). It is common knowledge that SGD on least-squares converges to the minimum $\ell_2$ norm solution given by $\bth=\X^\dagger \y$.  Next, we describe our pruning strategies to compress the model.

\vp
\noindent \textbf{Pruning strategies:} Given dataset $\Sc$ and target sparsity level $s$, a pruning function $P$ takes a model $\bt$ as input and outputs an $s$-sparse model $\bt^P_s$. Two popular pruning functions are magnitude-based (MP) and Hessian-based (HP) (a.k.a.~optimal brain damage) pruning \cite{lecun1990optimal}. The latter uses a diagonal approximation of the covariance via $\hat{\bSi}=\text{diag}(\X^T\X)/n$ to capture \emph{saliency} (see \eqref{saliency eq}). Formally, we have the following definitions:
%\cite{dobriban2018high}
\begin{itemize}
\item {\em{Magnitude-based pruning:}} $\bt^M_s=\Tb_s(\bt)$.
\item {\em{Hessian-based pruning:}} $\bt^H_s=\hat{\bSi}^{-1/2}\Tb_s(\hat{\bSi}^{1/2}\bt)$.
\item {\em{Oracle pruning:}} Let $\Delta^\st\subset[p]$ be the optimal $s$ indices so that $\bth(\Delta^\st)$ achieves the minimum population risk (in expectation over $\Sc$) among all $\bth(\Delta)$ and any subset $\Delta$ in \eqref{eq:ERM}.  When $\bSi$ is diagonal and $s<n$, using rather classical results, it can be shown that (see Lemma 7 in the Supplementary Material (SM)) these {\em{oracle indices}} are the ones with the top-$s$ saliency score given by %\vspace*{-20pt}.
\begin{align}
\text{Saliency score}=\bSi_{i,i}{\bt^\st_i}^2.\label{saliency eq}
\end{align} Oracle pruning {employs these latent saliency scores and} returns $\bt^O_s$ by pruning the weights of $\bt$ outside of $\Delta^\st$.
\end{itemize}
%A good baseline for pruning is training a small model by solving ERM with $s$ features. 
We remark that our distributional characterization might allow us to study more complex pruning strategies, such as optimal brain surgeon \cite{hassibi1994optimal}. However, we restrict our attention to the three aforementioned core strategies to keep the discussion focused. 

\noindent\textbf{Pruning algorithm:} To shed light on contemporary pruning practices, we will study the following three-stage {\em{train$\rightarrow$prune$\rightarrow$retrain}} algorithms.
\begin{enumerate}
\item Find the empirical risk minimizer $\bth=\X^\dagger \y$.
\item Prune $\bth$ with strategy $P$ to obtain $\bth^P_s$.
\item {\em{Retraining:}} Obtain $\bth^{RT}_s=\bth(\Ic(\bth^P_s))$.
\end{enumerate}
The last step obtains a new $s$-sparse model by solving ERM in \eqref{eq:ERM} with the features $\Delta=\Ic(\bth^P_s)$ identified by pruning. Figures \ref{figNN} and \ref{figRF} illustrate the performance of this procedure for ResNet-20~on the CIFAR-10 dataset and for a random feature regression on a synthetic problem, respectively . Our analytic formulas for RF, as seen in Fig.~\ref{figRF}, very closely match the empirical observations (see Sec.~\ref{sec mot}~for further explanations). Interestingly, the arguably simpler RF model already captures key behaviors (double-descent, better performance in the overparameterized regime, performance of sparse model comparable to large model) in ResNet.
Sections \ref{sec mot} and \ref{sec intu} present numerical experiments on pruning that verify our analytical predictions, as well as, our insights on the fundamental principles behind the roles of overparameterization and retraining. Sec \ref{sec main} establishes our theory on the DC of $\bth$ and provable guarantees on pruning. All proofs are deferred to the Supplementary Material (SM).%of all our results 
%We will then explore when and how retraining stage is critical for the pruning performance. 

%
\section{Motivating Examples}\label{sec mot}
%\newpage
%\newpage
% (Chang, Lizzie please fill here)
%\subsection{Verifying Analytical Predictions}

\begin{figure}[t!]
	\begin{subfigure}{1.5in}\vspace{-5pt}
		\centering
		\begin{tikzpicture}
		\node at (0,0) {\includegraphics[scale=0.2]{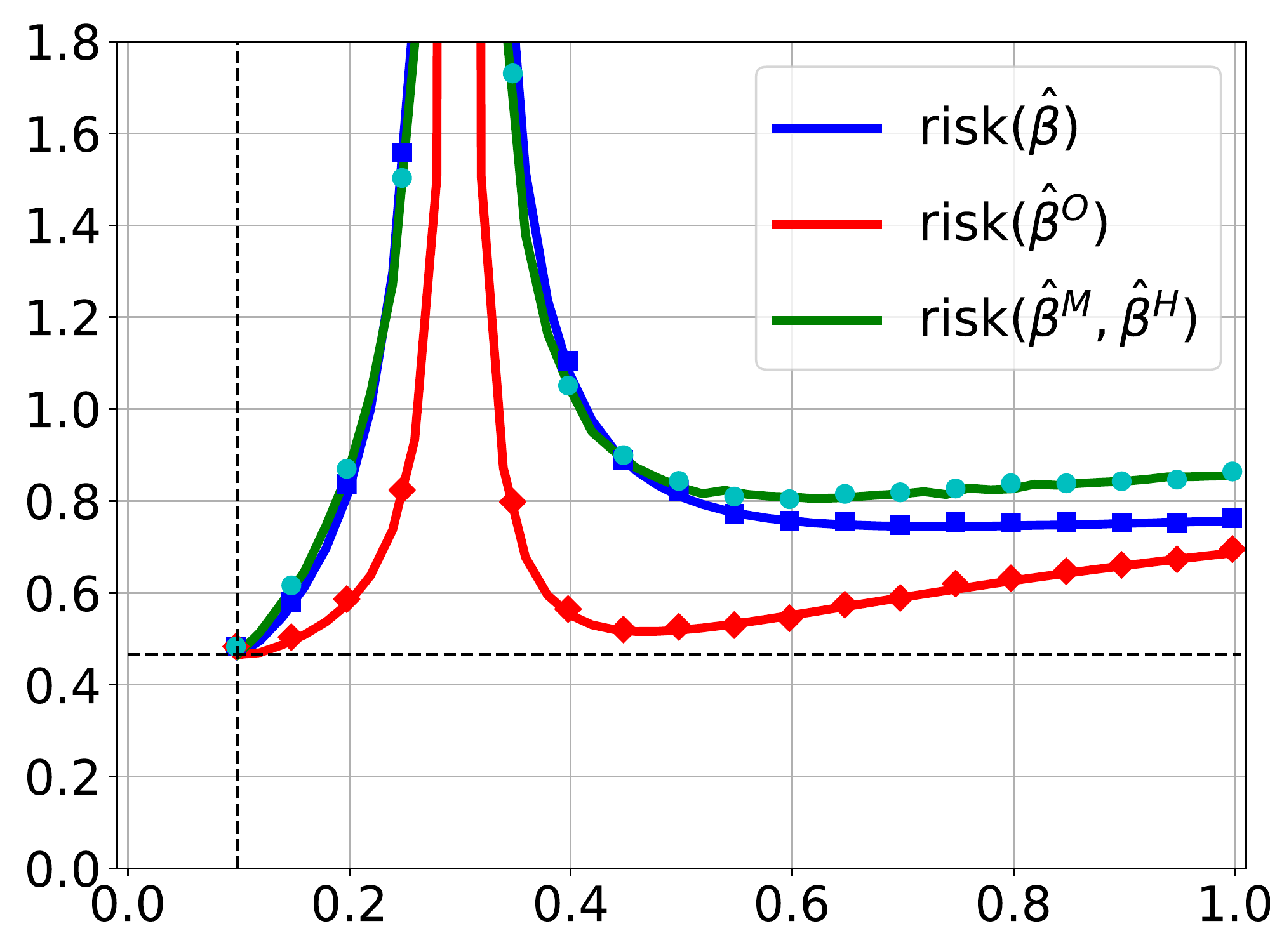}};
		\node at (-2.1,0) [rotate=90,scale=.9]{Test Risk};
		\node at (0,-1.7) [scale=.9]{Problem Size ($k/p$)};%Each bar in the figure shows the correlations' geometric mean of datasets in this group when training to corresponding non-zero level. 
		\end{tikzpicture}\caption{\small{Identity covariance, spiked latent weights.}}\label{fig1a}\vspace{-0.2cm}
	\end{subfigure}~~~~~\begin{subfigure}{1.5in}\vspace{-5pt}
	\centering
	\begin{tikzpicture}
	\node at (0,0) {\includegraphics[scale=0.2]{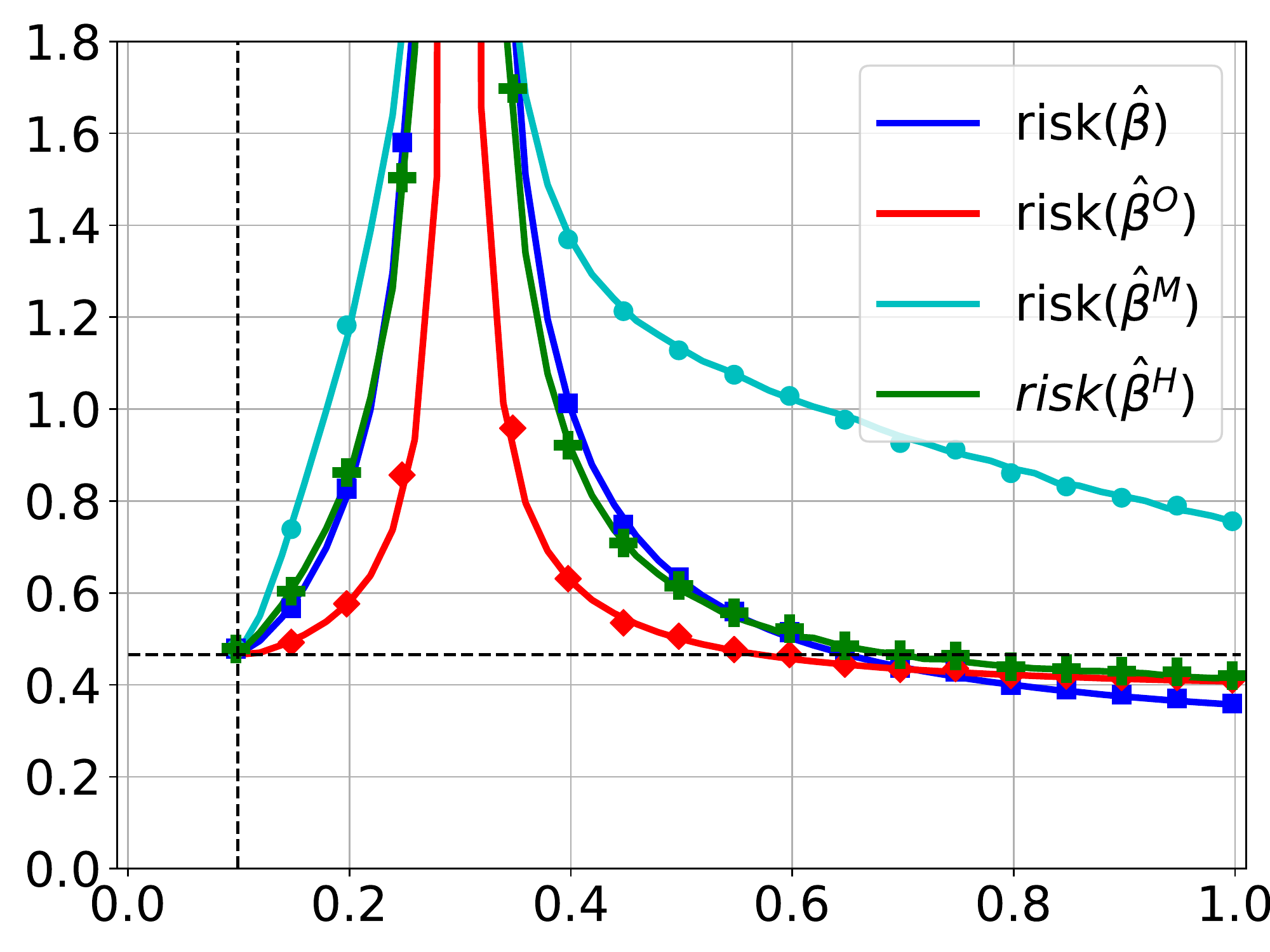}};
	%\node at (-2.4,0) [rotate=90,scale=1.]{Test Risk};
	\node at (0,-1.7) [scale=.9]{Problem Size ($k/p$)};%Each bar in the figure shows the correlations' geometric mean of datasets in this group when training to corresponding non-zero level. 
	\end{tikzpicture}\caption{\small{Spiked covariance, identical latent weights.}}\label{fig1b}\vspace{-0.2cm}
\end{subfigure}\caption{\small{Our theoretical predictions for various pruning strategies in linear models with $s/p=0.1$ and $n/p=0.3$. We solve ERM using the first $k$ features and then prune to obtain an $s$-sparse model. The vertical dashed line shows the $k=s$ point. The horizontal dashed line highlights the minimum risk among all underparameterized solutions ($k\leq n$) and all solutions obtained by a final retraining. \fy{Retraining curves are omitted here, but they can be found in Fig.~7 of SM.}}}\label{fig1}\vspace{-15pt}
\end{figure}
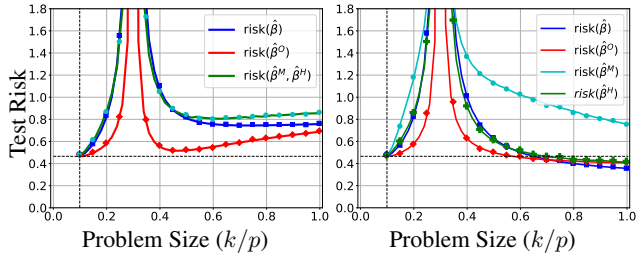
\subsection{Linear Gaussian Problems} 
We begin our study with linear Gaussian problems (LGP), which we formally define as follows. 
\begin{definition}[Linear Gaussian Problem (LGP)] \label{def LGP}Given latent vector $\bt^\st\in\R^d$, covariance $\bSi$ and noise level $\sigma$, assume that each example in $\Sc$ is generated independently as $y_i=\x_i^T\bt^\st+ \sigma z_i$ where $z_i\sim\Nn(0,1)$ and $\x_i\sim\Nn(0,\bSi)$. Additionally, the map $\phi(\cdot)$ is identity and $p=d$.
\end{definition}
Albeit simple, LGPs are of fundamental importance for the following reasons: (1) We show in Sec. \ref{sec main} that our theoretical framework rigorously characterizes pruning strategies for LGPs; (2) Through a ``linear Gaussian equivalence", we will use our results for linear models to obtain analytic predictions for nonlinear random features; (3) Our theoretical predictions and numerical experiments discussed next demonstrate that LGPs already capture key phenomena observed in more complex models (e.g., Fig. \ref{figNN}).
%. Moreover, 
%We first start with LGP experiments that are perfectly characterized within our theoretical framework. 

In Fig.~\ref{fig1}, we consider LGPs with diagonal covariance $\bSi$. We set  the sparsity  level $s/p=0.1$ and the relative dataset size $n/p=0.3$. To parameterize the covariance and $\bt^\st$, we use a {\em{spiked}} vector $\bla$, the first $s$ entries of which are set equal to $C=25\gg 1$ and the remaining entries equal to $1$. $\bla$ corresponds to the latent saliency score (cf. ~\eqref{saliency eq}) of the indices. To understand the role of overparameterization, we vary the number of features used in the optimization. Specifically, we solve \eqref{eq:ERM} with $\Delta=[k]$ and vary the number of features $k$ from $0$ to $p$. Here we consider the {\em{train$\rightarrow$prune}} algorithm, where we first solve for $\bth([k])$ and obtain our pruned model $\bth^P_s([k])$ by applying magnitude, Hessian or Oracle pruning (cf. ~$P\in \{M,H,O\}$). Since $\bla$ is non-increasing, the indices are sorted by saliency score; thus, Oracle pruning always picks the first $s$ indices. Solid lines represent analytic predictions, while markers are empirical results.
 %Our theoretical predictions and empirical results are shown in Fig.~\ref{fig1}, with simplified notation $\bth^P_s([k])\rightarrow \bth^P$. 
 The vertical dashed line is the sparsity level $s/p$. 

\cmt{confusing about why setups of Fig.~\ref{fig1a}, \ref{fig1b}. Are we trying to keep saliency score the same?}
In Fig.~\ref{fig1a}, we set $\bSi=\Iden_p$ and $\bt^\st=\sqrt{\bla}$. Note, that the analytic curves correctly predict the test risk and the double descent behavior. Observe that the Hessian and Magnitude pruning coincide here, since the diagonal of the empirical covariance is essentially identity. In contrast, Fig.~\ref{fig1b} emphasizes the role of the feature covariance by setting $\bSi=\text{diag}(\bla)$ and $\bt^\st$ to be the all ones vector. In this scenario, we observe that Hessian pruning performs better compared to Fig.~\ref{fig1a} and also outperforms Magnitude pruning. This is because the empirical covariance helps distinguish the salient indices. Importantly, for Hessian and Oracle pruning, the optimal sparse model is achieved in the highly overparameterized regime $k=p$. Notably, the achieved performance at $k=p$ is strictly better than the horizontal dashed line,  which highlights the optimal risk among all underparameterized solutions $k\leq n$ and all retraining solutions (see also SM Sec. A). This has two striking consequences. First, {\em{retraining can in fact hurt the performance}}; because the {\em{train$\rightarrow$prune}} performance at $k=p$ is strictly better than  {\em{train$\rightarrow$prune$\rightarrow$retrain}} for all $k$. Second, {\em{overparameterized pruning can be provably better than solving the sparse model with the knowledge of the most salient features}} as $k=p$ is also strictly better than $k=s$.
%{\em{train$\rightarrow$prune}} with 
% \textcolor{red}{this will be explained in Sec.~\ref{sec intu}}. % (also see SM A)
%The reason is that the population risk of retraining is guaranteed to be upper bounded by the risk of $\bth_s([s])$ which uses optimal features.
%Recalling features are sorted with decreasing saliency and observing $\bth_s([s])$ solves ERM with the knowledge of optimal feature locations, this leads to our initial claim.

\begin{figure}[t!]
\centering
	\begin{tikzpicture}
	\node at (0,0) {\includegraphics[scale=0.28]{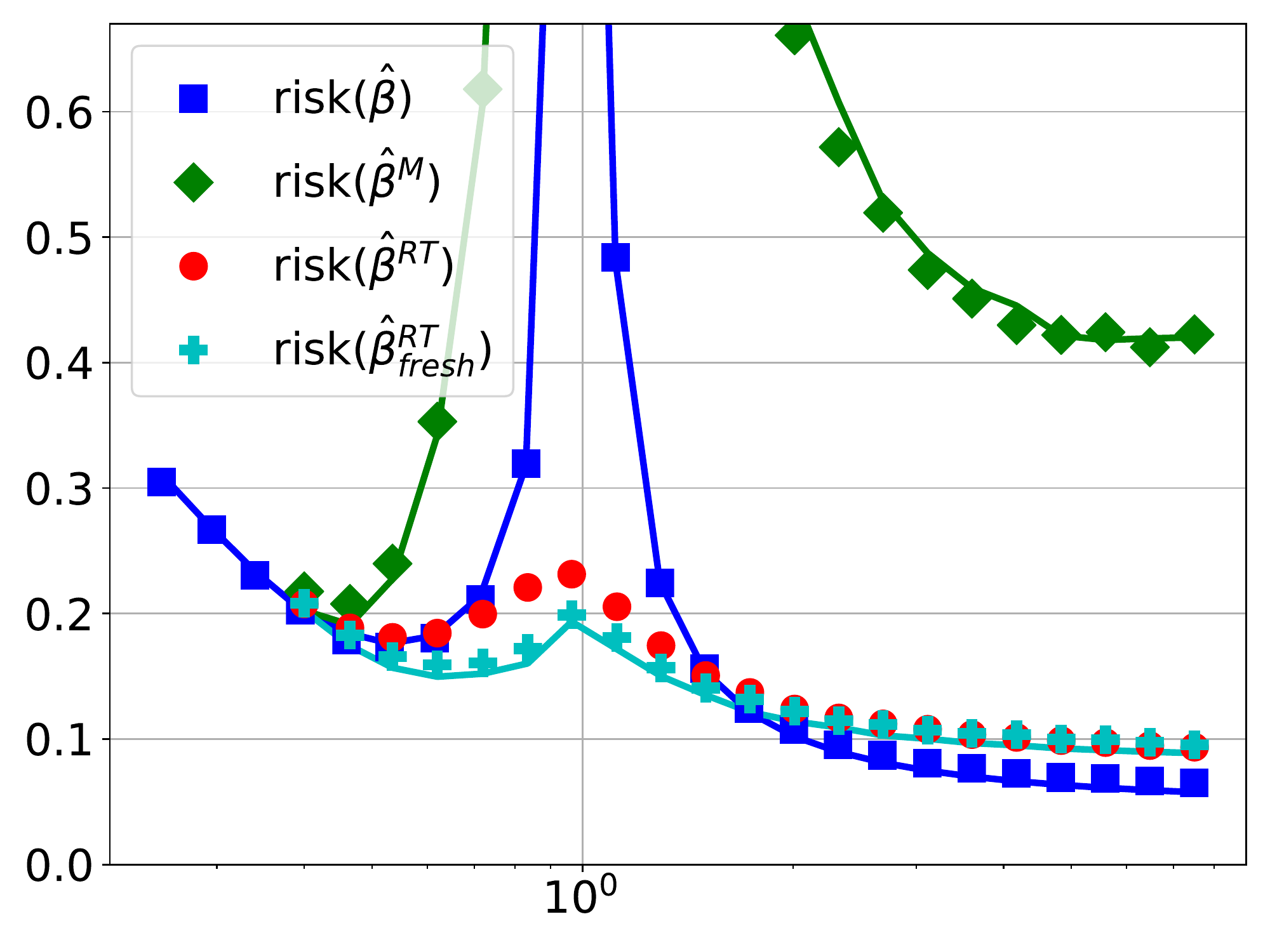}};
	\node at (-3.1,0) [rotate=90,scale=1.]{Test Risk};
	\node at (0,-2.2) [scale=.9]{Overparameterization ($p/n$)};%Each bar in the figure shows the correlations' geometric mean of datasets in this group when training to corresponding non-zero level. 
	\end{tikzpicture}
	\vspace{-10pt}
	\caption{\small{Illustration of the mismatch between pruning with retraining (red markers) and pruning with fresh samples (cyan markers/line). The setting here is exactly the same as in  Fig.~\ref{figRF}, but we only show the case of sparsity $4s$ for which the mismatch is observed. Observe that our analytical predictions accurately capture the risk of retraining with fresh samples. However, we observe a discrepancy with the true risk of retraining (without fresh samples) around the interpolation threshold. Also shown the risk of the original ERM solution before pruning (in blue) and of the magnitude-pruned model (before any retraining).
%	Same setup as Fig.~\ref{figRF} however we restrict our attention to sparsity $4s$. Green curve is pruning without retraining. Cyan curve retrains with independent samples (unlike red curve). This
	}
	}
	\label{figRF2}\vspace{-15pt}
\end{figure}
\subsection{Random Features Regression} We relate an ERM problem \eqref{eq:ERM} with nonlinear map $\phi$ to an equivalent LGP. This will allow us to use our theoretical results about the latter to characterize the properties of the original nonlinear map. We ensure the equivalence by properly setting up the LGP to exhibit similar second order statistics as the original problem.
\begin{definition}[Equivalent Linear Problem] Given distribution $(\x,y)$ $\sim\Dc$, the equivalent LGP($\bt,\bSi,\sigma$) with $n$ samples is given with parameters $\bSi=\E[\x\x^T]$, $\bt^\st=\bSi^{-1}\E[y\x]$ and $\sigma=\E[(y-\x^T\bt^\st)^2]^{1/2}$. 
\end{definition}
%from equivalent LGP to the original problem  %It should be emphasized that it is important
In Section \ref{sec main}, we formalize the DC of LGPs, which enables us to characterize pruning/retraining dynamics. Then, we empirically verify that DC and pruning dynamics of equivalent LGPs can successfully predict the original problem \eqref{eq:ERM} with non-linear features. The idea of setting up and studying equivalent LGPs as a proxy to nonlinear models, has been recently used in the emerging literature of high-dimensional learning, for predicting the performance of the original ERM task \cite{montanari2019generalization,goldt2020gaussian,abbasi2019universality,derezinski2019exact}. This work goes beyond prior art, which focuses on ERM, by demonstrating that we can also successfully predict the pruning/retraining dynamics. Formalizing the performance equivalence between LGP and equivalent problem is an important future research avenue and it can presumably be accomplished by building on the recent high-dimensional universality results such as \cite{oymak2018universality,hu2020universality,abbasi2019universality,goldt2020gaussian}.

In Fig. \ref{figRF}, we study random feature regression to approximate a synthetic nonlinear distribution. Specifically, data has the following distribution: Given input $\ab\sim\Nn(0,\Iden_d)$, we generate random unit norm $\bt^1\in\R^d,\bt^2\in\R^d$ and set the label to be a quadratic function given by $y=\ab^T\bt^1+(\ab^T\bt^2)^2$. Then, we fix $\Rb\distas\Nn(0,1)$ and we generate ReLU features $\x=\text{ReLU}(\Rb\ab)$, where $\Rb$ corresponds to the input layer of a two-layer network. The markers in Fig. \ref{figRF} are obtained by solving RFR and pruning and retraining with varying sparsity targets ($s,2s,4s$ with $s/n=10\%$). Here, $d=10, n=200$. For each marker, the results are averages of 50 $\Rb\in\R^{p\times d}$ realizations and 10 iterations for each choice of $\Rb$. The lines are obtained via our DC of the equivalent LGP (by using Defs.~\ref{aux_def} and \ref{RT_def}) where the latent parameter $\bt^\st$, noise $\sigma$ and the covariance $\bSi$ of the RFR problem are calculated for fixed realization of the input layer $\Rb$ (similarly averaged over 50 random $\Rb$). Our theory and empirical curves exhibit a good match. The results demonstrate the importance of overparameterization for RF pruning, which corresponds to picking {\em{random features smartly}}. Here, the coefficients of least-squares act like a scoring function for the saliency of random features and capture how well they are aligned with the target function. The fact that the risk of the pruned models is minimized in the overparameterized regime implies that least-squares regression succeeds in properly selecting salient random features from a larger candidate set. In the context of deep learning, our discussion can be interpreted as {\em{pruning hidden nodes of the network}}.

\noindent\textbf{Predicting retraining performance.} As discussed in Sec.~\ref{sec main} and Def.~\ref{RT_def}, for the retraining stage, our DC is accomplished by assuming that retraining phase uses $n$ fresh training examples (i.e.~a new dataset $\Sc_{\text{fresh}}$). Let us denote the resulting model by $\bth^{RT}_{\text{fresh}}$. Perhaps surprisingly, Fig.~\ref{figRF} shows that this DC correctly captures the performance of $\bth^{RT}$ with the exception of the red curve ($4s$). Fig.~\ref{figRF2} focuses on this instance and shows that our DC indeed perfectly predicts the fresh retraining performance and verifies the slight empirical mismatch between $\bth^{RT}$ and $\bth^{RT}_{\text{fresh}}$.
\vspace{-6pt}
\subsection{Neural Network Experiments}\label{ssec:NNE}
Finally, we study pruning deep neural networks.
 %to see the pruning performance under complex models. 
Inspired by  \cite{nakkiran2019deep}, we train ResNet-20 with changeable filters over CIFAR-10. Here, the filter number $k$ is equivalent to the width/channel of the model. As the width of ResNet-20 changes, the fitting performance of the dataset varies. Here, we apply {\em{train$\rightarrow$prune$\rightarrow$retrain}}.
%, since neural networks behave approximately to random feature regression where retraining is crucial to maintain the performance. 
Select $s$ as the sparsity target and $s$-filter ResNet-20 model as the base model with $N_s$ parameters. First, we train a dense model with $k$ filters and $N_k$ parameters, $N_k\gg N_s$, and prune it by only keeping the largest $N_s$ entries in absolute value non-zero. $N_k$ grows approximately quadratically in $k$. Now, the sparse model shares the same number of parameters amenable to training as the base model. Finally, we retain the pruned network and record its performance on the same dataset and same configuration. 
In Fig.~\ref{figNN}, we plot the training and test errors of dense and sparse models. All the neural experiments are trained with Adam optimization and $0.001$ learning rate for $1000$ epochs, with data augmentation. Green, yellow and red lines correspond to $5$, $8$ or $10$ sparsity targets, with around 28,000, 70,000 and 109,000 trainable parameters, respectively. As the width $k$ grows, the training and test errors decrease for all $5$-,$8$-,$10$-filter base models, except for the shaded double descent range. These experiments again verify the main insight revealed to us by studying simpler linear and random-feature models, that is, training a larger model, followed by appropriate pruning, can preform better than training a small model from the beginning.
%
%Our initial argument is proved in neural network experiments that pruning a larger model and pruning could preform better than training a small model from from the beginning. 
Another worth-mentioning observation is that with appropriate sparsity level (here, $10$) the pruned model has prediction performance comparable to the dense model. Finally and interestingly, the test error dynamics of the pruned model exhibit a double descent that resembles that of the dense model (previously observed in \cite{nakkiran2019deep}). 
%More, there is similar double descent behavior in training sparse model as dense model. 

%\newpage

\subsection{Further Intuitions on The Denoising Effect of Overparameterization}
To provide further insights into the pruning benefits of overparameterization, consider a simple linear model (as in Def~\ref{def LGP}) with $n\geq p\geq s$, noise level $\sigma=0$ and identity covariance. Suppose our goal is estimating the coefficients $\bt^\st_{\Delta}$ for some fixed index set $\Delta\subset[p]$ with $|\Delta|=s$. For pruning, we can pick $\Delta$ to be the most salient/largest entries. If we solve the smaller regression problem over $\Delta$, $\bth(\Delta)$ will only provide a noisy estimate of $\bt^\st_{\Delta}$. The reason is that, the signal energy of the missing features $[p]-\Delta$ acts as a noise uncorrelated with the features in $\Delta$. Conversely, if we solve ERM with all features (the larger problem), we perfectly recover  $\bt^\st$ due to zero noise and invertibility ($n\geq p$). Then one can also perfectly estimate $\bt^\st_{\Delta}$. This simple argument, which is partly inspired by the missing feature setup in \cite{hastie2019surprises}, shows that solving the larger problem with more parameters can have a ``denoising-like effect" and perform better than the small problem. Our contribution obviously goes well beyond this discussion and theoretically characterizes the exact asymptotics, handles the general covariance model and all $(n,p,s)$ regimes, and also highlights the importance of the overparameterized regime $n\ll p$.
%\textcolor{red}{It also explains why retraining again with few features can potentially hurt the performance. This is because when features are uncorrelated (e.g.~diagonal covariance), missing features (due to the smaller problem size) act like additive uncorrelated noise.} 
\section{Understanding the Benefits of Retraining}\label{sec intu}

On the one hand, the study of LGPs in \fx{Fig.~\ref{fig1} and Fig.~7 of SM} suggest that retraining can actually hurt the performance. On the other hand, in practice and in the RFR experiments of Fig.~\ref{figRF2}, retraining is crucial; compare the green $\bth^M$ and red $\bth^{RT}$ curves and see SM Section A for further DNN experiments. Here, we argue that the benefit of retraining is connected to the correlations between input features. Indeed, the covariance/Hessian matrices associated with RF and DNN regression are not diagonal (as was the case in Fig.~\ref{fig1}). To build intuition, imagine that only a single feature suffices to explain the label. If there are multiple other features that can similarly explain the label, the model prediction will be shared across these features. Then, pruning will lead to a biased estimate, which can be mitigated by retraining. The following lemma formalizes this intuition under an instructive setup, where the features are perfectly correlated.% and the pruning performance can be exactly characterized.

\cmt{
train$\rightarrow$prune With growing correlations, the This is in contrast to the diagonal covariance with orthogonal features where retraining may hurt. wheredemonstrate a simple scenario which shows that retraining can actually
%  i.e.~when the same observation can be explained by multiple features
 retraining is critical for training sparse neural networks \cite{frankle2019lottery}. In this section, we identify feature correlation as a critical factor that necessitates retraining. This is in contrast to our results in Section \ref{doubdec} which considers independent features.

%Perhaps surprisingly, our results so far show that retraining might hurt the performance since estimating the feature coefficients by fitting the important features alone can perform worse than fitting all features.  

In order to convey the point, let us focus on a simplistic scenario where covariance matrix is rank one and features are perfectly correlated with each other. In this case, the diagonal entries of the covariance matrix essentially dictates the form of the solution. %Suppose the covariance is simply 
}
%Let $(\la_i)_{i=1}^p$ be the diagonal entries of $\sqrt{\bSi}$. Draw $\y,\X$ according to Definition \ref{d model}. 
\begin{lemma} \label{lem rank one}Suppose $\Sc$ is drawn from an LGP$(\sigma,\bSi,\bts)$ as in Def.~\ref{def LGP} where $\text{rank}(\bSi)=1$ with $\bSi=\bla\bla^T$ for $\bla\in\R^p$. Define $\zeta=\Tb_s(\bla)^2/\tn{\bla}^2$. For magnitude and Hessian pruning ($P\in\{M,H\}$) and the associated retraining, we have the following excess risks with respect to $\bt^\st$%population minima
\begin{align}%\underbrace{\frac{\zeta^2\sigma^2}{p-2}}_{\text{Variance Term}}
&\E_{\Sc}[\Lc(\bth_s^P)]-\Lc(\bt^\st)={\frac{\zeta^2\sigma^2}{n-2}}+\underbrace{(1-\zeta)^2(\bla^T\bts)^2}_{\text{Error due to bias}}\\
&\E_{\Sc}[\Lc(\bth_s^{RT})]-\Lc(\bt^\st)={\sigma^2}/({n-2}).
\end{align}
\end{lemma}
The lemma reveals that pruning the model leads to a biased estimator of the label. Specifically, the bias coefficient $1-\zeta$ arises from the missing predictions of the pruned features (which correspond to the small coefficients of $|\bla|$). In contrast, regardless of $s$, retraining always results in an unbiased estimator with the exact same risk as the dense model which quickly decays in sample size $n$. The reason is that retraining enables the remaining features to account for the missing predictions. Here, this is accomplished perfectly, due to the fully correlated nature of the problem. {In particular, this is in contrast to the diagonal covariance (Fig.~\ref{fig1}), where the missing features act like uncorrelated noise during retraining.} \cmt{Our distributional predictions in the next section enable us to capture the benefits of retraining for arbitrary covariances going beyond this instructive example.}% / feature correlations

\section{Main Results}\label{sec main}

Here, we present our main theoretical result: a sharp asymptotic characterization of the distribution of the solution to overparameterized least-squares for correlated designs. We further show how this leads to a sharp prediction of the risk of magnitude-based pruning.  
Concretely, for the rest of this section, we assume the linear Gaussian problem (LGP) of Definition \ref{def LGP}, the overparameterized regime $k=p>n$ and the min-norm model
%\begin{align}
%\hat{\betab} = \left\{ \arg\min_{\betab} \tn{\betab} ~\text{s.t.}~\tn{\y-\X\betab}=\min_{\betab'}\tn{\y-\X\betab'} \right\}\label{eq:min_norm}
%\end{align} 
\begin{align}
\hat{\betab} = \arg\min_{\betab} \tn{\betab} ~\text{s.t.}~\y=\X\bt.\label{eq:min_norm}
\end{align} 
As mentioned in Sec.~\ref{sec:setup}, $\hat{\betab}$ is actually given in closed-form as $\bth=\X^\dagger \y$. Interestingly, our analysis of the distribution of $\bth$ does not rely on the closed-form expression, but rather follows by viewing $\bth$ as the solution to the convex linearly-constrained quadratic program in \eqref{eq:min_norm}. Specifically, our analysis uses the framework of the convex Gaussian min-max Theorem (CGMT) \cite{thrampoulidis2015regularized}, which allows to study rather general inference optimization problems such as the one in \eqref{eq:min_norm}, by relating them with an auxiliary optimization that is simpler to analyze \cite{StoLASSO,oymak2013squared,thrampoulidis2015regularized,thrampoulidis2018precise,salehi2019impact,taheri2020fundamental}.
Due to space considerations, we focus here on the more challenging overparameterized regime and defer the analysis of the underparameterized regime to the SM. 

%Note that $\hat{\betab}$ is the LS solution when $p<n$ and the min-norm interpolator when $p>n$. \so{Due to space considerations, here, we focus on the latter, more challenging, overparameterized regime and defer the underparameterized regime to SM.}
%%Here, we focus on the latter, more challenging, overparameterized regime.

\subsection{Distributional Characterization of the Overparameterized Linear Gaussian Models}

\noindent\emph{Notation:}  We first introduce additional notation necessary to state our theoretical results. $\odot$ denotes the entrywise product of two vectors and $\onebb_p$ is the all ones vector in $\R^p$.
 The \emph{empirical distribution} of a vector $\x\in\R^p$ is given by 
$ \frac{1}{p}\sum_{i=1}^p \delta_{x_i}$, where $\delta_{x_i}$ denotes a Dirac delta mass on $x_i$. Similarly, the empirical joint distribution of vectors $\x, \x'\in \R^p$ is $\frac{1}{p} \sum_{i=1}^p \delta_{(x_i, x'_i)}$. The \emph{Wasserstein-$k$} ($W_k$) distance between two measures $\mu$ and $\nu$ is defined as 
%Given two probability measures $\mu$ (on a space $\mathcal{X}$) and $\nu$ (on a space $\mathcal{Y}$), a coupling $\rho$ of $\mu$ and $\nu$ is a probability distribution on $\mathcal{X}\times\mathcal{Y}$ whose marginals coincide with $\mu$ and $\nu$, respectively. 
%For $k\ge 1$, the \emph{Wasserstein-$2$} ($W_2$) distance between two probability measures $\mu$, $\nu$ on $\R^n$ is defined by
%
%\begin{align}
%
$W_k(\mu,\nu) \equiv \left(\inf_{\rho}\E_{(X,Y)\sim \rho}|X-Y|^k\right)^{1/k},$ %\label{eq:WassersteinDef}
%
%\end{align}
%
where the infimum is over all the couplings of $\mu$ and $\nu$, i.e. all random variables $(X,Y)$ such that $X\sim\mu$ and $Y\sim\nu$ marginally.
%X ? µ and Y ? ? marginally
A sequence of probability distributions $\nu_p$ on $\R^m$ {converges in $W_k$} to 
$\nu$, written $\nu_p\stackrel{W_k}{\Longrightarrow} \nu$, if $W_k(\nu_p,\nu) \rightarrow 0$ as $p \rightarrow \infty$. Finally, we say that a function $f:\R^m\rightarrow\R$ is pseudo-Lipschitz of order $k$, denoted $f\in\rm{PL}(k)$, if there is a constant $L>0$ such that for all $\x,\y\in\R^m$,
$
|f(\x) - f(\y)|\leq L(1+\tn{\x}^{k-1}+\tn{\y}^{k-1})\|\x-\y\|_2.
$
\fy{We call $L$ the $\rm{PL}(k)$ constant of $f$.}
%\somm{Is this convergence in probability or almost sure?} \CT{I understand that here convergence is a.s.. All our results are in probability / wpa 1. So formally, speaking, we do not show $W_k$ convergence. But is ok because the theorem is stated in terms of convergence of test functions.} 
An equivalent definition of $W_k$ convergence is that, for any $f\in\rm{PL}(k)$, $\lim_{p\rightarrow\infty}\E f(X_p)=\E f(X)$, where expectation is with respect to $X_p\sim\nu_p$ and $X\sim\nu$. For a sequence of random variables $\mathcal{X}_{p}$ that converge in probability to some constant $c$ in the limit of Assumption \ref{ass:linear} below, we  write $\mathcal{X}_{p}\rP c$. 

%We say that a sequence of probability distributions $\nu_p$ on $\R^m$ {converges in $W_k$} to $\nu$, written $\nu_p\stackrel{W_k}{\Longrightarrow} \nu$, if $W_k(\nu_p,\nu) \rightarrow 0$ as $p \rightarrow \infty$. An equivalent definition is that, for any $f\in\rm{PL}(k)$, $\lim_{p\rightarrow\infty}\E f(X_p)=\E f(X)$, where expectation is with respect to $X_p\sim\nu_p$ and $X\sim\nu$ (e.g., \cite{montanari2017estimation}).

\vp
Next, we formalize the set of assumption under which our analysis applies. Our asymptotic results hold in the linear asymptotic regime specified below.
\begin{assumption}\label{ass:linear} We focus on a double asymptotic regime where $n,p,s\rightarrow\infty$ at fixed overparameterization ratio $\kappa:=p/n>0$ and sparsity level $\alpha:=s/p\in(0,1)$.  
\end{assumption}

Additionally, we require certain mild assumptions on the behavior of the covariance matrix $\Sigmab$ and of the true latent vector $\bt^\star$. For simplicity, we assume here that $\Sigmab = \diag{[\Sigmab_{1,1},\ldots,\Sigmab_{p,p}]}$.% and we also need that $\Sigmab_{i,i}$ are non-zero and bounded. 
%Moreover, we specify the following two assumptions.

\begin{restatable}{assumption}{asstwo}\label{ass:inv} {The covariance matrix $\Sigmab$ is diagonal} and there exist constants $\Sigma_{\min},\Sigma_{\max}\in(0,\infty)$ such that:
$
\Sigma_{\min}\leq\bSi_{i,i}\leq \Sigma_{\max},
$
for all $i\in[p].$
\end{restatable}

%Finally, we assume a limit on the joint empirical measure of $(\diag{\Sigma},\betas)$ as follows.
\begin{restatable}{assumption}{assthree}\label{ass:mu}
% Let $\nub_i:= {{p}(\bt^\star_i)^2{\lai}},~i\in[p]$. The joint empirical distribution of $\{(\lai,\nub_i)\}_{i\in[p]}$ converges in Wasserstein-2 distance to a probability distribution $\mu$ on $\R_{>0}\times\R_{>0}$, i.e.,
%$
%\frac{1}{p}\sum_{i\in[p]}\delta_{(\lai,\nub_i)} \stackrel{W_2}{\Longrightarrow} \mu.
%$
The joint empirical distribution of $\{(\lai,\sqrt{p}\betas_i)\}_{i\in[p]}$ converges in {Wasserstein-k} distance to a probability distribution $\mu$ on $\R_{>0}\times\R$ {for some {$k\geq 4$}}. That is
$
\frac{1}{p}\sum_{i\in[p]}\delta_{(\lai,\sqrt{p}\betas_i)} \stackrel{W_k}{\Longrightarrow} \mu.
$
\end{restatable}

With these, we are ready to define, what will turn out to be, the asymptotic DC in the overparameterized regime.
\begin{definition}[Asymptotic DC -- Overparameterized regime]\label{def:Xi}
Let random variables $(\Lambda,B)\sim \mu$ (where $\mu$ is defined in Assumption \ref{ass:mu}) and fix $\kappa>1$. Define parameter $\xi$ as the unique positive solution to the following equation
\begin{align}\label{eq:ksi}
\E_{\mu}\Big[ \big({1+(\xi\cdot\Lambda)^{-1}}\big)^{-1} \Big] = {\kappa^{-1}}\,.
\end{align}
Further define the positive parameter $\gamma$ as follows:
\begin{align}
\label{eq:gamma}
\hspace{-0.1in}\gamma := 
%\frac{\sigma^2 + \E_{\mu}\left[\frac{N^2}{(\Lambda^{-1}+\xi)^2}\right]}{1-\kappa\E_{\mu}\left[\frac{1}{\left(1+(\xi\Lambda)^{-1}\right)^2}\right]}  
\Big({\sigma^2 + \E_{\mu}\Big[\frac{B^2\Lambda}{(1+\xi\Lambda)^2}\Big]}\Big)\Big/\Big({1-\kappa\E_{\mu}\Big[\frac{1}{\left(1+(\xi\Lambda)^{-1}\right)^2}\Big]}\Big).
\end{align}
%\ct{@Samet: In your notation $\gamma$ should be $\Gamma\kappa$. Everything matches except the second term in the numerator. In your notation, that second term becomes $\int_0^1\bz^2(t)\bt^2(t) \Sigmab(t) dt$ (note the extra $\Sigmab(t)$ compared to your formula)
%}
With these and $H\sim\Nn(0,1)$, define the random variable
\begin{align}
X_{\kappa,\sigma^2}(\Lambda,B,H) := \Big(1-\frac{1}{1+ \xi\Lambda}\Big) B + \sqrt{\kappa}\frac{\sqrt{\gamma}\,\Lambda^{-1/2}}{1+(\xi\Lambda)^{-1}} H, \label{eq:X}
\end{align}
and let $\Pi_{\kappa,\sigma^2}$ be its distribution.
\end{definition}

\fy{Our main result establishes asymptotic convergence of the empirical distribution of $(\sqrt{p}\bth,\sqrt{p}\betas,\Sigmab)$ for a rich class of test functions. These are the functions within $\rm{PL}(3)$ that become $\rm{PL}(2)$ when restricted to the first two indices. Formally, we define this class of functions as follows
\begin{align}\label{eq:pdef}
\Plt :=\{&f:\R^2\times \Zc\rightarrow\R,~f\in \rm{PL}(3)~\text{and}~\\
&\sup_{z\in\Zc}\text{``}\rm{PL}(2)~\text{constant~of}~f(\cdot,\cdot,z)\text{''}<\infty\}.\nn
\end{align}
For pruning analysis, we set $\Zc=[\Sigma_{\min},\Sigma_{\max}]$ and define %the following functions%class are critical %has the form%It applies to all functions in $\rm{PL}(2)$ as well as the following class of functions
\begin{align}\label{eq:fdef}
\Fl:= \{ f:\R^2\times \Zc\rightarrow\R~\big|~&f(x,y,z) = z(y-g(x))^2\nn\\
&~\text{where}~g(\cdot)~\text{is Lipschitz}\}.
%f_\Lc(x,y,z) := \fx{z(y-g(x))}^2. %:\R\rightarrow \R
\end{align}
%where $g$ is Lipschitz. 
%\CT{I rewrote this as a family. Alternatively, Thm. 1 could be stated to hold for family of functions $f:\R^3\rightarrow\R$ such that $f\in\rm{PL}(3)$ and $g(x,y):=f(x,y,z)$ is $\rm{PL}(2)$ with some $\rm{PL}$-constant $L$ that uniformly over $z$ in the range of $f$.}
As discussed below, $\Fl$ is important for predicting the risk of the (pruned) model. In the SM, we prove that $\Fl\subset \Plt$. We are now ready to state our main theoretical result.}% which applies to $\Plt$
%\begin{theorem}[Asymptotic DC for LGP]\label{thm:master_W2} %Fix $\kappa=p/n>1$.
%Let Assumption \ref{ass:linear} hold with $\kappa>1$ and  further let Assumptions \ref{ass:inv} and \ref{ass:mu} hold. Consider $\hat{\bt}$ as in \eqref{eq:min_norm} and $\hat\Pi_n(\y,\X,\betas,\Sigmab):=\frac{1}{p}\sum_{i=1}^{p}\delta_{\sqrt{p}\hat{\bt}_i,\sqrt{p}\betas_i,\Sigmab_{i,i}}$, the joint empirical distribution of $(\sqrt{p}\hat\betab,\sqrt{p}\betas,\Sigmab)$. Recall the definition of the measure $\Pi_{\kappa,\sigma^2}$ in Definition \ref{def:Xi}. Then, $\hat\Pi_n(\y,\X,\betas,\Sigmab)$ converges in Wasserstein-k distance to $\Pi_{\kappa,\sigma^2}\otimes\mu$.
%Specifically, for any function $f:\R^3\rightarrow\R$, $f\in\rm{PL}(k)$ with $k\geq 3$, it holds that
%\begin{align}\label{eq:thm}
%\hspace{-0.1in}p^{-1} \sum_{i=1}^{p} f(\sqrt{p}\hat\betab_i,\sqrt{p}\betas_i,\Sigmab_{i,i}) \rP \E\left[f(X_{\kappa,\sigma^2},B,\Lambda) \right],
%\end{align}
%where the expectation is over $(\Lambda,B,H)\sim\mu\otimes\Nn(0,1).$
%\end{theorem}
% be any of the following two: (a) $f\in\rm{PL}(2)$, or, (b) $f=f_\Lc$ defined in \eqref{eq:fdef} \fx{where $g$ is a Lipschitz function}. 
\cts{
\begin{restatable}[Asymptotic DC -- Overparameterized LGP]{theorem}{mainthm}\label{thm:master_W2} %Fix $\kappa=p/n>1$.
Fix $\kappa>1$ and suppose Assumptions \ref{ass:inv} and \ref{ass:mu} hold. Recall the solution $\bth$ from \eqref{eq:min_norm} and let 
\[
\hat\Pi_n(\y,\X,\betas,\Sigmab):=\frac{1}{p}\sum_{i=1}^{p}\delta_{(\sqrt{p}\bth_i,\sqrt{p}\betas_i,\Sigmab_{i,i})}
\] be the joint empirical distribution of $(\sqrt{p}\bth,\sqrt{p}\betas,\Sigmab)$. Let $f:\R^3\rightarrow\R$ be a function in $\Plt$ defined in \eqref{eq:pdef}. We have that
%\begin{align}\label{eq:thm_app_f}
%\frac{1}{p} \sum_{i=1}^{p} f(\sqrt{p}\bth_i,\sqrt{p}\betas_i,\Sigmab_{i,i}) \rP \E_{(\Lambda,B,H)\sim\mu\otimes\Nn(0,1)}\left[f(X_{\kappa,\sigma^2},B,\Lambda) \right].
%\end{align}
\begin{align}\label{eq:thm}
\hspace{-0.1in}\frac{1}{p} \sum_{i=1}^{p} f(\sqrt{p}\hat\betab_i,\sqrt{p}\betas_i,\Sigmab_{i,i}) \rP \E\left[f(X_{\kappa,\sigma^2},B,\Lambda) \right].
\end{align}
\end{restatable}
}

%\begin{restatable}[Goldbach's conjecture]{theorem}{goldbach}
%\label{thm:goldbach}
%Every even integer greater than 2 can be expressed as the sum of two primes.
%\end{restatable}
%
%\goldbach*

%%%%%%%%%%%%%%%%%%%%%%%%%%%%%%%%%%%%%%%%%%%%%
%Note that the integer $k$ appearing in the statement of the Theorem \ref{thm:master_W2} is the same 
%\begin{remark}(Novelty) Compare to Surprises. Compare to CGMT.
%\end{remark}

As advertised, Theorem \ref{thm:master_W2} fully characterizes the joint empirical distribution of the min-norm solution, the latent vector and the covariance spectrum. The asymptotic DC allows us to precisely characterize several quantities of interest, such as estimation error, generalization error etc.. For example, a direct application of \eqref{eq:thm} to the function $f(x,y,z)=z(y-x)^2\in \Fl\subset\Plt$ directly yields the risk prediction of the min-norm solution recovering \cite[Thm.~3]{hastie2019surprises} as a special case. Later in this section, we show how to use Theorem \ref{thm:master_W2} towards the more challenging task of precisely characterizing the risk of magnitude-based pruning.

Before that, let us quickly remark on the technical novelty of the theorem. Prior work has mostly applied the CGMT to isotropic features. Out of these, only very few obtain DC, \cite{thrampoulidis2018symbol,miolane2018distribution}, while the majority focuses on simpler metrics, such as squared-error. Instead, Theorem \ref{thm:master_W2} considers correlated designs and the overparameterized regime. The most closely related work in that respect is \cite{montanari2019generalization}, which very recently obtained the DC of the max-margin classifier. Similar to us, they use the CGMT, but their analysis of the auxiliary optimization is technically different to ours. Our approach is similar to \cite{thrampoulidis2018symbol}, but extra technical effort is needed to account for correlated designs and the overparameterized regime.

\subsection{From DC to Risk Characterization}\label{sec:risk}
%In the remaining of this section, we show how to use Theorem \ref{thm:master_W2} to precisely characterize the risk of the magnitude-based pruning. 
First, we consider a simpler ``threshold-based" pruning method that applies a fixed threshold at every entry of $\hat\betab$. Next, we relate this to magnitude-based pruning and obtain a characterization for the performance of the latter.
In order to define the threshold-based pruning vector, let \[
\Tc_t(x) = \begin{cases} x & \text{if } |x|>t \\ 0 & \text{otherwise} \end{cases},
\] be the hard-thresholding function with fixed threshold $t\in\R_+$. Define $\bth^\Tc_{t} := \Tc_{t/\sqrt{p}}(\hat\bt)$, where $\Tc_t$ acts component-wise. Then, the population risk of   $\bth^\Tc_t$ becomes
\begin{align}
\Lc(\bth^\Tc_t) &= \E_{\Dc}[(\x^T(\bt^\star-\bth^\Tc_t) + \sigma z)^2] \nn\\
 \cmt{&= \sigma^2 + (\bt^\star-\bth^\Tc_t)^T\Sigmab(\bt^\star-\bth^\Tc_t) \nn \\}
  &= \sigma^2 + \frac{1}{p}\sum_{i=1}^p\Sigmab_{i,i}\big(\sqrt{p}\betas_i-\Tc_{t}(\sqrt{p}\hat\betab_i)\big)^2\nn\\
  &\rP \sigma^2 + \E\left[ \Lambda\left(B-\Tc_t(X_{\kappa,\sigma^2})\right)^2 \right]\,.
  \label{eq:loss_w2}
%\\ &= \sigma^2 + \tn{\Tc_t\left(\sqrt{\Sigmab}(\hat\bt-\bt^\star)\right)}^2 + \tn{\sqrt{\Sigmab}(\bt^\star-\Tc_t(\bt^\star))}^2\nn
%\\ &= \sigma^2 + \frac{1}{p}\tn{\Tc_t(\sqrt{\p}\hat\w)}^2 + \tn{\sqrt{\Sigmab}(\bt^\star-\Tc_t(\bt^\star))}^2 \label{eq:loss_w2}
\end{align}
%In the first equality, we let $z\sim\Nn(0,1)$. 
%In the second, we used $\E[\x\x^T]=\Sigmab$. In the third line, we wrote $\bt^\star = \Tc_t(\bt^\star) + (\bt^\star - \Tc_t(\bt^\star))$ and we also used the fact that $\Sigmab$ is diagonal so that $\sqrt{\Sigmab}\Tc_t(\vb)=\Tc_t(\sqrt{\Sigmab}\vb)$ for any vector $\vb\in\R^p$. In the last line, we recalled the definition of $\hat\w$ in Theorem \ref{thm:master_W2}. 
In the second line above, we note that $\sqrt{p}\Tc_{t'}(x)=\Tc_{\sqrt{p}t'}(\sqrt{p}x)$. In the last line, we apply \eqref{eq:thm}, after recognizing that the function $(x,y,z)\mapsto z(y-\Tc_t(x))^2$ 
%is pseudo-Lipschitz of order $3$ (see SM \fx{Lemma \ref{lem:fL}} for a proof)%\footnote{Notice that the function $\Tc_t$ is discontinuous. However, it can be %approximated arbitrarily closely with Lipschitz functions.}. \fx{Additionally observe %that this function 
is a member of the $\Fl$ family defined in \eqref{eq:fdef}.
%} 
As in \eqref{eq:thm}, the expectation here is with respect to $(\Lambda,B,H)\sim\mu\otimes\Nn(0,1)$. 
%\somm{Here, more justification on why Lip approx is sufficient would be great. Obviously Lip approx should be fine because asymptotic distribution is continuous.}
%\so{$\E[\Tc_t(X_{\kappa,\sigma^2}(\Lambda,N,H))^2]$. 
%The third term converges to }\som{Forgot covariance multiplier?}

Now, we show how to use \eqref{eq:loss_w2} and Theorem \ref{thm:master_W2} to characterize the risk of the magnitude-based pruned vector $\betab^M_s:=\Tb_s(\bth)$. Recall, here from Assumption \ref{ass:linear} that $s=\alpha p$. To relate $\hat{\betab}^M_s$ to $\bth^\Tc_t$, consider the set  $\Sc_t:=\{i\in[p]\,:\,\sqrt{p}|\hat\betab_i| \geq t \}$ for some constant $t\in\R_+$ (not scaling with $n,p,s$). Note that the ratio ${|\Sc_t|}/{p}$ is equal to
\begin{align}
p^{-1}\sum_{i=1}^p\mathbb{1}_{[\sqrt{p}|\hat\betab_i| \geq t]} \rP \E[\mathbb{1}_{[|X_{\kappa,\sigma^2}|\geq t]}] = \P\left(|X_{\kappa,\sigma^2}|\geq t\right).\nn
\end{align}
Here, $\mathbb{1}$ denotes the indicator function and the convergence follows from Theorem \ref{thm:master_W2} when applied to a sequence of bounded Lipschitz functions approximating the indicator. Thus, by choosing 
\begin{align}\label{eq:tstar}
t^\star:=\sup\left\{t\in\R\,:\, \P(|X_{\kappa,\sigma^2}|\geq t) \geq \alpha \right\},
\end{align}
it holds that ${|\Sc_t|}/{p}\rP\alpha$. In words, and {observing that $X_{\kappa,\sigma^2}$ admits a continuous density (due to the Gaussian variable $H$)}: for any  $\eps>0$, in the limit of $n,p,s\rightarrow\infty$, the vector $\bth^\Tc_{t^\star}$ has $(1\pm\eps) \alpha p= (1\pm\eps) s$ non-zero entries, which correspond to the largest magnitude entries of $\bth$, with probability approaching1. Since this holds for arbitrarily small $\eps>0$, recalling $t^\star$ as in \eqref{eq:tstar}, we can conclude from \eqref{eq:loss_w2} that the risk of the magnitude-pruned model converges as follows.
{\begin{restatable}[Risk of Magnitude-pruning]{corollary}{cormag}\label{cor:mag}
Let the same assumptions and notation as in the statement of Theorem \ref{thm:master_W2} hold. Specifically, let $\hat\betab$ be the min-norm solution in \eqref{eq:min_norm} and $\bth_s^M:=\Tb_s(\bt)$ the magnitude-pruned model at sparsity $s$. Recall the threshold $t^\st$ from \eqref{eq:tstar}. The risk of $\bth^M_s$ satisfies the following in the limit of $n,p,s\rightarrow\infty$ at rates $\kappa:=p/n>1$ and $\alpha:=s/p\in(0,1)$ (cf. Assumption \ref{ass:linear}):
\begin{align}
\Lc(\bth^M_s) \rP \sigma^2 + \E\left[ \Lambda\left(B-\Tc_{t^\star}(X_{\kappa,\sigma^2})\right)^2 \right],\nn
\end{align}
where the expectation is over $(\Lambda,B,H)\sim\mu\otimes\Nn(0,1).$
\end{restatable}}
%\begin{align}
%\Lc(\bth^M_s) \rP \sigma^2 + \E\left[ \Lambda\left(B-\Tc_{t^\star}(X_{\kappa,\sigma^2})\right) \right].\nn
%\end{align}
%\somm{One way to fully rigorize this line and Lip approx is showing that any eps-fraction of entries of $||\hat{\beta}||$ is small in $\eps$. With that, it becomes crystal clear I think.}
%where .

\cmt{LET US NOT FORGET TO ADD OR COMMENT ABOUT UNDERPARAM!}
% Prediction of
\subsection{Non-asymptotic DC and Retraining Formula}
While Theorem \ref{thm:master_W2} is stated in the asymptotic regime, during analysis, the DC arises in a non-asymptotic fashion. The following definition is the non-asymptotic counterpart of Def.~\ref{def:Xi}. We remark that this definition applies to arbitrary covariance (not necessarily diagonal) by applying a simple eigen-rotation before and after the DC formula associated with the diagonalized covariance.
\cmt{
\begin{definition}[Non-asymptotic DC] \label{aux_def}Fix $p>n\geq 1$ and set $\kappa=p/n>1$. Given $\sigma>0$, covariance $\bSi=\Ub\text{diag}(\bla)\Ub^T$ and latent vector $\bt$, set $\bar{\bt}=\Ub^T\bt$ and define the unique non-negative terms $\Xi,\Gamma,\bz\in\R^p$ and $\bg\in\R^p$ as follows\vspace{-0pt}
\begin{align}
&\Xi>0\quad\text{is the solution of}\quad 1=\frac{\kappa}{p}\sum_{i=1}^p\frac{1}{1+(\Xi\bla_i)^{-1}},\nn\\
&\Gamma=\frac{\sigma^2+\sum_{i=1}^p\bla_i\bz_i^2\bar{\bt}_i^2}{\kappa(1-\frac{\kappa}{p}\sum_{i=1}^p{(1+(\Xi\bla_i)^{-1})^{-2}})},\nn\\
&\bz_i=\frac{1}{1+\Xi\bla_i}\quad\text{and}\quad \bg_i=\frac{\kappa\sqrt{\Gamma}}{1+(\Xi\bla_i)^{-1}}\quad\text{for}\quad 1\leq i\leq p.\nn
\end{align}
The non-asymptotic OP-LS distribution is defined as the following $\Ub$-rotated normal distribution
\[
\Dc_{\sigma,\bSi,\bt}=\Ub\Nn((\onebb_p-\bz)\odot\bar{\bt},p^{-1}\text{diag}(\bla^{-1}\odot\bg^2)).
\]
\end{definition}}
%\somm{Justify this definition in more detail}
\begin{definition}[Non-asymptotic DC] \label{aux_def}Fix $p>n\geq 1$ and set $\kappa=p/n>1$. Given $\sigma>0$, covariance $\bSi=\Ub\text{diag}(\bla)\Ub^T$ and latent vector $\bt$, set $\bar{\bt}=\Ub^T\bt$ and define the unique non-negative terms $\xi,\gamma,\bz\in\R^p$ and $\bg\in\R^p$ as follows:\vspace{-0pt}
\begin{align}
&\xi>0\quad\text{is the solution of}\quad \kappa^{-1}={p^{-1}}\sum_{i=1}^p\big({1+(\xi\bla_i)^{-1}}\big)^{-1},\nn\\
&\gamma=\frac{\sigma^2+\sum_{i=1}^p\bla_i\bz_i^2\bar{\bt}_i^2}{1-\frac{\kappa}{p}\sum_{i=1}^p{(1+(\xi\bla_i)^{-1})^{-2}}},\nn\\
&\bz_i=({1+\xi\bla_i})^{-1}\quad\text{,}\quad \bg_i={\kappa\gamma}(1+(\xi\bla_i)^{-1})^{-2},~ 1\leq i\leq p.\nn
\end{align}
The non-asymptotic distributional prediction is given by the following $\Ub$-rotated normal distribution
\[
\Dc_{\sigma,\bSi,\bt}=\Ub\Nn((\onebb_p-\bz)\odot\bar{\bt},p^{-1}\text{diag}(\bla^{-1}\odot\bg)).
\]
\end{definition}
{We remark that this definition is similar in  spirit to the concurrent/recent work \cite{li2020exploring}. However, unlike this work, here we prove the asymptotic correctness of the DC, we use it to rigorously predict the pruning performance and also extend this to retraining DC as discussed next.} 

%E[x_Ix_I^T]^(-1)E[xx^T]E[xx^T]^(-1)E[yx] = E[x_Ix_I^T]^(-1)E[yx] = E[x_Ix_I^T]^(-1)E[yx_I]
\noindent \textbf{Retraining DC.} As the next step, we would like to characterize the DC of the solution after retraining, i.e.,~$\bth^{RT}$. We carry out the retraining derivation (for magnitude pruning) as follows. Let $\Ic\subset[p]$ be the nonzero support of the pruned vector $\bth^M_s$. Re-solving \eqref{eq:ERM} restricted to the features over $\Ic$ corresponds to a linear problem with effective feature covariance $\bSi_\Ic$ with support of non-zeros restricted to $\Ic\times \Ic$. For this feature covariance, we can also calculate the effective noise level and global minima of the population risk $\bt^\st_\Ic$. The latter has the closed-form solution $\bt^\st_\Ic=\bSi_\Ic^\dagger\bSi\bt^\st$. The effective noise is given by accounting for the risk change due to the missing features via $\sigma_\Ic=(\sigma^2+{\bt^\st}^T\bSi\bt^\st-{\bt^\st_\Ic}^T\bSi_\Ic\bt^\st_\Ic)^{1/2}$. With these terms in place, fixing $\Ic$ and using Def.~\ref{aux_def}, the retraining prediction becomes $\Dc_{\sigma_{\Ic},\bSi_{\Ic},\bts_\Ic}$. This process is summarized below.
\begin{definition}[Retraining DC] \label{RT_def}Consider the setting of Def.~\ref{aux_def} with $\sigma,\bSi,\bt^\st$ and sparsity target $s$. The sample $\bth^{RT}$ from the retraining distribution $\Dc^{\text{RT,s}}_{\sigma,\bSi,\bt^\st}$ is constructed as follows. Sample $\bth\sim \Dc_{\sigma,\bSi,\bt^\st}$ and compute the set of the top-$s$ indices $\Ic=\Ic(\Tb_s(\bth))$. Given $\Ic$, obtain the effective covariance $\bSi_\Ic\in\R^{p\times p}$, population minima $\bt^\st_\Ic\in\R^p$, and the noise level $\sigma_{\Ic}>0$ as described above. Draw $\bth^{RT}\sim\Dc_{\sigma_{\Ic},\bSi_{\Ic},\bts_\Ic}$.

\cmt{Obtain $\bSi_\Ic\in\R^{p\times p}$ by restricting the nonzero support of $\bSi$ to $\Ic\times \Ic$. Set $\Ic$ restricted population minima $\bt^\st_\Ic=\bSi_\Ic^\dagger\bSi\bt^\st$ and set the new noise level $\sigma_{\Ic}=(\sigma^2+{\bt^\st}^T\bSi\bt^\st-{\bt^\st_\Ic}^T\bSi_\Ic\bt^\st_\Ic)^{1/2}$. Draw $\bth^{RT}\sim\Dc_{\sigma_{\Ic},\bSi_{\Ic},\bt_\Ic}$. }
\end{definition}
%\som{Clarify here!}
Observe that, the support $\Ic$ depends on the samples $\Sc$ via $\bth$. Thus, our retraining DC is actually derived for the scenario when the retraining phase uses a fresh set of $n$ samples to break the dependence between $\Ic,\Sc$ (which obtains $\bth^{RT}_{\text{fresh}}$). Despite this, we empirically observe that the retraining DC predicts the regular retraining (reusing $\Sc$) performance remarkably well and perfectly predicts $\bth^{RT}_{\text{fresh}}$ as discussed in Figs.~\ref{figRF} and \ref{figRF2}. Finally, we defer the formalization of the retraining analysis to a future work. This includes proving that $\bth^{RT}_{\text{fresh}}$ obeys Def.~\ref{RT_def} asymptotically as well as directly studying $\bth^{RT}$ by capturing the impact of the $\Ic,\Sc$ dependency.
 
%\so{Here, the core technical challenges are addressing the randomness in $\Ic$ by characterizing its asymptotic behavior and properly mapping this behavior to the asymptotic retraining distribution.}

 %Additionally, if we indeed use fresh samples, it indeed exactly predicts the retraining performance (see Fig. \ref{figRF2}). 

%showing that asymptotically all $\Ic$'s will lead to the same retraining distribution
%theshowing that asymptotically all $\Ic$'s will lead to the same retraining distribution.

\section{Conclusions and Future Directions}\label{sec discuss}
This paper sheds light on under-explored phenomena in pruning practices for neural network model compression. On a theoretical level, we prove an accurate distributional characterization (DC) for the solution of overparameterized least-squares for linear models with correlated Gaussian features. Our DC allows to precisely characterize the pruning performance of popular pruning methods, such as magnitude pruning. Importantly, our DC combined with a linear Gaussian equivalence, leads to precise analytic formulas for the pruning performance of nonlinear random feature models. On the experimental side, we provide a thorough study of overparameterization and pruning with experiments on linear models, random features and neural nets with growing complexity. Our experiments reveal striking phenomena such as a novel double descent behavior for model pruning and the power of overparameterization. They also shed light on common practices such as retraining after pruning. 
% \cts{Finish}

Going forward, there are several exciting directions to pursue. First, it would be insightful to study whether same phenomena occur for other loss functions in particular for cross-entropy. Second, this work focuses on unregularized regression tasks and it is important to identify optimal regularization schemes for pruning purposes. For instance, should we use classical $\ell_1/\ell_2$ regularization or can we refine them by injecting problem priors such as covariance information? Finally, going beyond pruning, using DC, one can investigate other compression techniques that process the output of the initial overparameterized learning problem, such as model quantization and distillation. 

\section*{Acknowledgments}
S. Oymak is partially supported by the NSF award CNS-1932254. C. Thrampoulidis is partially supported by the NSF under Grant Numbers CCF-2009030. 

%\newpage
\section*{Potential Ethical Impacts}
While deep learning is transformative in wide swath of applications, it comes at a cost: State-of-the-art deep learning models tend to be very large and consume significant energy during inference. The race for larger and better models and growing list of applications exacerbates this carbon footprint problem. Thus there is an urgent need for better and more principled model compression methods to help build environmentally friendly ML models. This work responds to this need by establishing the fundamental algorithmic principles and guarantees behind the contemporary model compression algorithms and by shedding light on the design of lightweight energy- and compute-efficient neural networks. We do not see an ethical concern associated with this work.
%We anticipate that proposed theories and associated novel insights will also guide .

%\input{sparse_main}

\cmt{\subsection{Random Features}}
% \begin{proof}
% Let $\F\in\R^{s\times p}$ be the feature matrix. Set $\bSi'=\F\bSi\F^T$. The least-squares solution has the form $\tbh=\theta\F\vb$
% \[
% \theta
% %=\arg\min_{\theta}\tn{\y-\g\vb^T\F\tbh}\\
% =\arg\min_{\theta}\tn{\vb^T\bt-(\tn{\F\vb}^2\theta)\g+\sigma\z}\implies\theta=\frac{1}{\tn{\F\vb}^2}(\vb^T\bt+\frac{\sigma}{\tn{\g}^2}\g^T\z)
% \]
% \[
% %\Lc(\tbh)\simeq\sigma^2(n-1)
% \]
% \end{proof}

%\newpage
%\bibliographystyle{aaai21}
\bibliography{Bibfiles2}

%\newpage
\appendix
\onecolumn

\section*{Organization of the Supplementary Material}
The supplementary material (SM) 
%starts after references (at Page 11) and 
is organized as follows. 
\begin{enumerate}
\item In Section \ref{SM mot} we provide additional experiments on neural networks and linear models further supporting the main results. 
\item In Section \ref{sec proof thm 1} we prove our main Theorem \ref{thm:master_W2}.
\item Section \ref{sec prune risk} provides our analysis and results on magnitude and Hessian pruning.
\item Section \ref{SM useful fact} provides further supporting technical results used in our proofs.
\item Section \ref{SM overdet} provides our asymptotic analysis of overdetermined problems complementing our main results on overparameterized problems. 
%\item Section \ref{SM cgmt res} provides supporting results on CGMT analysis. 
\item In Section \ref{SM lem proof} we prove Lemma \ref{lem rank one}. 
\end{enumerate}
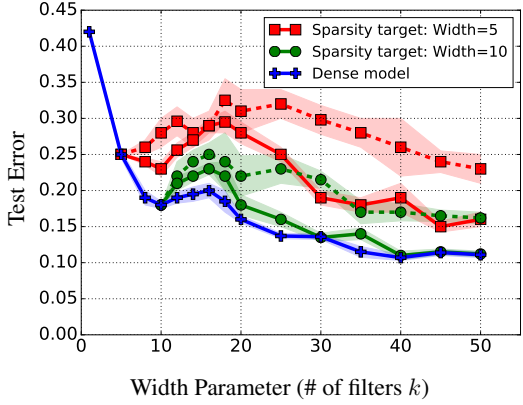
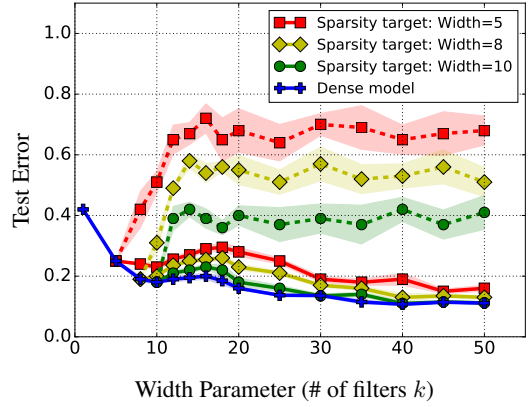
\begin{figure}[t!]
        \centering
	\begin{subfigure}{3.5in}\vspace{-5pt}
		\centering
		\begin{tikzpicture}
		\node at (0,0) {\includegraphics[scale=0.33]{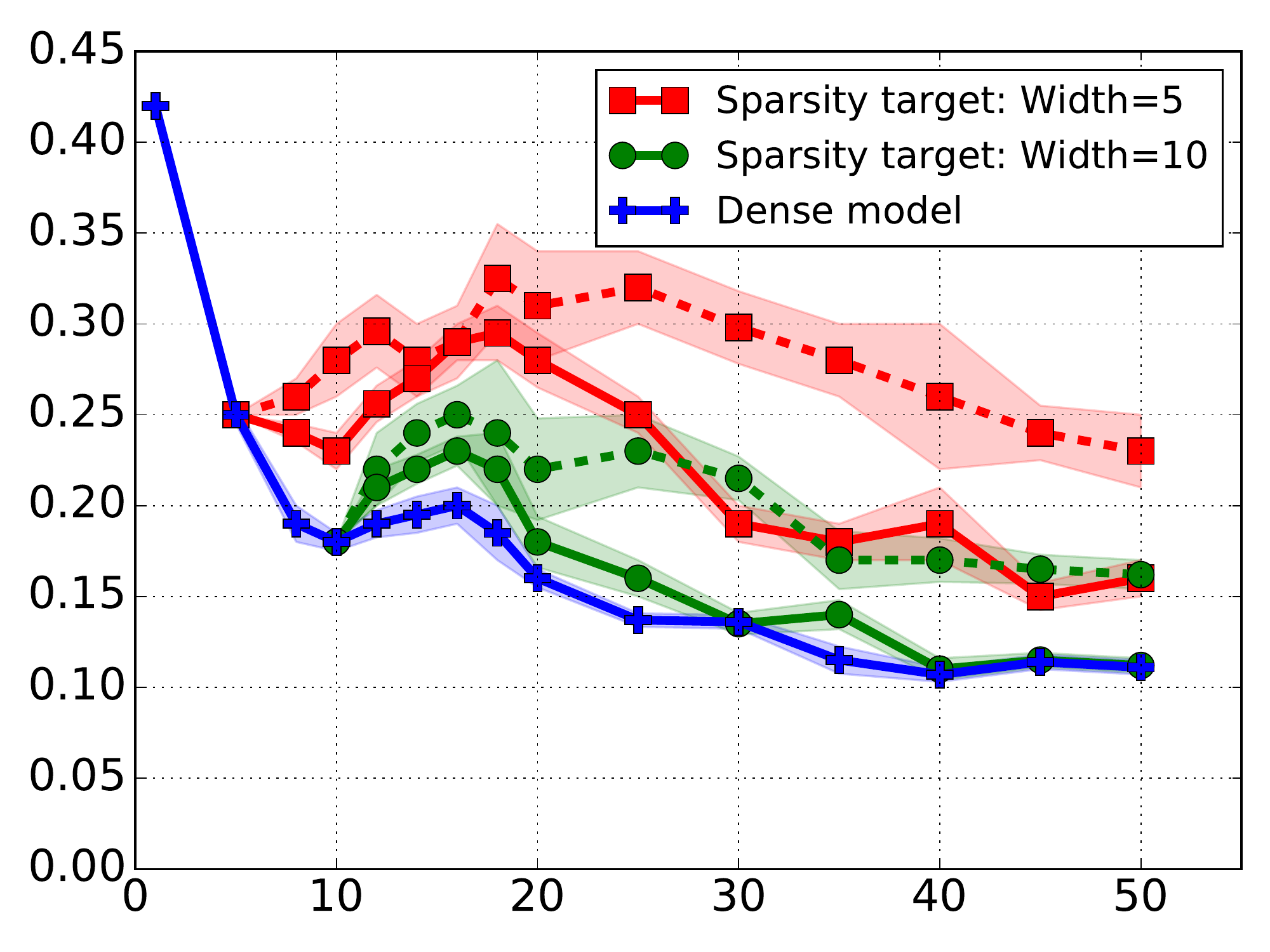}};
		\node at (-3.5,0) [rotate=90,scale=.9]{Test Error};
		\node at (0,-2.8) [scale=.9]{Width Parameter (\# of filters $k$)};%Each bar in the figure shows the correlations' geometric mean of datasets in this group when training to corresponding non-zero level. 
		\end{tikzpicture}\caption{\small{Magnitude-based and randomly pruned models.}}\label{figCFa}\vspace{-0.2cm}
	\end{subfigure}~~~~~\begin{subfigure}{3.5in}\vspace{-5pt}
	\centering
	\begin{tikzpicture}
	\node at (0,0) {\includegraphics[scale=0.33]{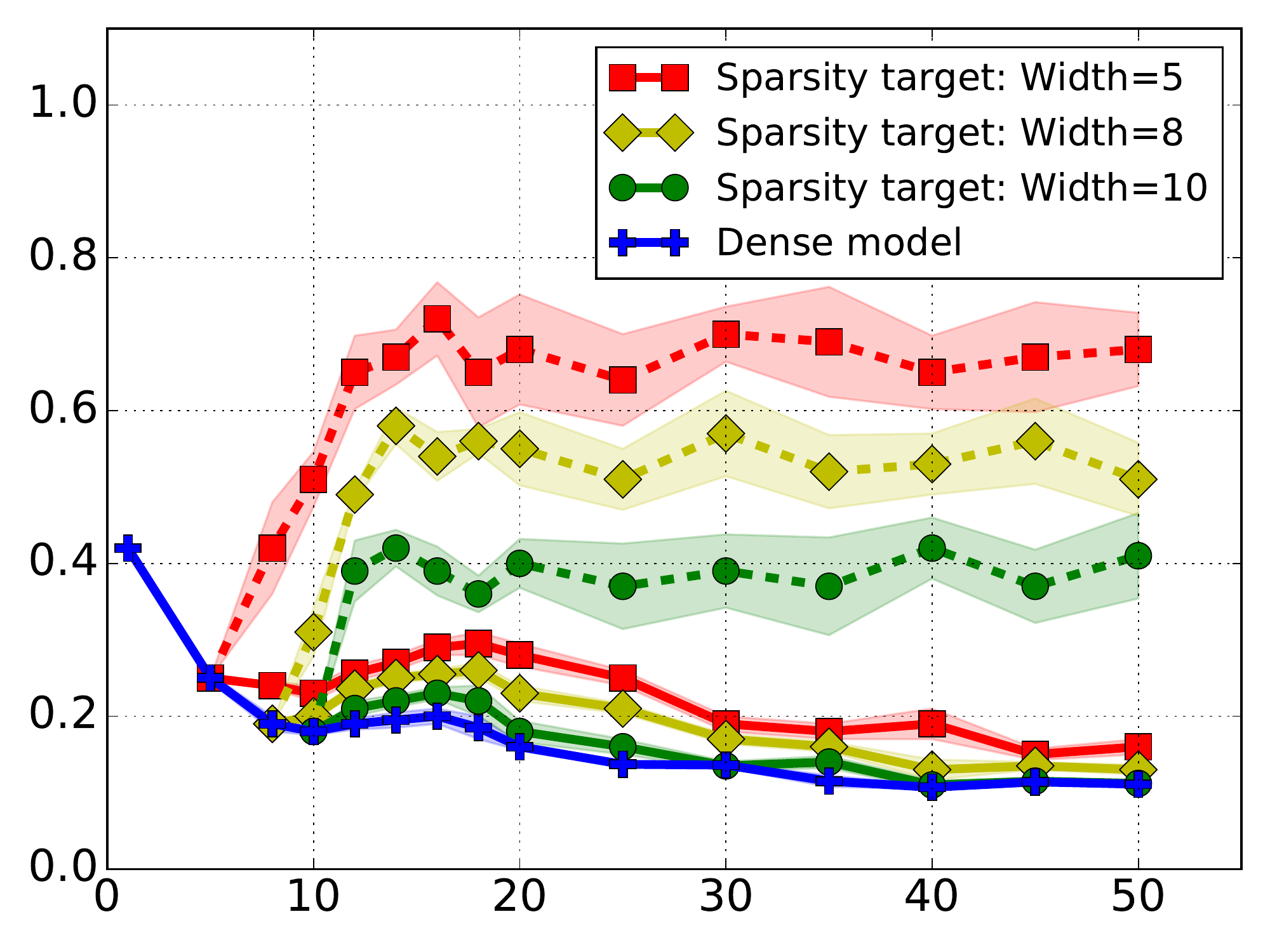}};
	%\node at (-2.4,0) [rotate=90,scale=1.]{Test Risk};
		\node at (-3.5,0) [rotate=90,scale=.9]{Test Error};
	\node at (0,-2.8) [scale=.9]{Width Parameter (\# of filters $k$)};%Each bar in the figure shows the correlations' geometric mean of datasets in this group when training to corresponding non-zero level. 
	\end{tikzpicture}\caption{\small{With and without retraining pruned models.}}\label{figCFb}\vspace{-0.2cm}
\end{subfigure}\caption{\small{We train and prune ResNet-20 models on CIFAR-10 and also add randomly pruning and non-retraining curves. Solid lines here are exactly the same as Figure \ref{figNN} which show test errors of dense and $s$-sparse models. Dotted lines are test errors of random-based pruned models in (a) and of magnitude-based pruned models however without retraining in (b).}}\label{figCF}\vspace{-10pt}
\end{figure}

\begin{figure}[t!]
        \centering
	\begin{subfigure}{1.8in}\vspace{-5pt}
		\centering
		\begin{tikzpicture}
		\node at (0,0) {\includegraphics[scale=0.23]{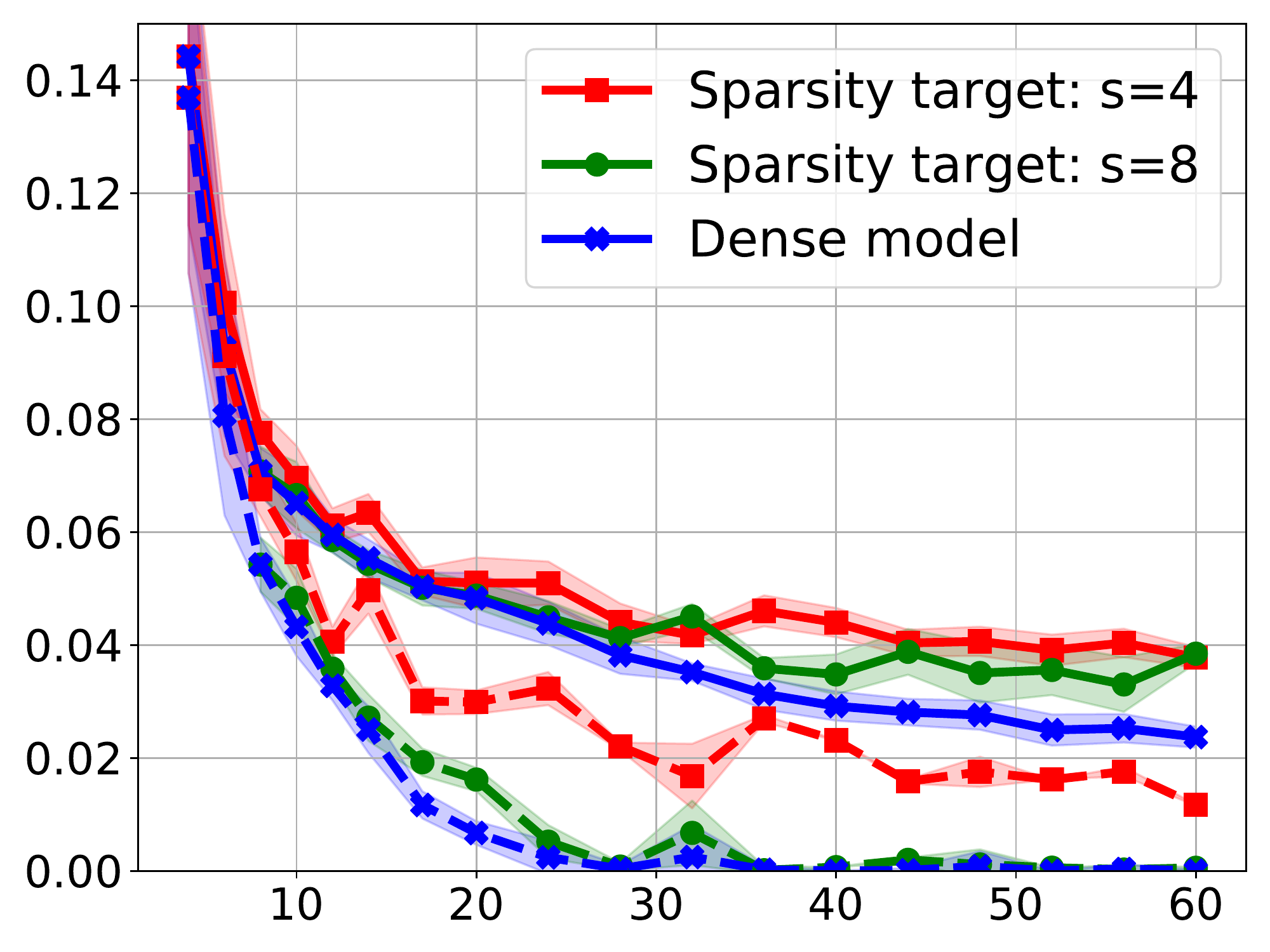}};
		\node at (-2.4,0) [rotate=90,scale=.9]{Training / Test Error};
		\node at (0,-1.9) [scale=.9]{Width Parameter (\# of nodes $k$)};%Each bar in the figure shows the correlations' geometric mean of datasets in this group when training to corresponding non-zero level. 
		\end{tikzpicture}\caption{\small{Dense and magnitude-based pruned and retrained models.}}\label{figMNISTa}\vspace{-0.2cm}
	\end{subfigure}~~~~~\begin{subfigure}{1.8in}\vspace{-5pt}
	\centering
	\begin{tikzpicture}
	\node at (0,0) {\includegraphics[scale=0.23]{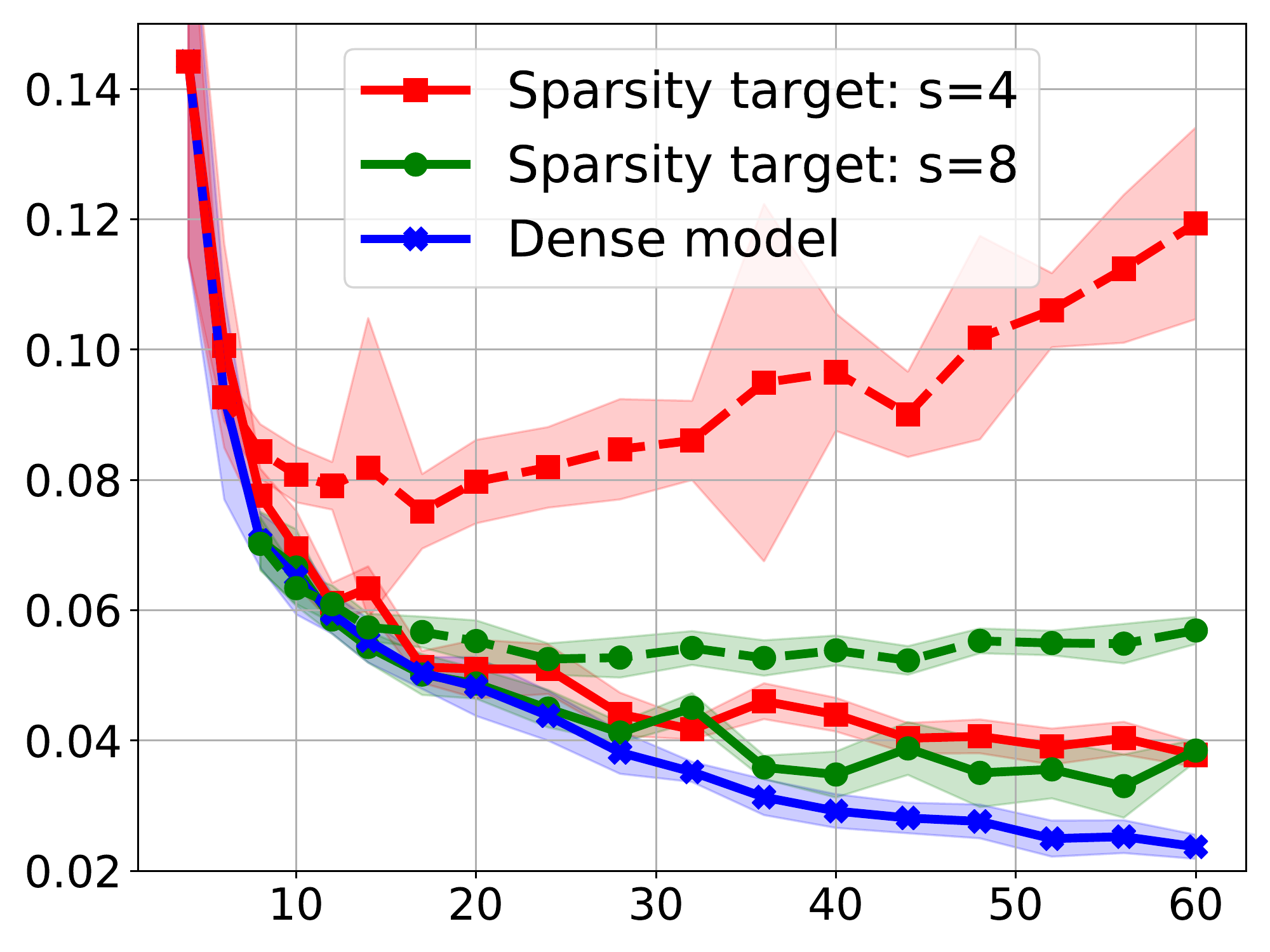}};
	\node at (-2.4,0) [rotate=90,scale=.9]{Test Error};
	\node at (0,-1.9) [scale=.9]{Width Parameter (\# of nodes $k$)};%Each bar in the figure shows the correlations' geometric mean of datasets in this group when training to corresponding non-zero level. 
	\end{tikzpicture}\caption{\small{Magnitude-based and randomly pruned models.}}\label{figMNISTb}\vspace{-0.2cm}
\end{subfigure}~~~~~\begin{subfigure}{1.8in}\vspace{-5pt}
	\centering
	\begin{tikzpicture}
	\node at (0,0) {\includegraphics[scale=0.23]{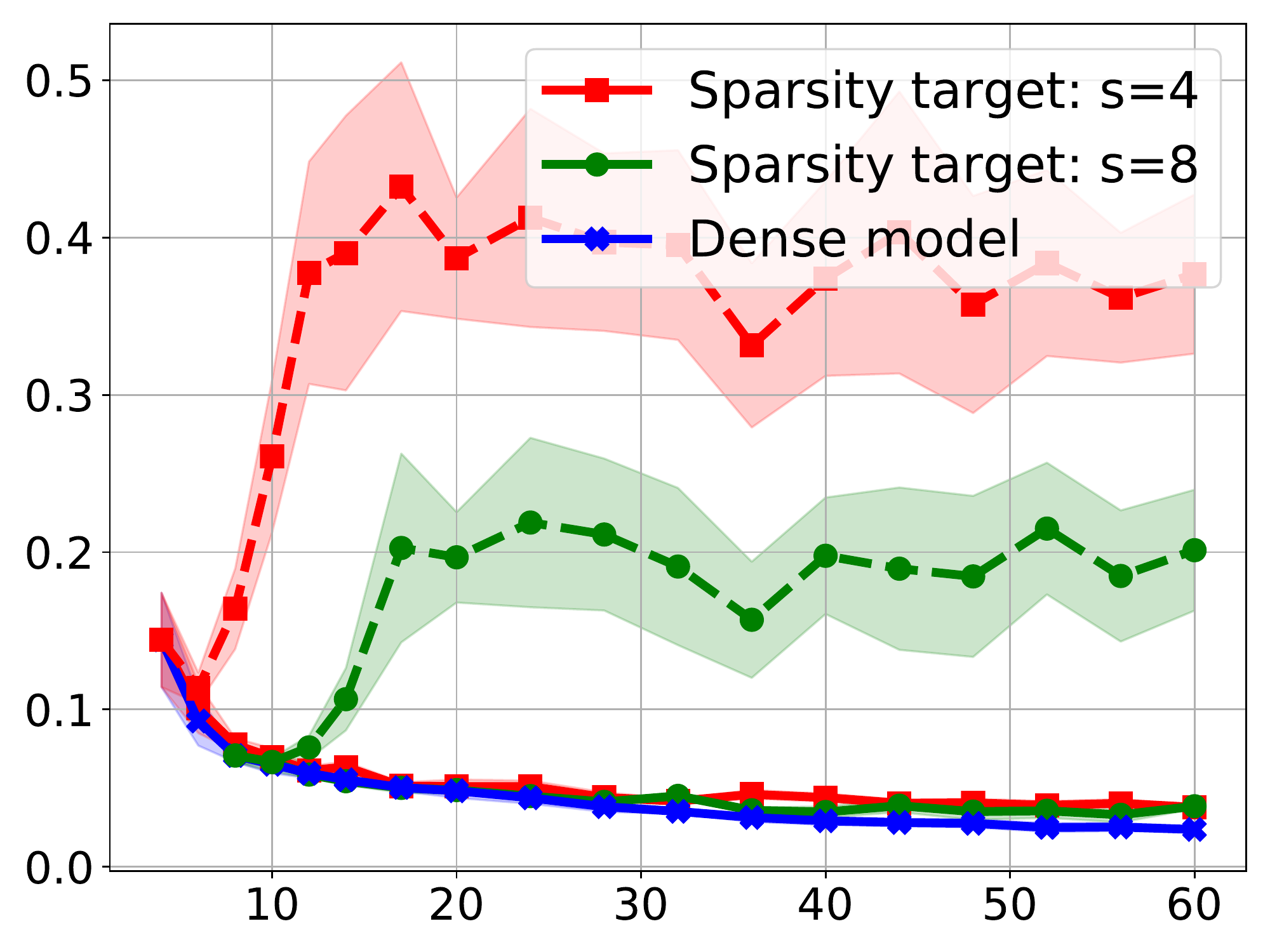}};
	\node at (-2.4,0) [rotate=90,scale=.9]{Test Error};
	\node at (0,-1.9) [scale=.9]{Width Parameter (\# of nodes $k$)};%Each bar in the figure shows the correlations' geometric mean of datasets in this group when training to corresponding non-zero level. 
	\end{tikzpicture}\caption{\small{With and without retraining pruned models.}}\label{figMNISTc}\vspace{-0.2cm}
\end{subfigure}\caption{\small{Here we use the simplest neural network consisting of two fully-connected layers to train MNIST with various width and sparsity targets. In this architecture the width parameter equivalent to the number of hidden layer nodes controls model size. Same as Figure~\ref{figNN} the blue line corresponds to training dense models with width $k$. As for the red and green lines we choose $4$- and $8$-width models as base models respectively and prune $k$-width dense model to corresponding sparsity targets. (a) shows both training and test errors after magnitude-based pruning and retraining. In (b) solid and dotted lines are test errors of magnitude- and random-based pruned models with retraining. To verify the importance of retraining in neural networks we present the test errors of magnitude-based models without retraining in (c).}}\label{figMNIST}\vspace{-10pt}
\end{figure}
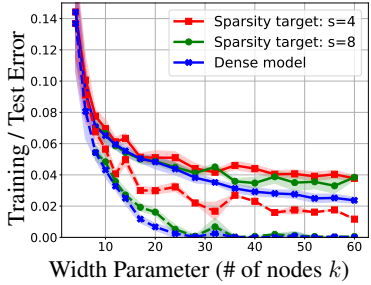
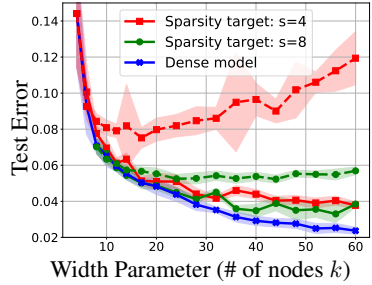
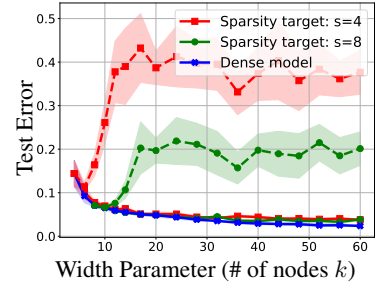

\section{Further Experiments and Intuitions}\label{SM mot}
%\newpage
%\newpage
% (Chang, Lizzie please fill here)
%\subsection{Verifying Analytical Predictions

%\subsection{Experiments on Neural Networks}
%\som{Explain retraining of Figure 1.}

\subsection{Further discussion and experiments on CIFAR-10} \label{ssec:SMCF}
First, we provide further discussion on Figure \ref{figNN}. Recall that, in this figure, we apply \emph{train$\rightarrow$prune$\rightarrow$retrain} to obtain the sparse neural networks. The test error and the training errors for the sparse models are evaluated at the end of the retraining phase. Thus, it is rather surprising that sparse models manage to achieve zero training error around the same width parameter $k$ as the dense model because while parameter count of the dense model increases in $k$, it is fixed for sparse models.

Secondly, we complement Figure \ref{figNN} with two additional experiments. The first experiment assesses the benefit of pruning compared to using a random nonzero pattern with the same sparsity. The second experiment assesses the benefit of retraining by comparing the curves in Fig \ref{figNN} with the test errors obtained without retraining. These two experiments are shown in Figure \ref{figCFa} and \ref{figCFb} and all of them are trained over same dataset and configured as given in Section \ref{ssec:NNE}. Instead of applying magnitude-based pruning, dotted red and green lines in Fig \ref{figCFa} are sparse models over $5$ or $10$ sparsity targets, pruned randomly to achieve same number of nonzeros as magnitude-based pruning strategy. Although the double descent phenomenon and downward trend are still present on the dotted lines, the performance is worse than magnitude-base pruning method. Fig. \ref{figCFb} shows how retraining benefits pruning ResNet-20 models. The results agree with our intuition that the non-retrained (dotted) lines achieve much bigger test error than retrained (solid) lines and overparameterization does not help in improving performance.

\subsection{MNIST Experiments with two layers} 
In Figure \ref{figMNIST} we train the simplest neural model with only 2 fully-connected layers over MNIST with various number of nodes to explore properties of magnitude-based pruning, random pruning and non-retraining on simple neural networks. Here, the number of nodes is equivalent to the width of the model, which directly controls the model size. Same as in Section \ref{ssec:NNE}, we select an $s$-width model as base model and prune trained dense models to the same sparsity. All experiments are trained with Adam optimization, $0.001$ learning rate and $200$ epochs under MNIST.  Solid red, green and blue lines in Figure \ref{figMNIST} correspond to test error of $4$-, $8$-sparsity targets and dense models. In Figure \ref{figMNISTa} dotted lines are training errors of dense and $s$-sparse models. As the width $k$ grows, the training and test error decrease for all dense and sparse models. The behavior is similar to Figure \ref{figNN} and training larger models benefit pruned-model accuracy however double descent is not really visible. We suspect that this may be because of the simpler nature of the MNIST dataset and LeNet architecture compared to the CIFAR10 dataset and ResNet-20 model. In Figure \ref{figMNISTb}, dotted lines apply randomly pruning. Different to magnitude pruning, where training bigger models and then pruning results in better performance, randomly pruning hurts when the sparsity level $s/k$ is relatively low. This is because under magnitude-based pruning, we can identity most of optimal entries of weights from trained dense model and retraining with these entries achieves lower errors. In constrast, random pruning learns nothing from the trained model and as the sparsity level decreases, the probability that random operator selects the limited optimal entries by chance also decreases, leading to worse performance. Dotted lines in Figure \ref{figMNISTc} show test errors of sparse models before retraining which educes the same conclusion in Section \ref{ssec:SMCF}, that is retraining is crucial to improve the performance in neural networks.

\subsection{Experiments on LGP}
\begin{figure}[t!]
        \centering
	\begin{subfigure}{3.5in}\vspace{-5pt}
		\centering
		\begin{tikzpicture}
		\node at (0,0) {\includegraphics[scale=0.33]{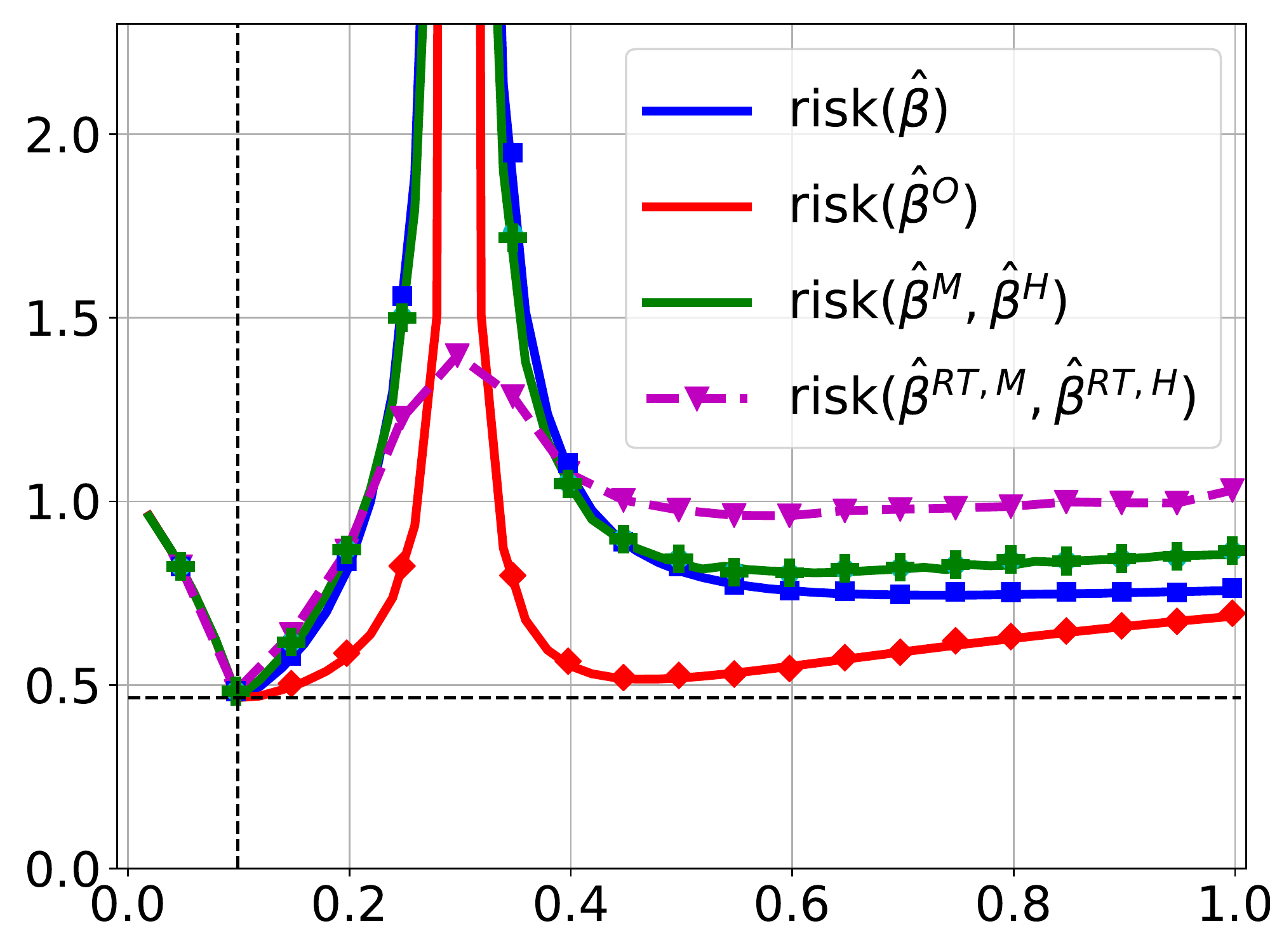}};
		\node at (-3.5,0) [rotate=90,scale=.9]{Test Risk};
		\node at (0,-2.8) [scale=.9]{Problem Size ($k/p$)};%Each bar in the figure shows the correlations' geometric mean of datasets in this group when training to corresponding non-zero level. 
		\end{tikzpicture}\caption{\small{Identity covariance, spiked latent weights.}}\label{figSM1a}\vspace{-0.2cm}
	\end{subfigure}~~~~~\begin{subfigure}{3.5in}\vspace{-5pt}
	\centering
	\begin{tikzpicture}
	\node at (0,0) {\includegraphics[scale=0.33]{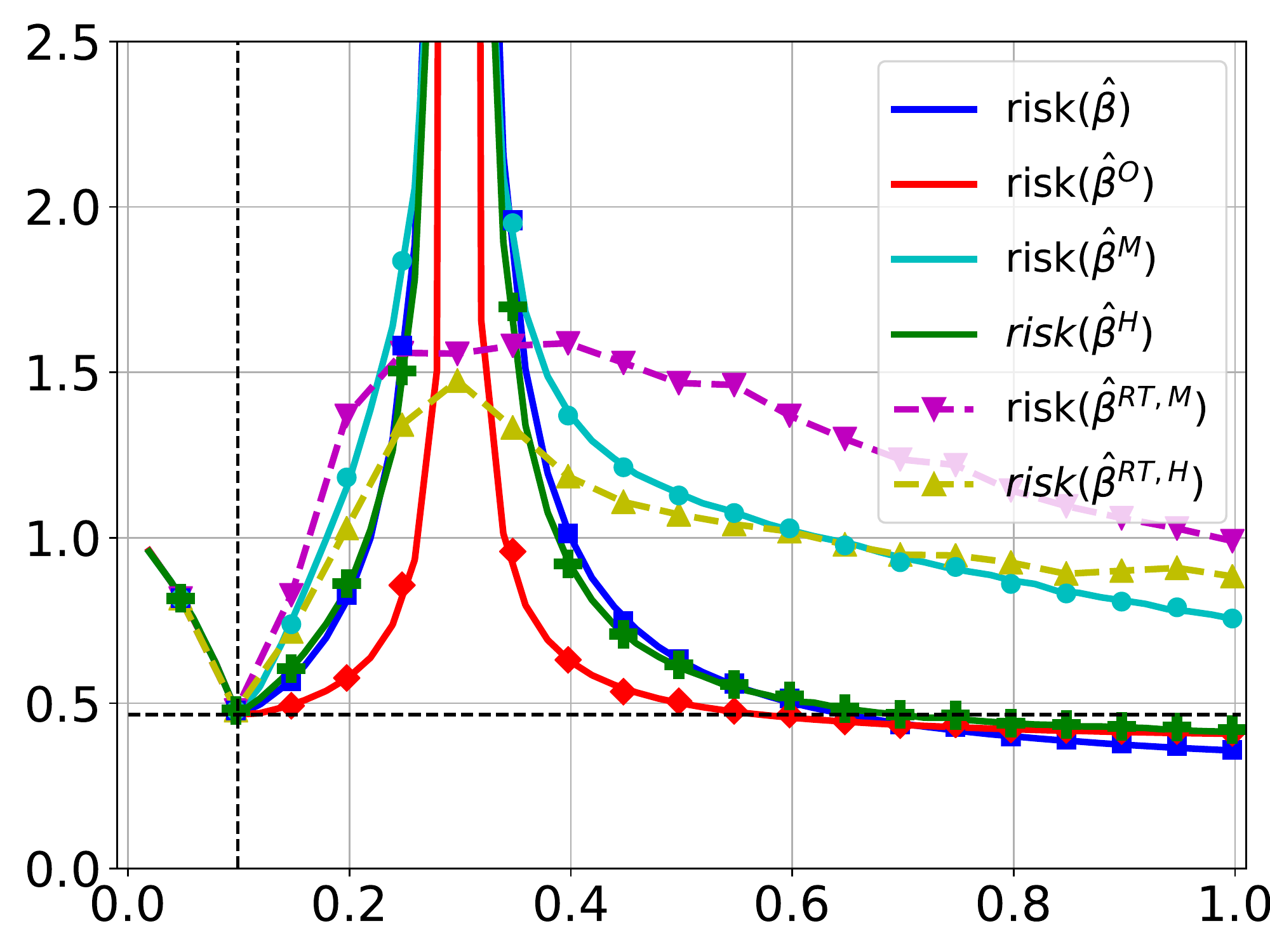}};
	%\node at (-2.4,0) [rotate=90,scale=1.]{Test Risk};
		\node at (-3.5,0) [rotate=90,scale=.9]{Test Risk};
	\node at (0,-2.8) [scale=.9]{Problem Size ($k/p$)};%Each bar in the figure shows the correlations' geometric mean of datasets in this group when training to corresponding non-zero level. 
	\end{tikzpicture}\caption{\small{Spiked covariance, identical latent weights.}}\label{figSM1b}\vspace{-0.2cm}
\end{subfigure}\caption{\small{Same as Figure \ref{fig1} however retraining curves are included. Our asymptotic prediction for various pruning strategies in linear gaussian models with $s/p=0.1$ and $n/p=0.3$. We solve ERM using the first $k$ features and then prune to obtain an $s$-sparse model. The vertical dashed line shows the $k=s$ point. The horizontal dashed line highlights the minimum risk among all underparameterized solutions including retraining. Retraining curves are not displayed.}}\label{figSM1}\vspace{-15pt}
\end{figure}
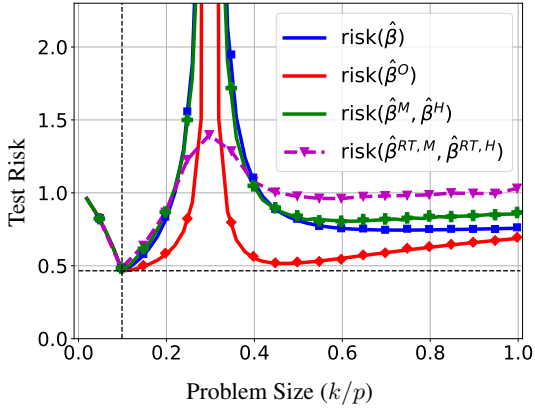
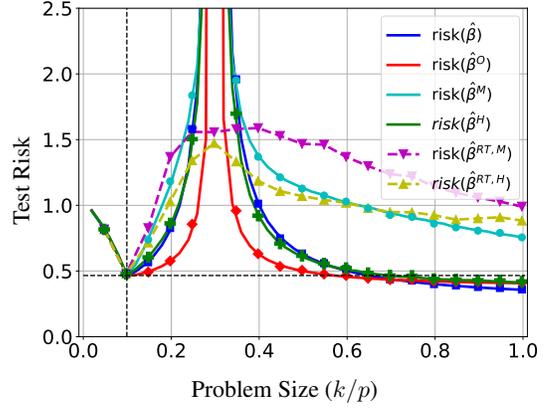

In Figure \ref{figSM1}, we carry out the identical experiments as in Figure \ref{fig1}. The difference is that we display two more figures which are the retraining curves for Magnitude- and Hessian-based pruning strategies shown in purple and yellow lines respectively. Figure \ref{figSM1a} is the counterpart of Figure \ref{fig1a} and Figure \ref{figSM1b} is the counterpart of Figure \ref{fig1b}. The main message in these experiments is that \emph{retraining hurts the performance}. This performance degradation is more emphasized in the overparameterized regime. Specifically, both retrained versions of Magnitude and Hessian pruning $\bth^{RT,M}$ and $\bth^{RT,H}$ perform consistently worse compared to their pruning-only counterparts $\bth^{M}$ and $\bth^{H}$. Observe that, the only regime where retraining outperforms the pruning-only approach is at the peak of double descent. This is the region where pruning-only risk diverges to infinity whereas retraining attains finite risk. This is because the retraining stage solves a well-conditioned problem and avoids the ill-conditioning occurring at $n=k$. Recall that, in light of Lemma \ref{lem rank one}, unlike the rank-one covariance case, retraining hurts because covariance is diagonal; thus, features are uncorrelated and do not have overlapping predictions.

%\subsection{Further Intuitions: Denoising affect of overparameterization}
%To provide further insights into the pruning benefits of overparameterization, consider a simple linear model with noise level $\sigma=0$, $n\geq p\gg s$ and identity covariance. Pick index set $\Delta\subset[p]$ with $|\Delta|=s$. Suppose we wish to estimate the coefficients $\bt^\st_{\Delta}$. For pruning, we can pick $\Delta$ to be the most salient/largest entries. If we solve the smaller problem over $\Delta$, $\bth(\Delta)$ will only provide a noisy estimate of $\bt^\st_{\Delta}$. The reason is that, the signal energy of missing features $[p]-\Delta$ act as noise uncorrelated with features in $\Delta$. Conversely, if we solve ERM with all features (the larger problem), we perfectly recover  $\bt^\st$ due to lack of noise and $n>p$ from which we perfectly estimate $\bt^\st_{\Delta}$. This simple argument, which is partly inspired by the missing feature setup in \cite{hastie2019surprises}, shows that solving the larger problem with more parameters can have a ``denoising-like effect" and perform better than the small problem. It also explains why retraining again with few features can potentially hurt the performance. This is because when features are uncorrelated (e.g.~diagonal covariance), missing features (due to the smaller problem size) act like additive uncorrelated noise. Our contribution obviously goes well beyond this discussion and theoretically characterizes exact asymptotics, addresses a general covariance model, and highlights the importance of the  overparameterized regime $n\ll p$.

\section{Proofs for overparameterized least-squares}\label{sec proof thm 1}
In this section, we assume the linear Gaussian problem (LGP) of Definition \ref{def LGP}, the overparameterized regime $k=p>n$ and the min-norm model $\bth$ of \eqref{eq:min_norm}. We prove Theorem \ref{thm:master_W2} that derives the asymptotic DC of $\bth$ and we show how this leads to sharp formulae for the risk of the Magnitude- and Hessian-pruned models.

\subsection{Notation and Assumptions}\label{sec:ass_app}
%\noindent\textbf{Notation.}~
For the reader's convenience, we recall some necessary notation and assumptions from Section \ref{sec main}.  We say that a function $f:\R^m\rightarrow\R$ is pseudo-Lipschitz of order $k$, denoted $f\in\rm{PL}(k)$, if there is a constant $L>0$ such that for all $\x,\y\in\R^m$,
$
|f(\x) - f(\y)|\leq L(1+\tn{\x}^{k-1}+\tn{\y}^{k-1})\|\x-\y\|_2 
$
(See also Section \ref{SM useful fact}). 
We say that a sequence of probability distributions $\nu_p$ on $\R^m$ {converges in $W_k$} to 
$\nu$, written $\nu_p\stackrel{W_k}{\Longrightarrow} \nu$, if $W_k(\nu_p,\nu) \rightarrow 0$ as $p \rightarrow \infty$. An equivalent definition is that, for any $f\in\rm{PL}(k)$, $\lim_{p\rightarrow\infty}\E f(X_p)=\E f(X)$, where expectation is with respect to $X_p\sim\nu_p$ and $X\sim\nu$ (e.g., \cite{montanari2017estimation}).  
Finally, recall that a sequence of probability distributions $\nu_n$ on $\R^m$ \emph{converges weakly} to $\nu$, if for any bounded Lipschitz function $f$: $\lim_{p\rightarrow\infty}\E f(X_p)=\E f(X)$, where expectation is with respect to $X_p\sim\nu_p$ and $X\sim\nu$.
Throughout, we use $C,C',c,c'$ to denote absolute constants (not depending on $n,p$) whose value might change from line to line.

We focus on a double asymptotic regime where:
$$n,p,s\rightarrow\infty \text{ at fixed overparameterization ratio } \kappa:=p/n>1 \text{ and sparsity level } \alpha:=s/p\in(0,1).$$  
For a sequence of random variables $\mathcal{X}_{p}$ that converge in probability to some constant $c$ in the limit of Assumption \ref{ass:linear} below, we  write $\mathcal{X}_{p}\rP c$. For a sequence of event $\Ec_p$ for which $\lim_{p\rightarrow}\Pro(\Ec_p) = 1$, we say that $\Ec_p$ occurs \emph{with probability approaching 1}. For this, we will often use the shorthand ``wpa 1". 

\vspace{5pt}
Next, we recall the set of assumption under which our analysis applies:

\asstwo*
\assthree*

%
%\vp
%\textbf{(A1)[Diagonal covariance].}~~The covariance matrix $\Sigmab$ is diagonal.
%
%%
%\vp
%\textbf{\ref{ass:mu}[Boundedness and empirical distribution].}~~There exist constants $\Sigma_{\min},\Sigma_{\max}\in(0,\infty)$ such that:
%$
%\Sigma_{\min}\leq\bSi_{i,i}\leq \Sigma_{\max}.
%$
%for all $i\in[p].$ 
%%There exists absolute constant  $C>0$ such that $\sqrt{p}|\betas_i|<C$, for all $i\in[p]$. 
%\cts{Furthermore, the joint empirical distribution of $\{(\lai,\sqrt{p}\betas_i)\}_{i\in[p]}$ converges weakly to a probability distribution $\mu$ on $\R_{>0}\times\R$ with bounded $4$th moment 
%%$\E_{(\Lambda,B)}\left[\left(\sqrt{\Lambda^2+B^2}\right)^3\right]$, 
%and assume that as $p\rightarrow\infty$, the $4$th moment of the empirical distribution converges to the $4$th moment of $\mu$, i.e.,
%$$
%\frac{1}{p}\sum_{i=1}^p\left(\sqrt{(\sqrt{p}\betas_i)^{2}+\lai^2}\right)^4 \rP \E_{(\Lambda,B)\sim\mu}\left[\left(\sqrt{\Lambda^2+B^2}\right)^4\right]<\infty.
%$$
%}
% for some $k\geq 3$.}

%\cts{Furthermore, the joint empirical distribution of $\{(\lai,\sqrt{p}\betas_i)\}_{i\in[p]}$ converges weakly to a probability distribution $\mu$ on $\R_{>0}\times\R$ with bounded $(2k-2)$-th moment, and assume that 
%$$
%\frac{1}{p}\sum_{i\in[p]}\sum_{i=1}^p(\sqrt{p}\betas_i)^{2k-2} \rP \E_{(\Lambda,B)\sim\mu}\left[ B^{2k-2}\right]<\infty,
%$$
%as $p\rightarrow\infty$ for some $k\geq 3$.}

\noindent We remark that Assumption \ref{ass:mu} above implies (see \cite[Lem.~4]{bayati2011dynamics} and \cite[Lem.~A3]{javanmard2013state}) that for any pseudo-Lipschitz function $\psi:\R^2\rightarrow\R$ of order $4$, i.e., $\psi\in\rm{PL}(4)$:
$$
\frac{1}{p}\sum_{i=1}^p\psi(\lai,\sqrt{p}\betas_i) \rP \E_{(\Lambda,B)\sim\mu}\left[ \psi(\Lambda,B)\right].
$$

\subsection{Asymptotic distribution and risk characterizations}
%, \fx{which is a special case of Theorem \ref{thm:master_W2}, tailored to our specific need in this paper, that is, characterizing the risk of pruned solutions.} \somm{Theorem 1 should be the special case. Not the other way around!!!}

{In this section, we prove our main result Theorem \ref{thm:master_W2}. Recall that $\bth$ is the min-norm solution. Since the distribution of $\bth$ depends on the problem dimensions (as it is a function of $\X,\y$), when necessary, we will use $\bth_n$ notation to make its dimension dependence explicit.} Let $\bth^P=\Pc(\bth)$ be a pruned version of the min-norm solution $\bth$. Recall from Section \ref{sec:risk}, that the first crucial step in characterizing the risk $\Lc(\bth^P)$ is studying the risk $\Lc(\bth^\Tc_t)$ of a threshold-based pruned vector. 
%Theorem \ref{thm:master_W2} below shows how this is possible. 

To keep things slightly more general,  consider $\bth^{g}$ defined such that $\sqrt{p}\bth^{g}=g(\sqrt{p}\bth)$, where $g$ is a Lipschitz function acting entry-wise on $\bth$ (for example, $g$ can be the {(arbitrarily close Lipschitz approximation of the)} thresholding operator $\Tc_t$ of Section \ref{sec:risk}). Then,  the risk of $\bth^g$ can be written as
\begin{align}
\Lc(\bth^g) &= \E_{\Dc}[(\x^T(\bt^\star-\bth^g) + \sigma z)^2] = \sigma^2 + (\bt^\star-\bth^g)^T\Sigmab(\bt^\star-\bth^g) \nn \\
  &= \sigma^2 + \frac{1}{p}\sum_{i=1}^p\Sigmab_{i,i}\big(\sqrt{p}\betas_i-g(\sqrt{p}\bth_i)\big)^2\nn
  \\
    &=: \sigma^2 + \frac{1}{p}\sum_{i=1}^p f\big(\sqrt{p}\bth_i,\sqrt{p}\betas_i,\Sigmab_{i,i}\big),\label{eq:risk_app_f}
% \\
%  &\rP \sigma^2 + \E\left[ \Lambda\left(B-\Tc_t(X_{\kappa,\sigma^2})\right) \right]\,.
\end{align}
where in the last line, we defined the function $f$ as $f=\fl\in \Fl\subset\Plt$ given by
\[
\fl(x,y,z) := z(y-g(x))^2\quad\text{where}~g~\text{is Lipschitz.}
\]
Here, recall the definition of the families $\Fl$ in \eqref{eq:fdef} and $\Plt$ in \eqref{eq:pdef}.

The following theorem establishes the asymptotic limit of \eqref{eq:risk_app_f}. For the reader's convenience, we repeat the notation introduced in Definition \ref{def:Xi}.
%\begin{definition}[Asymptotic DC -- Overparameterized regime]\label{def:Xi}
Let random variables $(\Lambda,B)\sim \mu$ (where $\mu$ is defined in Assumption \ref{ass:mu}) and fix $\kappa>1$. Define parameter $\xi$ as the unique positive solution to the following equation
%\begin{align}\label{eq:ksi}
$$
\E_{\mu}\Big[ \big({1+(\xi\cdot\Lambda)^{-1}}\big)^{-1} \Big] = {\kappa^{-1}}\,.
$$
Further define the positive parameter $\gamma$ as follows:
$$
\hspace{-0.1in}\gamma := 
%\frac{\sigma^2 + \E_{\mu}\left[\frac{N^2}{(\Lambda^{-1}+\xi)^2}\right]}{1-\kappa\E_{\mu}\left[\frac{1}{\left(1+(\xi\Lambda)^{-1}\right)^2}\right]}  
\Big({\sigma^2 + \E_{\mu}\Big[\frac{B^2\Lambda}{(1+\xi\Lambda)^2}\Big]}\Big)\Big/\Big({1-\kappa\E_{\mu}\Big[\frac{1}{\left(1+(\xi\Lambda)^{-1}\right)^2}\Big]}\Big).
$$
%\ct{@Samet: In your notation $\gamma$ should be $\Gamma\kappa$. Everything matches except the second term in the numerator. In your notation, that second term becomes $\int_0^1\bz^2(t)\bt^2(t) \Sigmab(t) dt$ (note the extra $\Sigmab(t)$ compared to your formula)
%}
With these and $H\sim\Nn(0,1)$, define the random variable
$$
X_{\kappa,\sigma^2}:=X_{\kappa,\sigma^2}(\Lambda,B,H) := \Big(1-\frac{1}{1+ \xi\Lambda}\Big) B + \sqrt{\kappa}\frac{\sqrt{\gamma}\,\Lambda^{-1/2}}{1+(\xi\Lambda)^{-1}} H, 
$$
and let $\Pi_{\kappa,\sigma^2}$ be its distribution.
%\end{definition}

%Recall the definition of the measure $\Pi_{\kappa,\sigma^2}$ in Definition \ref{def:Xi}. Then, 
%$\hat\Pi_n(\y,\X,\betas,\Sigmab)$ converges in Wasserstein-2 distance to $\Pi_{\kappa,\sigma^2}\otimes\mu$.
%Let $\cal{H}$ be the hard-thresholding operation. Draw $\hat{\bt}\sim \Dca(\bt,\bSi)$.
%\[
%\text{risk}_{\bSi^p}(\Hc(\hat{\bt}^p))\rightarrow \text{risk}_{\bSi}(\Hc(\hat{\bt}))
%\]
%\end{theorem}
%
%\begin{corollary} 
%Moreover, for the function $\fl:\R^3\rightarrow\R$, defined in \eqref{eq:fdef} it holds that
%Specifically, for any function $f:\R^k\rightarrow\R$, $f\in\rm{PL}(k)$, it holds that
%where the expectation is over $(\Lambda,B,H)\sim\mu\otimes\Nn(0,1).$
%Specifically, letting $\cal{H}:\R\rightarrow\R$ denote the hard-thresholding operation, it holds that
%\begin{align}
%\| \cal{H}( \hat\w^p )\|_2^2 \rP %\E\left[\left(\cal{H}\left(X_{\kappa,\sigma^2}(\Lambda,N,H) \right)\right)^2\right]
%\end{align}
%%\end{corollary}

\mainthm*

Before we prove the theorem, let us show how it immediately leads to a sharp prediction of the risk behavior. Indeed, a direct application of \eqref{eq:thm} for $f=\fl$ to \eqref{eq:risk_app_f} shows that
\begin{align}
\Lc(\bth^g)\rP \sigma^2 +  \E_{(\Lambda,B,H)\sim\mu\otimes\Nn(0,1)}\left[\fl(X_{\kappa,\sigma^2},B,\Lambda) \right] = \sigma^2 +  \E_{(\Lambda,B,H)\sim\mu\otimes\Nn(0,1)}\left[\Sigma\left(B-g(X_{\kappa,\sigma^2})\right)^2 \right].\label{eq:risk_app_f2}
\end{align}

We further remark on the following two consequences of Theorem \ref{thm:master_W2}. 

First, since \eqref{eq:thm} holds for any $\rm{PL}(2)$ function, we have essentially shown that $\hat\Pi_n(\y,\X,\betas,\Sigmab)$ converges in Wasserstein-2 distance to $\Pi_{\kappa,\sigma^2}\otimes\mu$, where recall that $\Pi_{\kappa,\sigma^2}$ is the distribution of the random variable $X_{\kappa,\sigma^2}$.

Second, the theorem implies that the empirical distribution of $\sqrt{p}\bth_n$ converges weakly to $\Pi_{\kappa,\sigma^2}$. To see this, apply \eqref{eq:thm} for the $\rm{PL}(2)$ function $f(x,y,z) = \psi(x)$ where  $\psi:\R\rightarrow\R$ is a bounded Lipschitz test function. 

\subsection{Proof of Theorem \ref{thm:master_W2}}
\vspace{5pt}
Let $\X\in\R^{n\times p}$ have zero-mean and normally distributed rows with a diagonal covariance matrix $\bSi=\E[\x\x^T]$. Given a ground-truth vector $\betas$ and labels $\y=\X\betas+\sigma \z,~\z\sim\Nn(0,\Iden_n)$, we consider the least-squares problem subject to the minimum Euclidian norm constraint (as $\kappa=p/n>1$) given by
\begin{align}\label{eq:PO_beta}
\min_{\bt}\frac{1}{2}\tn{\bt}^2\quad\text{subject to}\quad \y=\X\bt.
\end{align}
%, we solve the least-squares problem  \eqref{optim me2} subject to the . 
It is more convenient to work with the following change of variable: 
\begin{align}\label{eq:w}
\w:=\sqrt{\Sigmab}(\bt-\betas).
\end{align}
 With this, the optimization problem  in \eqref{eq:min_norm} can be rewritten as
\begin{align}\label{eq:PO}
\Phi(\X)=\min_{\w} \frac{1}{2}\tn{\Sigmab^{-1/2}\w+\betas}^2\quad\text{subject to}\quad \Xb\w=\sigma \z,
\end{align}
where we set $\Xb=\X\Sigmab^{-1/2}\distas\Nn(0,1)$. First, using standard arguments, we show that the solution of \eqref{eq:PO} is bounded. Hence, we can constraint the optimization in a sufficiently large compact set without loss of generality.

\begin{lemma}[Boundedness of the solution]\label{lem:bd_PO}
Let $\wh_n:=\wh_n(\X,\z)$ be the minimizer in \eqref{eq:PO}. Then, with probability approaching 1, it holds that $\wh_n\in\Bc$, where
$$\Bc:=\left\{\w\,|\,\|\w\|_2\leq B_{+} \right\},\qquad B_+:=5\sqrt{\frac{\Sigma_{\max}}{\Sigma_{\min}}}\frac{\sqrt{\kappa}+1}{\sqrt{\kappa}-1}(\sqrt{\Sigma_{\max}\E\left[B^2\right]} + \sigma).
% + \sqrt{\Sigma_{\max}\E\left[B^2\right]}.
$$ 
\end{lemma}
\begin{proof}
First, we show that the min-norm solution $\bth=\X^T(\X\X^T)^{-1}\y$ of \eqref{eq:PO_beta} is bounded. Note that $\kappa>1$, thus $\X\X^T$ is invertible wpa 1. We
have,
\begin{align}
\tn{\bth_n}^2 = \y^T(\X\X^T)^{-1}\y \leq \frac{\tn{\y}^2}{\la_{\min}(\X\X^T)}  = \frac{\tn{\y}^2}{\la_{\min}(\Xb\Sigmab\Xb^T)} \leq  \frac{\tn{\y}^2}{\la_{\min}(\Xb\Xb^T)\,\Sigma_{\min}} = \frac{\tn{\y}^2}{\sigma_{\min}^2(\Xb)\,\Sigma_{\min}}. \label{eq:Ubb}
\end{align}
But, wpa 1, 
$
\sigma_{\min}(\Xb)/\sqrt{n} \geq \frac{1}{2}\left(\sqrt{\kappa}-1\right).
$
Furthermore, 
$
\|\y\|_2 \leq \|\Xb\Sigmab^{1/2}\betas\|_2 + \sigma\|\z\|_2 \leq \sigma_{\max}(\Xb)\sqrt{\Sigma_{\max}} \|\betas\|_2 + \sigma\|\z\|_2.
$
Hence, wpa 1,
$$
\|\y\|_2/\sqrt{n} \leq 2(\sqrt{\kappa}+1)\sqrt{\Sigma_{\max}} \sqrt{\E\left[B^2\right]} + 2\sigma,
$$
where we used the facts that wpa 1: $\|z\|_2/\sqrt{n}\rP 1$, $\sigma_{\max}(\Xb)<\sqrt{2n}(\sqrt{\kappa}+1)$ and \cts{by Assumption \ref{ass:mu}}:
$$
\|\betas\|_2^2 = \frac{1}{p}\sum_{i=1}^{p}(\sqrt{p}\betas_i)^2 \rP \E\left[B^2\right].
$$
Put together in \eqref{eq:Ubb}, shows that
\begin{align}
\tn{\bth_n} < \frac{2(\sqrt{\kappa}+1)\sqrt{\Sigma_{\max}} \sqrt{\E\left[B^2\right]} + 2\sigma}{\sqrt{\Sigma_{\min}}(\sqrt{\kappa}-1)/2} =: \tilde{B}_+.\label{eq:bd_beta}
\end{align}
Recalling that $\wh_n= \sqrt{\Sigmab}\bth_n - \sqrt{\Sigmab}\betas$, we conclude, as desired, that wpa 1,
$
\tn{\wh_n} \leq \sqrt{\Sigma_{\max}}\tilde{B}_+ + \sqrt{\Sigma_{\max}} \sqrt{\E[B^2]} \leq B_+.
$
%{\color{red}Missing proof}
\end{proof}

%%%%%%%%%%%%%%%%%%%%%%%%%%%%%%%%%
%%%% Samet's k-th bounded %%%%%%%%%%%%%%%%%
%\input{kth_bounded}
%%%%%%%%%%%%%%%%%%%%%%%%%%%%%%%%%

Lemma \ref{lem:bd_PO} implies that nothing changes in \eqref{eq:PO} if we further constrain $\w\in\Bc$ in \eqref{eq:PO}. Henceforth, with some abuse of notation, we let
\begin{align}\label{eq:PO_bd}
\Phi(\X):=\min_{\w\in\Bc} \frac{1}{2}\tn{\Sigmab^{-1/2}\w+\betas}^2\quad\text{subject to}\quad \Xb\w=\sigma \z,
\end{align}

Next, in order to analyze the primary optimization (PO) problem in \eqref{eq:PO_bd} in apply the CGMT \cite{thrampoulidis2015regularized,thrampoulidis2018precise}. Specifically, we use the constrained formulation of the CGMT given by Theorem \ref{thm closed}.  Specifically, the auxiliary problem (AO) corresponding to \eqref{eq:PO_bd} takes the following form with $\g\sim\Nn(0,\Iden_n)$, $\h\sim\Nn(0,\Iden_p)$, $h\sim \Nn(0,1)$:% (non-rigorously)
\begin{align}
\phi(\g,\h) = \min_{\w{\in\Bc}} \frac{1}{2}\tn{\Sigmab^{-1/2}\w+\betas}^2\quad\text{subject to}\quad \tn{\g}\tn{\w~{\sigma}}\leq \h^T\w+\sigma h.\label{eq:AO_con}
\end{align}%\tn{\h}
%Set $\bar{\h}=\h/\sqrt{p}$.

We will prove the following techincal result about the AO problem.
%\somm{Re-specify the assumptions of this lemma!}
\begin{lemma}[Properties of the AO -- Overparameterized regime]\label{lem:AO}
{Let the assumptions of Theorem \ref{thm:master_W2} hold.} Let $\phi_n=\phi(\g,\h)$ be the optimal cost of the minimization in \eqref{eq:AO_con}. Define $\bar\phi$ as the optimal cost of the following deterministic min-max problem
\begin{align}\label{eq:AO_det}
\bar\phi:=\max_{u\geq 0}\min_{\tau>0}~ \Dc(u,\tau):=\frac{1}{2}\left({u\tau} + \frac{u\sigma^2}{\tau} - u^2\kappa\,\E\left[ \frac{1}{\Lambda^{-1}+\frac{u}{\tau}} \right] - {\E\left[\frac{B^2}{1+\frac{u}{\tau}\Lambda}\right]} \right).
\end{align}
The following statements are true.
%\somm{Important: Added a AO subscript to AO related terms!}

\noindent{(i).}~The AO minimization in \eqref{eq:AO_con} is $\frac{1}{\Sigma_{\max}}$-strongly convex and has a unique minimizer $\wha_n:=\wha_n(\g,\h)$.

\noindent{(ii).}~In the limit of $n,p\rightarrow\infty, p/n=\kappa$, it holds that $\phi(\g,\h)\rP\bar\phi$, i.e., for any $\eps>0$:
$$
\lim_{n\rightarrow\infty}\P\left(|\phi(\g,\h)-\bar\phi|>\eps\right) = 0.
$$

\noindent{(iii).} The max-min optimization in \eqref{eq:AO_det} has a unique saddle point $(u_*,\tau_*)$ satisfying the following: 
$$
u_*/\tau_* = \xi\quad\text{and}\quad\tau_* = \gamma,
$$
where $\xi, \gamma$ are defined in Definition \ref{def:Xi}.

\noindent{(iv).}~Let $f:\R^3\rightarrow\R$ be a function in \fx{$\rm{PL}(3)$}. Let $\btha_n=\btha_n(\g,\h)=\Sigmab^{-1/2}\wha_n + \betas$. Then,
$$
\frac{1}{p}\sum_{i=1}^{p}f\left(\sqrt{p}\btha_n,\sqrt{p}\betas,\Sigmab\right) \rP \E_{(B,\Lambda,H)\sim\mu\otimes	\Nn(0,1)}\left[f\left(X_{\kappa,\sigma^2}(B,\Lambda,H),B,\Lambda\right) \right].
$$
\fx{In particular, this holds for all functions  $f\in \Plt$ defined in \eqref{eq:pdef}.}%\CT{Stating this way so that it becomes more clear where the $W_4$ convergence requirement comes from. See also remark below.}

\noindent{(v).}~\cts{The empirical distribution of $\btha_n$ converges weakly to the measure of $X_{\kappa,\sigma^2}$, and also,  for some absolute constant $C>0$:
\begin{align}\label{eq:k_AO}
\tn{\btha}^2 < C\quad\text{wpa 1.}
%\sum_{i=1}^{p}(\btha_n)^2
% \rP \E\left[(X_{\kappa,\sigma^2})^{2k-2}\right] < \infty.
\end{align}
}

\end{lemma}

%\begin{lemma}\label{lem:deviation}
%Let $\wh_n=\wh_n(\X,\z)$ be a minimizer of the PO in \eqref{eq:PO}. 
%%Further let $\tau_n=\tau_n(\g,\h,\z)$ and $u_n=u_n(\g,\h,\z)$ be the unique saddle point of the min-max optimization in \eqref{eq:AO_4}. Define $\wh_n=\wh_n(\tau_n,u_n)$ as in \eqref{eq:w_n}. 
%For any pseudo-Lipschitz function $f:\R\rightarrow\R$ of order 2, it hold that
%\begin{align}
%\frac{1}{p} \sum_{i=1}^{p} f\left(\hat\w_{n,i}\right) \rP \E_\mu\left[X_{\kappa,\sigma^2}(\Lambda,N,H)  \right].
%\end{align}
%\end{lemma}

%\begin{proof}[Proof of Lemma \ref{lem:deviation}]
We prove Lemma \ref{lem:AO} in Section \ref{sec:proofAO}. \fx{We remark that Assumption \ref{ass:mu} on $W_4$-convergence of the joint empirical distribution of $\{(\Sigmab_{i,i},\sqrt{p}\bts_i)\}_{i\in[p]}$ is required in the proof of the statement (iv) above. More generally if $W_k$-convergence is known for some integer $k$, then statement (iv) above holds for test functions $f\in\rm{PL}(k-1)$. This is the first place in the proof of Theorem \ref{thm:master_W2}, where we use the assumption $f\in\Plt$; indeed, we show in Lemma \ref{lem:fL} that $\Fl\subset\Plt\subset\rm{PL}(3)$. The second part is in proving the perturbation result in \eqref{eq:dev2show} below. Unlike the former, when proving the perturbation result, the requirement $f\in\Plt$ cannot be relaxed (e.g.~to $f\in\rm{PL}(k-1)$) by simply increasing the order of $W_k$-convergence in Assumption \ref{ass:mu}.}

\subsubsection{Finalizing the proof of Theorem \ref{thm:master_W2}:}
Here, we show how Lemma \ref{lem:AO} leads to the proof of Theorem \ref{thm:master_W2} when combined with the CGMT framework \cite{thrampoulidis2015regularized,thrampoulidis2018precise}.

\cts{Let $f:\R^3\rightarrow\R$ be a function in $\Plt$, where $\Plt$ was defined in \eqref{eq:pdef}.} For convenience, 
 define 
$$F_n(\bth_n,\betas,\Sigmab):=\frac{1}{p} \sum_{i=1}^{p} f\left(\sqrt{p}\bth_{n,i},\sqrt{p}\betas_{i},\Sigmab_{ii}\right)\quad\text{and}\quad\alpha_*:=\E_\mu\left[f(X_{\kappa,\sigma^2}(\Lambda,B,H),B,\Lambda)\right].$$
Fix  any $\eps>0$ and define the set 
\begin{align}\label{eq:S_set}
\Sc = \Sc(\betas,\Sigmab) 
%=\{\bt\bgl |F_n(\bth_n,\betas,\Sigmab)-\alpha_*|\geq 2\eps\}.
=\big\{\w=\sqrt{\Sigmab}(\betab-\bts) \in\Bc \bgl |F_n(\bt,\betas,\Sigmab)-\alpha_*|\geq 2\eps\big\}.
\end{align}
\fy{With this definition, observe that, it suffices to prove that the solution $\hat\w_n=\sqrt{\Sigmab}(\bth_n-\bts)$ of the PO in \eqref{eq:PO_beta} satisfies $\hat\w_n\not\in\Sc$ wpa 1.}
% To see that this is sufficient, note the following. On the one hand, setting $f=\fl\in \Fl$, directly proves \eqref{eq:thm}. On the other hand, recall that $W_2$-convergence is equivalent to convergence of any $\rm{PL}(2)$ test function $f$, e.g. see \cite[Sec. 6.1]{montanari2017estimation}. 

%\ct{Note that this $\Sc$ is a random set now.}
To prove the desired, we need to consider the ``perturbed" PO and AO problems (compare to \eqref{eq:PO} and \eqref{eq:AO_con}) as:
\begin{align}\label{eq:PO_S}
\Phi_S(\X)=\min_{\w\in\Sc} \frac{1}{2}\tn{\Sigmab^{-1/2}\w+\betas}^2\quad\text{subject to}\quad \Xb\w=\sigma \z,
\end{align}
and
\begin{align}
\phi_S(\g,\h)=\min_{\w\in\Sc} \frac{1}{2}\tn{\Sigmab^{-1/2}\w+\betas}^2\quad\text{subject to}\quad \tn{\g}\tn{\w~{\sigma}}\leq \h^T\w+\sigma h.\label{eq:AO_S}
\end{align}%\tn{\h}
Recall here, that $\Xb=\X\Sigmab^{-1/2}\distas\Nn(0,1)$, $\g\sim\Nn(0,\Iden_n)$, $\h\sim\Nn(0,\Iden_p)$, $h\sim \Nn(0,1)$ and we have used the change of variables $\w:=\sqrt{\Sigmab}(\bt-\betas)$ for convenience. 

 Using \cite[Theorem 6.1(iii)]{thrampoulidis2018precise} it suffices to find costants $\bar\phi, \bar\phi_S$ and $\eta>0$ such that the following three conditions hold:
\begin{enumerate}
\item $\bar\phi_S \geq \bar\phi + 3\eta$,
\item $\phi(\g,\h) \leq \bar\phi + \eta$, with probability approaching 1,
\item $\phi_S(\g,\h) \geq \bar\phi_S - \eta$, with probability approaching 1.
\end{enumerate}
In what follows, we  explicitly find $\bar\phi, \bar\phi_S,\eta$ such that the three conditions above hold.

\vspace{5pt}
\noindent\underline{Satisfying Condition 2}: Recall the deterministic min-max optimization in \eqref{eq:AO_det}. Choose $\bar\phi=\Dc(u_*,\tau_*)$ be the optimal cost of this optimization. From Lemma \ref{lem:AO}(ii), $\phi(\g,\h)\rP\bar\phi$. Thus, for any $\eta>0$, with probability approaching 1:
\begin{align}\label{eq:phi_lim}
\bar\phi + \eta \geq \phi(\g,\h) \geq \bar\phi - \eta.
\end{align}
Clearly then, Condition 2 above holds for any $\eta>0$. 

\vspace{5pt}
\noindent\underline{Satisfying Condition 3}: Next, we will show that the third condition holds for appropriate $\bar\phi$. 
%Let $\tau_n=\tau_n(\g,\h,\z)$ and $u_n=u_n(\g,\h,\z)$ be the unique saddle point of the min-max optimization in \eqref{eq:AO_4}. Define 
Let $\wha_n=\wha_n(\g,\h)$ be the unique minimizer of \eqref{eq:AO_con} as per Lemma \ref{lem:AO}(i), i.e., $\frac{1}{2}\tn{\Sigmab^{-1/2}\wha_n+\bt}^2 = \phi(\g,\h)$. Again from Lemma \ref{lem:AO}, the minimization in \eqref{eq:AO_con} is $1/\Sigma_{\max}$-strongly convex in $\w$. Here, $\Sigma_{\max}$ is the upper bound on the eigenvalues of $\Sigmab$ as per Assumption \ref{ass:mu}. Thus, for any $\tilde\eps>0$ and any feasible $\w$ the following holds (deterministically):
\begin{align}\label{eq:sc}
\frac{1}{2}\tn{\Sigmab^{-1/2}\w+\bt}^2 \geq \phi(\g,\h) + \frac{\tilde\epsilon^2}{2{\Sigma_{\max}}},~\text{provided that}~ \|\w-\wha_n\|_2 \geq \tilde\epsilon.
\end{align}

Now, we argue that {wpa 1,}
\begin{align}\label{eq:dev_arg}
%~\bt\in\Sc \text{ implies that }
\text{for all}~\w\in \Sc~\text{it holds that}~\|\w-\wha_n\|_2\geq \tilde\eps,
\end{align}
for an appropriate value of a constant $\tilde\eps>0$.

 Consider any $\w\in\Sc$. 

 First, by definition in \eqref{eq:S_set}, for $\betab=\Sigmab^{-1/2}\w+\bts$ we have that
$$
|F_n(\bt,\betas,\Sigmab)-\alpha_*| \geq 2\eps.
$$

Second, by Lemma \ref{lem:AO}(iv), with probability approaching 1,
$$
|F(\btha_n,\betas,\Sigmab) - \alpha_*| \leq \epsilon.
$$

Third, we will show that wpa 1, there exists universal constant $C>0$ such that
\begin{align}
|F_n(\btha_n,\betas,\Sigmab) - F_n(\bt,\betas,\Sigmab)| \leq C {\|\btha_n - \bt\|_2}\label{eq:dev2show}.
\end{align}

Before proving \eqref{eq:dev2show}, let us argue how combining the above three displays shows the desired. Indeed, in that case, wpa 1,
\begin{align*}
2\eps &\leq |F_n(\bt,\betas,\Sigmab)-\alpha_*| \leq |F_n(\btha_n,\betas,\Sigmab) - F_n(\bt,\betas,\Sigmab)| + |F_n(\btha_n,\betas,\Sigmab) - \alpha_*| \\
&\leq \epsilon + C \,\|\bt-\btha_n\|_2. \\
&\qquad\Longrightarrow \|\bt-\btha_n\|_2 \geq {\eps}/{C}=:\hat\eps\\
&\qquad\Longrightarrow \|\w-\wha_n\|_2 \geq \hat\eps\sqrt{\Sigma_{\min}}=:\tilde\eps.
\end{align*}
In the last line above, we recalled that $\bt=\Sigmab^{-1/2}\w+\betas$ and $\Sigmab_{i,i}\geq\Sigma_{\min},~i\in[p]$ by Assumption \ref{ass:mu}. This proves \eqref{eq:dev_arg}.

Next, combining \eqref{eq:dev_arg} and \eqref{eq:sc}, we find that wpa 1,
$
\frac{1}{2}\tn{\Sigmab^{-1/2}\w+\bt}^2 \geq \phi(\g,\h) + \frac{\tilde\epsilon^2}{2\Sigma_{\max}},~\text{for all}~ \w\in\Sc.
$
Thus, 
\begin{align}
\phi_S(\g,\h) \geq \phi(\g,\h) + \frac{\tilde\epsilon^2}{2\Sigma_{\max}}.\nn
\end{align}
%\end{proof}
When combined with \eqref{eq:phi_lim}, this shows that
\begin{align}
\phi_S(\g,\h) \geq \bar\phi + \frac{\tilde\epsilon^2}{2\Sigma_{\max}} - \eta.
\end{align}
Thus, choosing $\bar\phi_S = \bar\phi + \frac{\tilde\epsilon^2}{2\Sigma_{\max}}$ proves the Condition 3 above.

\vp
\noindent{\textbf{Perturbation analysis via Pseudo-Lipschitzness (Proof of \eqref{eq:dev2show}).}} To complete the proof, let us now show \eqref{eq:dev2show}. Henceforth, $C$ is used to denote a universal constant whose value can change from line to line. \fy{Recall that $f\in \Plt$ where $\Plt:\R^2\times \Zc\rightarrow\R$ is the set of $\rm{PL}(3)$ functions such that $f(\cdot,\cdot,z)$ is $\rm{PL}(2)$ for all $z\in\Zc$. Suppose that the $\rm{PL}(2)$ constant of $f(\cdot,\cdot,z)$ is upper bounded over $z\in \Zc$ by some $C>0$. We also let $C$ change from line to line for notational simplicity. Then, we have the following chain of inequalities:
\begin{align}
|F_n(\btha_n(\g,\h),\betas,\Sigmab) - F_n(\bt,\betas,\Sigmab)|
&=\frac{1}{p}\sum_{i=1}^p|f(\sqrt{p}\btha_{n,i},\sqrt{p}\betas_i,\Sigmab_{i,i})-f(\sqrt{p}\bt_i,\sqrt{p}\betas_i,\Sigmab_{i,i})|
\nn\\
&\leq
 \frac{C}{p}\sum_{i=1}^p (1+ \|\sqrt{p}[\betas_i,\btha_{n,i}]\|_2 +  \|\sqrt{p}[\betas_i,\bt_{i}]\|_2) \sqrt{p}|\btha_{n,i}-\bt_{i}|\nn\\
&\leq
C \Big(1+ \frac{1}{\sqrt{p}}\big(\sum_{i=1}^{p}\|\sqrt{p}[\betas_i,\btha_{n,i}]\|_2^2\big)^{1/2} + \frac{1}{\sqrt{p}}\big(\sum_{i=1}^p\|\sqrt{p}[\betas_i,\bt_{i}]\|_2^2\big)^{1/2}\Big) \|\btha_n-\bt\|_2\nn\\
&\leq
%C  \left(1+ \|\betas\|_2^2+ \|\btha_n\|_2^2 + \|\bt\|_2^2\right) \|\btha_n-\bt\|_2.
C \left(1+ \max\{\|\betas\|_2^2,\|\btha_n\|_2^2,\|\bt\|_2^2\}^{1/2} \right) \|\btha_n-\bt\|_2.\label{eq:fcase_main}
 \end{align}
 %boundedness of $\lai$ as per Assumption \ref{ass:mu}. {In the third line, we used the fact that the function $\psi(a,b) = (a-g(b))^2$ is $\rm{PL}(2)$ as the quadratic of a Lipschitz function.}
 In the second line above, we used the fact that $f(\cdot,\cdot,z)$ is $\rm{PL}(2)$. The third line follows by Cauchy-Schwartz  inequality. Finally, in the last line, we used the elementary fact that $a+b+c\leq 3\max\{a,b,c\}$ for $a=2\sum_{i=1}^p(\betas_i)^2$ and $b=\sum_{i=1}^p(\btha_{n,i})^2$ and $c=\sum_{i=1}^p\bt_{i}^2$.}

Hence, it follows from \eqref{eq:fcase_main} that in order to prove \eqref{eq:dev2show}, we need to show boundedness of the following terms: $\|\btha_n\|_2$, $\|\betas\|_2$ and $\|\bt\|_2$. {By feasibility of $\btha_n$ and $\bt$, we know that $\btha_n,\bt\in\Bc$. Thus, the desired $\|\bt\|_2<\infty$ and $\|\btha_n\|_2<\infty$ follow directly by Lemma \ref{lem:bd_PO} (Alternatively, for $\btha_n$ we conclude the desired by directly applying Lemma \ref{lem:AO}(v)).} Finally, to prove $\|\betas\|_2<\infty$, note that
$$
\|\betas\|_2^2 =\frac{1}{p}\sum_{i=1}^p(\sqrt{p}\betas_i)^2,
$$
\cts{which is bounded wpa 1 by Assumption \ref{ass:mu}, which implies bounded second moments of $\sqrt{p}\betas$. } This completes the proof of \eqref{eq:dev2show}, as desired.
%And everything is easy now since $\|\betas\|_2^2$ is bounded as long as second moments of empirical distribution as bounded, $\|\bth_n\|_2^2$ is bounded if $\|\betas\|_2^2$ is bounded, and, $\|\bt\|_2^2$ is bounded, since $\bt\in\Bc$ by Lemma 1. To prove 

\vspace{5pt}
\noindent\underline{Satisfying Condition 1:} To prove Condition 1, we simply pick $\eta$ to satisfy the following
\begin{align}
\bar\phi_S > \bar\phi + 3 \eta ~\Leftarrow~ \bar\phi + \frac{\tilde\epsilon^2}{2\Sigma_{\max}} - \eta \geq \bar\phi + 3 \eta ~\Leftarrow~ \eta \leq \frac{\tilde\epsilon^2}{8\Sigma_{\max}}.\nn
\end{align}

This completes the proof of Theorem \ref{thm:master_W2}.

~~~~
~~~~

\subsection{Proof of Lemma \ref{lem:AO}}\label{sec:proofAO}

\vp
%\noindent\underline{Proof of (i):}~

\subsubsection{Proof of (i).} Strong convexity of the objective function in \eqref{eq:AO_con} is easily verified by the second derivative test and use of Assumption \ref{ass:mu} that $\lai\leq\Sigma_{\max},~i\in[p].$ Uniqueness of the solution follows directly from strong convexity. \ct{Strictly speaking we might need to also argue existence, i.e., feasibility of the AO. An indirect way is to show feasibility using the CGMT, but it  seems unnecessarily complicated?}

\vspace{5pt}
\subsubsection{Proof of (ii).}Using Lagrangian formulation, the solution $\wha_n$ to \eqref{eq:AO_con} is the same as the solution to the following:
\begin{align}
\left(\wha_n,u_n\right) :=\arg\min_{\w\in\Bc}\max_{u\geq 0} ~\frac{1}{2}\tn{\Sigmab^{-1/2}\w+\betas}^2 + u \left( \sqrt{\tn{\w}^2+\sigma^2} \tn{\gba} - \sqrt{\kappa}\,\hba^T\w + \frac{\sigma h}{\sqrt{n}} \right)\label{eq:AO_2}
\end{align}
where we have: (i) set $\gba := \g/\sqrt{n}$ and $\hba:= \h/\sqrt{p}$; (ii) recalled that $p/n=\kappa$; and, (iii) used $\left(\wha_n,u_n\right)$ to denote the optimal solutions in \eqref{eq:AO_2}. The subscript $n$ emphasizes the dependence of $\left(\wha_n,u_n\right)$ on the problem dimensions. Also note that (even though not explicit in the notation) $\left(\wha_n,u_n\right)$ are random variables depending on the realizations of $\gba,\hba$ and $h$.

 Notice that the objective function above is convex in $\w$ and linear (thus, concave) in $u$. Also, $\Bc$ is compact. Thus, strong duality holds and we can flip the order of min-max \cite{fan1953minimax}. Moreover, in order to make the objective easy to optimize with respect to $\w$, we use the following variational expression for the square-root term $\sqrt{\tn{\w}^2+\sigma^2}$:
$$
\tn{\gba}\sqrt{\tn{\w}^2+\sigma^2} = \tn{\gba}\cdot\min_{\tau\in[\sigma,\sqrt{\sigma^2+B_+^2}]} \left\{ \frac{\tau}{2} + \frac{\tn{\w}^2+\sigma^2}{2\tau} \right\} = \min_{\tau\in[\sigma,\sqrt{\sigma^2+B_+^2}]} \left\{ \frac{\tau\tn{\gba}^2}{2} + \frac{\sigma^2}{2\tau} + \frac{\tn{\w}^2}{2\tau} \right\},
$$
where $B_+$ is defined in Lemma \ref{lem:bd_PO}. For convenience define the constraint set for the variable $\tau$ as $\Tc':=[\sigma,\sqrt{\sigma^2+B_+^2}]$. For reasons to be made clear later in the proof (see proof of statement (iii)), we consider the (possibly larger) set:
\[
\Tc:=[\sigma,\max\{\sqrt{\sigma^2+B_+^2},2\tau_*\}]\,
\]
where $\tau_*$ is as in the statement of the lemma.

The above lead to the following equivalent formulation of \eqref{eq:AO_2}, 
\begin{align}
\left(\wha_n,u_n,\tau_n\right) = \max_{u\geq 0}\min_{\w\in\Bc,\tau\in\Tc} ~ \frac{u\tau\tn{\gba}^2}{2} + \frac{u\sigma^2}{2\tau} + \frac{u\sigma h}{\sqrt{n}} + \min_{\w\in\Bc} \left\{ \frac{1}{2}\tn{\Sigmab^{-1/2}\w+\betas}^2 + \frac{u}{2\tau}\tn{\w}^2 - u \sqrt{\kappa}\,{\hba}^T\w \right\} .\label{eq:AO_3}
\end{align}
%\ct{To be fully rigorous, need to show here that the unconstrained min over $\w$ is the same as the constrained $\w\in\Bc$.}
The minimization over $\w$ is easy as it involves a strongly convex quadratic function. First, note that the unconstrained optimal $\w':=\w'(\tau,u)$ (for fixed $(\tau,u)$) is given by
\begin{align}\label{eq:w'}
\w':=\w'(\tau,u) = -\left(\Sigmab^{-1}+\frac{u}{\tau}\Iden\right)^{-1}\left(\Sigmab^{-1/2}\betas-u\sqrt{\kappa}\hba\right), 
\end{align}
and \eqref{eq:AO_3} simplifies to
\begin{align}
\left(u_n,\tau_n\right)=\max_{u\geq 0}\min_{\tau\in\Tc} ~ \frac{u\tau\tn{\gba}^2}{2} + \frac{u\sigma^2}{2\tau} + \frac{u\sigma h}{\sqrt{n}} - \frac{1}{2} \left(\Sigmab^{-1/2}\betas-u\sqrt{\kappa}\hba\right)^T\left(\Sigmab^{-1}+\frac{u}{\tau}\Iden\right)^{-1} \left(\Sigmab^{-1/2}\betas-u\sqrt{\kappa}\hba\right)\,=:\Rc(u,\tau) .\label{eq:AO_4}
\end{align}
%\somm{Does $u=0$ scenario create a problem in saddle point uniqueness?} 
It can be checked by direct differentiation and the second-derivative test that the objective function in \eqref{eq:AO_4} is strictly convex in $\tau$ and strictly concave in $u$ over the domain $\{(u,\tau)\in\R_+\times\R_+\}$ \footnote{{To analyze the matrix-vector product term in \eqref{eq:AO_4} for $(\tau,u)$ one can use the fact that $\bSi$ is diagonal. This way, as a function of $u$ and $\tau$ the analysis reduces to the properties of relatively simple functions. For instance, for $\tau$, this function is in the form $f(\tau)=-(a+b/\tau)^{-1}$ for $a,b>0$, which is strictly convex.}}. Thus, the saddle point $(u_n,\tau_n)$ is unique. Specifically, this implies that the optimal $\wha_n$ in \eqref{eq:AO_3} is given by (cf. \eqref{eq:w'})
\begin{align}\label{eq:w_n}
\wha_n=\w'(\tau_n,u_n) = -\left(\Sigmab^{-1}+\frac{u_n}{\tau_n}\Iden\right)^{-1}\left(\Sigmab^{-1/2}\betas-u_n\sqrt{\kappa}\hba\right).
\end{align}
In Lemma \ref{lem:AO}(v) we will prove that wpa 1, in the limit of $p\rightarrow\infty$, $\|\wha_n\|_2\leq C$ for sufficiently large absolute constant $C>0$. Thus, by choosing the upper bound in the definition of $\Bc$ in Lemma \ref{lem:bd_PO} strictly larger than C, guarantees that the unconstrained $\wha_n$ in \eqref{eq:w_n} is feasible in \eqref{eq:AO_3}.

\noindent{\textbf{Asymptotic limit of the key quantities $\tau_n,u_n$:}} In what follows, we characterize the high-dimensional limit of the optimal pair $(u_n,\tau_n)$ in the limit $n,p\rightarrow\infty,~p/n\rightarrow\kappa$.
%Now that we have reached a min-max problem over only scalar variables, we are ready to study its convergence properties in the limit $n,p\rightarrow\infty,~p/n\rightarrow\kappa$. 
We start by analyzing the (point-wise) convergence of $\Rc(u,\tau)$. 
For the first three summands in \eqref{eq:AO_4}, we easily find that
$$
\left\{\frac{u\tau\tn{\gba}^2}{2} + \frac{u\sigma^2}{2\tau} + \frac{u\sigma h}{\sqrt{n}}\right\}~ \rP~\left\{
\frac{u\tau}{2} + \frac{u\sigma^2}{2\tau} \right\}.
$$
Next, we study the fourth summand. First, note that
\begin{align}
(u\sqrt{\kappa}\hba)^T\left(\Sigmab^{-1}+\frac{u}{\tau}\Iden\right)^{-1}(u\sqrt{\kappa}\hba) &= u^2\kappa\,\frac{1}{p}\h^T\left(\Sigmab^{-1}+\frac{u}{\tau}\Iden\right)^{-1}\h \nn\\
&= u^2\kappa\,\frac{1}{p}\sum_{i=1}^{p}\frac{\h_i^2}{\lai^{-1}+\frac{u}{\tau}} \nn\\
&\rP u^2\kappa\,\E\left[ \frac{1}{\Lambda^{-1}+\frac{u}{\tau}} \right].
\end{align}
% the quadratic form
%\begin{align}
%(u\sqrt{\kappa}\hba)^T\left(\Sigmab^{-1}+\frac{u}{\tau}\Iden\right)^{-1}(u\sqrt{\kappa}\hba) = u^2\kappa\,\frac{1}{p}\h^T\left(\Sigmab^{-1}+\frac{u}{\tau}\Iden\right)^{-1}\h
%\end{align}
%concentrates around its expectation \cite{?} $u^2\kappa\,\tr\left(\left(\Sigmab^{-1}+\frac{u}{\tau}\Iden\right)\right)^{-1}.$

%\somm{Explain/justify PL(2)'ness of these functions. Clearly explain how A2 is used\dots}
In the last line, $\Lambda$ is a random variable as in Definition \ref{def:Xi}. {Also, we used Assumption \ref{ass:mu} together with the facts that $\h$ is independent of $\Sigmab$ and  that the function $(x_1,x_2)\mapsto x_1^2(x_2^{-1}+u/\tau)^{-1}$ is $\rm{PL}(3)$ assuming $x_2$ is bounded (see Lemma \ref{lem:PLbdd} for proof).}
\ct{Question: Is it immediate that the empirical distribution of $(\beta,\Sigmab,\h)$ converges in $W_k$ to $\mu\otimes\Nn(0,1)$ given that $(\beta,\Sigmab)$ converges to $\mu$ and $\h$ is independent???}
 Second, we find that
\begin{align}
(\betas)^T\Sigmab^{-1/2}\left(\Sigmab^{-1}+\frac{u}{\tau}\Iden\right)^{-1}\Sigmab^{-1/2}\betas &= \frac{1}{p}(\sqrt{p}\betas)^T\left(\Iden+\frac{u}{\tau}\Sigmab\right)^{-1}(\sqrt{p}\betas) \nn\\
&= \frac{1}{p}\sum_{i=1}^{p}\frac{\left(\sqrt{p}\betas_i/\sqrt{\lai}\right)^2}{\lai^{-1}+\frac{u}{\tau}}\nn\\
&\rP \E\left[\frac{B^2\Lambda^{-1}}{\Lambda^{-1}+\frac{u}{\tau}}\right].
\end{align}
Here, $\Lambda,B$ are random variables as in Definition \ref{def:Xi} and {we also used Assumption \ref{ass:mu} together with the fact that the function $(x_1,x_2)\mapsto x_1^2x_2^{-1}(x_2^{-1}+u/\tau)^{-1}$ is $\rm{PL}(3)$ assuming $x_2$ is bounded (see Lemma \ref{lem:PLbdd} for proof).}
Third, \cts{by independence of $(\betas, \Sigmab)$ from $\h$}
\begin{align}
(u\sqrt{\kappa}\hba)^T\left(\Sigmab^{-1}+\frac{u}{\tau}\Iden\right)^{-1}\Sigmab^{-1/2}\betas = u\sqrt{\kappa} \cdot \frac{1}{p}\sum_{i=1}
^p \frac{\h_i\Sigmab_{i,i}^{-1/2}(\sqrt{p}\betas_i)}{\Sigmab_{i,i}^{-1}+\frac{u}{\tau}\Iden} ~\rP~ 0.
\end{align}
Putting these together, the objective $\Rc(u,\tau)$ in \eqref{eq:AO_4} converges point-wise in $u,\tau$ to
\begin{align}
\Rc(u,\tau)\rP\Dc(u,\tau) := \frac{1}{2}\left({u\tau} + \frac{u\sigma^2}{\tau} - u^2\kappa\,\E\left[ \frac{1}{\Lambda^{-1}+\frac{u}{\tau}} \right] - \E\left[\frac{B^2\Lambda^{-1}}{\Lambda^{-1}+\frac{u}{\tau}}\right] \right).\label{eq:conv_pt}
\end{align}
Note that $\Rc(u,\tau)$ (and thus, $\Dc(u,\tau)$) is convex in $\tau$ and concave in $u$. Thus, the convergence in \eqref{eq:conv_pt} is in fact uniform (e.g., \cite{AG1982}) and we can conclude that 
\begin{align}\label{eq:Dc0}
\phi(\g,\h) \rP \max_{u\geq 0}\min_{\tau\in\Tc}~ \Dc(u,\tau).
\end{align}
and {using strict concave/convexity of $\Dc(u,\tau)$, we also have the parameter convergence \cite[Lem. 7.75]{NF36}}
%\somm{This line follows from strict convex/concavity right?}
\begin{align}\label{eq:Dc}
{(u_n,\tau_n) \rP (u_*,\tau_*):=\arg\max_{u\geq 0}\min_{\tau\in\Tc}~ \Dc(u,\tau).}
\end{align}
In the proof of statement (iii) below, we show that the saddle point of \eqref{eq:Dc0} is $(u_*,\tau_*)$. In particular, $\tau_*$ is strictly in the interior of $\Tc$, which combined with convexity implies that
$$
 \max_{u\geq 0}\min_{\tau\in\Tc}~ \Dc(u,\tau) =  \max_{u\geq 0}\min_{\tau>0}~ \Dc(u,\tau) =: \bar\phi.
$$
This, together with the first display above proves the second statement of the lemma.

\vspace{5pt}
\subsubsection{Proof of (iii).}  Next, we compute the saddle point $(u_*,\tau_*)$ by studying the first-order optimality conditions of the strictly concave-convex $\Dc(u,\tau)$. Specifically, we consider the unconstrained minimization over $\tau$ and we will show that the minimum is achieved in the strict interior of $\Tc$. Direct differentiation of $\Dc(u,\tau)$ gives
\begin{subequations}
\begin{align}
{\tau} + \frac{\sigma^2}{\tau} - 2u\kappa\E\left[ \frac{1}{\Lambda^{-1}+\frac{u}{\tau}} \right] + \frac{u^2}{\tau}\kappa \E\left[ \frac{1}{\left(\Lambda^{-1}+\frac{u}{\tau}\right)^2}\right] + \frac{1}{\tau}\E\left[\frac{B^2\Lambda^{-1}}{\left(\Lambda^{-1}+\frac{u}{\tau}\right)^2}\right] &= 0, \label{eq:fo1}\\
{u} - \frac{u\sigma^2}{\tau^2} - \frac{u^3}{\tau^2}\kappa \E\left[ \frac{1}{\left(\Lambda^{-1} + \frac{u}{\tau}\right)^2}\right] - \frac{u}{\tau^2} \E\left[\frac{B^2\Lambda^{-1}}{\left(\Lambda^{-1}+\frac{u}{\tau}\right)^2}\right] &= 0,\label{eq:fo2}
\end{align}
\end{subequations}
Multiplying \eqref{eq:fo2}  with $\frac{\tau}{u}$ and adding to \eqref{eq:fo1} results in the following equation
\begin{align}
\tau = u\kappa\E\left[ \frac{1}{\Lambda^{-1}+\frac{u}{\tau}} \right] ~\Leftrightarrow ~
\E\left[ \frac{1}{(\frac{u}{\tau}\Lambda)^{-1}+1} \right] = \frac{1}{\kappa} \label{eq:tauu}\,.
\end{align}
Thus, we have found that the ratio $\frac{u_*}{\tau_*}$ is the unique solution to the equation in \eqref{eq:tauu}. Note that this coincides with the Equation \eqref{eq:ksi} that defines the parameter $\xi$ in Definition \ref{def:Xi}.  The fact that \eqref{eq:tauu} has a unique solution for all $\kappa>1$ can be easily seen as $F(x)=\E\left[ \frac{1}{(x\Lambda)^{-1}+1} \right], x\in\R_+$ has range $(0,1)$ and is strictly increasing (by differentiation).

Thus, we call $\xi=\frac{u_*}{\tau_*}$. Moreover, multiplying \eqref{eq:fo2} with $u$ leads to the following equation for $\tau_*$:
\begin{align}
u_*^2 = \sigma^2\xi^2 + u_*^2 \xi^2  \kappa \E\left[ \frac{1}{\left(\Lambda^{-1} +\xi\right)^2}\right] + \xi^2  \E\left[ \frac{B^2\Lambda^{-1}}{\left(\Lambda^{-1} + \xi\right)^2}\right] ~\Rightarrow~ \tau_*^2 = \frac{\sigma^2 + \E\left[ \frac{B^2\Lambda^{-1}}{\left(\Lambda^{-1} + \xi\right)^2}\right]}{1-\xi^2\kappa \E\left[ \frac{1}{\left(\Lambda^{-1} +\xi\right)^2}\right]} =  \frac{\sigma^2 + \E\left[ \frac{B^2\Lambda^{-1}}{\left(\Lambda^{-1} + \xi\right)^2}\right]}{1-\kappa \E\left[ \frac{1}{\left((\xi\Lambda)^{-1} +1\right)^2}\right]}.
\end{align}
{Again, note that this coincides with Equation \eqref{eq:gamma} that determines the parameter $\gamma$ in Definition \ref{def:Xi}, i.e., $\tau_*^2 = \gamma.$ By definition of $\Tc$ and of $\tau_*$, it is clear that $\tau_\star$ is in the strict interior of $\Tc$.}

\vspace{5pt}
\subsubsection{Proof of (iv).}  For convenience, define 
$$F_n(\btha,\betas,\Sigmab):=\frac{1}{p} \sum_{i=1}^{p} f\left(\sqrt{p}\btha_{n,i},\sqrt{p}\betas_{i},\Sigmab_{ii}\right)\quad\text{and}\quad\alpha_*:=\E_\mu\left[f(X_{\kappa,\sigma^2}(\Lambda,B,H),B,\Lambda)\right].$$
Recall from \eqref{eq:w_n} the explicit expression for $\wha_n$, repeated here for convenience.
\begin{align}
\wha_n = -\left(\Sigmab^{-1}+\frac{u_n}{\tau_n}\Iden\right)^{-1}\left(\Sigmab^{-1/2}\betas-u_n\sqrt{\kappa}\hba\right).\nn
\end{align}
Also, recall that $\btha_n = \Sigmab^{-1/2}\wha_n+\betas$. Thus,  (and using the fact that $\hba$ is distributed as $-\hba$),
\begin{align}
\btha_n &=\Sigmab^{-1/2}\left(\Sigmab^{-1}+\frac{u_n}{\tau_n}\Iden\right)^{-1}u_n\sqrt{\kappa}\hba + \left(\Iden-\Sigmab^{-1/2}\left(\Sigmab^{-1}+\frac{u_n}{\tau_n}\Iden\right)^{-1}\Sigmab^{-1/2}\right)\betas\nn\\
\Longrightarrow\btha_{n,i}&=\frac{{\Sigmab_{i,i}^{-1/2}}}{1+(\ksi_n\Sigmab_{i,i})^{-1}}\sqrt{\kappa}\tau_n\hba_i +\left(1-\frac{1}{1+\ksi_n\Sigmab_{i,i}}\right)\betas_i.\label{eq:betan}
\end{align}
For $i\in[p]$, define 
\begin{align}
\vb_{n,i} = \frac{{\Sigmab_{i,i}^{-1/2}}}{1+(\ksi_*\Sigmab_{i,i})^{-1}}\sqrt{\kappa}\tau_*\hba_i +\left(1-\frac{1}{1+\ksi_*\Sigmab_{i,i}}\right)\betas_i
%-\left(\Sigmab^{-1}+\frac{u_*}{\tau_*}\Iden\right)^{-1}\left(\Sigmab^{-1/2}\betas-u_*\sqrt{\kappa}\hba\right).
\end{align}
In the above, for convenience, we have denoted $\xi_n:=u_n/\tau_n$ and recall that $\xi_*:=u_*/\tau_*$.

The proof proceeds in two steps. In the first step, we use the fact that $\ksi_n\rP\ksi_*$ and $u_n\rP u_\star$ (see \eqref{eq:Dc}) to prove that for any $\eps\in(0,\ksi_*/2)$, there exists an absolute constant $C>0$ such that wpa 1:
\begin{align}\label{eq:ivstep1}
|F_n(\sqrt{p}\btha_n,\sqrt{p}\betas,\Sigmab) - F_n(\sqrt{p}\vb_n,\sqrt{p}\betas,\Sigmab) | \leq C\eps.
\end{align}
In the second step, we use pseudo-Lipschitzness of $f$ and Assumption \ref{ass:mu} to prove that
\begin{align}
|F_n(\vb_n,\betas,\Sigmab)| \rP \alpha_*.\label{eq:ivstep2}
\end{align}
The desired follows by combining \eqref{eq:ivstep1} and \eqref{eq:ivstep2}. Thus, in what follows, we prove \eqref{eq:ivstep1} and \eqref{eq:ivstep2}.

\vspace{2pt}
\noindent\underline{Proof of \eqref{eq:ivstep1}.}~~Fix some $\eps\in(0,\ksi_*/2)$. From \eqref{eq:Dc}, we know that w.p.a. 1 $|\xi_n-\xi_*|\leq \eps$ and $|u_n-u_*|\leq \eps$. Thus, $\wha_n$ is close to $\vb_n$. Specifically, in this event, for every $i\in[p]$, it holds that:
\begin{align}
\nn|\btha_{n,i} - \vb_{n,i}| &\leq {|\betas_i|}\left|\frac{1}{1+\xi_n\lai}-\frac{1}{1+\xi_*\lai}\right| + \sqrt{\kappa}\frac{|\hba_i|}{\sqrt{\Sigmab_{i,i}}}\left|\frac{\tau_n}{1+(\xi_n\lai)^{-1}}-\frac{\tau_*}{1+(\xi_*\lai)^{-1}}\right| \\
\nn&= {|\betas_i|}\left|\frac{1}{1+\xi_n\lai}-\frac{1}{1+\xi_*\lai}\right| + \sqrt{\kappa}\frac{|\hba_i|}{\sqrt{\Sigmab_{i,i}}}\left|\frac{u_n}{\xi_n+\lai^{-1}}-\frac{u_*}{\xi_*+\lai^{-1}}\right| \\
\nn& \leq
{|\betas_i|}\frac{|\lai||\xi_n-\xi_*|}{|1+\xi_n\lai||1+\xi_*\lai|} + \sqrt{\kappa}\frac{|\hba_i|}{\sqrt{\lai}}  \frac{u_*|\ksi_n-\ksi_*|}{(\ksi_n+\lai^{-1})(\ksi_*+\lai^{-1})} + 
\sqrt{\kappa}\frac{|\hba_i|}{\sqrt{\lai}}\frac{|u_n-u_*|}{\xi_n+\lai^{-1}}
\\
\nn& \leq
{|\betas_i|}{\Sigma_{\max}\eps} + \sqrt{\kappa}{|\hba_i|} u_*\Sigma_{\max}^{3/2} \eps+ 
\sqrt{\kappa}|\hba_i|\Sigma_{\max}^{1/2}\eps
\\
&\leq C\eps  \left(|\hba_i| + |\betas_i|\right).\label{eq:betav}
%\cdot \max\left\{\Sigma_{\max}^{3/2},\Sigma_{\max}^{1/2}\right\}\,
%\nn& \leq
%\frac{|\betas_i|}{\sqrt{\lai}} \frac{\eps}{(\xi-\eps)\xi} + \sqrt{\kappa}|\hba_i|\left(\frac{\eps(u_*+\eps)}{(\xi-\eps)\xi} + \frac{\eps}{\xi}\right)\\
%& \leq
%\frac{|\betas_i|}{\sqrt{\lai}} c_1(\eps) + |\hba_i|c_2(\eps).\label{eq:wivi}
\end{align}%\som{$\eps$ missing}
where $C=C(\Sigma_{\max},\kappa,u_*)$ is an absolute constant. In the second line above, we recalled that $u_n=\tau_n\xi_n$ and $u_*=\tau_*\xi_*$. In the third line, we used the triangle inequality. In the fourth line, we used that $\ksi_*>0$, $0<\lai\leq\Sigma_{\max}$ and $\ksi_n\geq \ksi_*-\eps \geq \ksi_*/2 >0$. 
%In the second line above, we used the triangle inequality. In the third inequality we used Assumption \ref{ass:inv} that $\lai>0, i\in[p]$, and in the last line we defined appropriate constants $c_1(\eps),c_2(\eps)>0$. Importantly,  ... 

Now, we will use this and Lipschitzness of $f$ to argue that there exists absolute constant $C>0$ such that wpa 1,
\begin{align}
|F_n(\sqrt{p}\btha_n,\sqrt{p}\betas,\Sigmab) - F_n(\sqrt{p}\vb_n,\sqrt{p}\betas,\Sigmab) | \leq C\eps.\nn
\end{align}

Denote, $\ab_i=(\sqrt{p}\btha_{n,i},\sqrt{p}\betas_i,\Sigmab_{i,i})$ and $\bb_i=(\sqrt{p}\vb_{n,i},\sqrt{p}\betas_i,\Sigmab_{i,i})$. Following the exact same argument as in \eqref{eq:dev2show} (just substitute $\bt\leftrightarrow\vb_n$ in the derivation), we have that for some absolute constant $C>0$ wpa 1:
\begin{align}
|F_n(\btha_n(\g,\h),\betas,\Sigmab) - F_n(\vb_n,\betas,\Sigmab)| &\leq C \|\btha_n-\vb_n\|_2.
\end{align}
%\begin{align}
%|F_n(\btha_n(\g,\h),\betas,\Sigmab) - F_n(\vb_n,\betas,\Sigmab)| &\leq \frac{L}{p}\sum_{i=1}^p (1+\max\{\|\ab_i\|_2^{k-1},\|\bb_i\|_2^{k-1}\})\|\ab_i-\bb_i\|_2\nn \\
%&\leq \frac{L}{p}\sum_{i=1}^p \left(1+\max\{\|\ab_i\|_2^{k-1},\|\bb_i\|_2^{k-1}\}\right)\cdot|\sqrt{p}\btha_{n,i}(\g,\h) - \sqrt{p}\vb_{n,i}|\nn\\
%&\leq L\left(1+\max\{\frac{1}{p}\sum_{i=1}^p \|\ab_i\|_2^{2k-2},\frac{1}{p}\sum_{i=1}^p\|\bb_i\|_2^{2k-2}\} \right)^{1/2}{\|\btha_n(\g,\h) - \vb_n\|_2}\nn.
%\end{align}
%\max\left\{\Sigma_{\max}^{3/2},\Sigma_{\max}^{1/2}\right\}
From this and \eqref{eq:betav}, we find that
\begin{align}
|F_n(\btha_n(\g,\h),\betas,\Sigmab) - F_n(\vb_n,\betas,\Sigmab)|
&\leq  
C\eps \left(\sum_{i=1}^p \left(|\hba_i|+|\betas_i|\right)^2\right)^{1/2}\nn
\\
 &\leq 
C\eps\sqrt{2}\sqrt{\|\betas\|_2^2 + \|\hba\|_2^2}.\label{eq:epsS}
\end{align}
%As in \eqref{eq:bdmom}, it can be shown that wpa 1: $\frac{1}{p}\sum_{i=1}^p \|\ab_i\|_2^{2k-2}<\infty$ and $\frac{1}{p}\sum_{i=1}^p \|\bb_i\|_2^{2k-2}<\infty$, as $p\rightarrow\infty$. 
But, \cts{recall that $\|\betas\|_2^2 = \frac{1}{p}\sum_{i=1}^p(\sqrt{p}\betas_i)^2<\infty$, as $p\rightarrow \infty$ by Assumption \ref{ass:mu}.}
%{\color{red} by assumption on second moment convergence of $\sqrt{p}\betas$.} 
Also, since $\hba_i\sim\Nn(0,1/p)$, it holds that $\|\hba\|_2^2\leq 2$, wpa 1 as $p\rightarrow\infty$. Therefore, from \eqref{eq:epsS}, wpa 1, there exists constant $C>0$ such that 
\begin{align}
|F_n(\btha_n(\g,\h),\betas,\Sigmab) - F_n(\vb_n,\betas,\Sigmab)| \leq C \cdot\eps,\nn
\end{align}
as desired.

\vspace{2pt}
\noindent\underline{Proof of \eqref{eq:ivstep2}.}~~Next, we will use Assumption \ref{ass:mu} to show that 
\begin{align}
|F_n(\vb_n,\betas,\Sigmab)| \rP \alpha_*.\label{eq:AO_conv}
\end{align}
Notice that $\vb_n$ is a function of $\betas,\Sigmab,\hba$. Concretely, define ${\ggt}:\R^3\rightarrow\R$, such that
$$
\ggt(x_1,x_2,x_3) := \frac{x_2^{-1/2}}{1+(\ksi_* x_2)^{-1}}\sqrt{\kappa}\tau_* x_3 + (1-(1+\ksi_*x_2)^{-1})x_1,
$$
and notice that
$$
\sqrt{p}\vb_{n,i} = \ggt\left(\sqrt{p}\betas_i,\lai,\h_i\right) = \frac{{\Sigmab_{i,i}^{-1/2}}}{1+(\ksi_*\Sigmab_{i,i})^{-1}}\sqrt{\kappa}\tau_*\h_i +\left(1-\frac{1}{1+\ksi_*\Sigmab_{i,i}}\right)\sqrt{p}\betas_i.
$$
Thus,
$$
F_n(\vb_n,\betas,\Sigmab) = \frac{1}{p}\sum_{i=1}^p f\left(g\left(\sqrt{p}\betas_i,\lai,\h_i\right),\sqrt{p}\betas_i,\lai\right) =: \frac{1}{p}\sum_{i=1}^p h\left(\h_i,\sqrt{p}\betas_i,\lai\right),
$$
where we have defined $h:\R^3\rightarrow\R$:
\begin{align}
h(x_1,x_2,x_3) := f\left(\ggt(x_2,x_3,x_1),x_2,x_3\right).\label{h func}
\end{align}
\cts{We will prove that $h\in\rm{PL}(4)$. Indeed, if that were the case, then Assumption \ref{ass:mu} gives}
\begin{align}
\frac{1}{p}\sum_{i=1}^p h\left(\h_i,\sqrt{p}\betas_i,\lai\right) \rP \E_{\Nn(0,1)\otimes \mu}\left[h(H,B,\Lambda)\right] &=  \E_{\Nn(0,1)\otimes \mu}\left[f\left(\ggt(B,\Lambda,H),B,\Lambda\right))\right]\\
& = \E[f\left(X_{\kappa,\sigma^2}(\Lambda,B,H),B,\Lambda\right)] = \alpha_*,
\end{align}
where the penultimate equality follows by recognizing that (cf. Eqn. \eqref{eq:X})
$$
\ggt(B,\Lambda,H) = (1-(1+\ksi_*\Lambda)^{-1})B + \sqrt{\kappa}\frac{\tau_*\Lambda^{-1/2}}{1+(\ksi_*\Lambda)^{-1}}H =  X_{\kappa,\sigma^2}(\Lambda,B,H).
$$
It remains to show that $h\in\rm{PL}(4)$. Lemma \ref{lem:hPL} in Section \ref{SM useful fact} shows that if $f\in\rm{PL}(k)$, then $h\in\rm{PL}(k+1)$ for all integers $k\geq 2$. \fy{Using this and the fact that $\Plt\subset\rm{PL}(3)$, for any $f\in\Plt$, we find that $h\in \rm{PL}(4)$. This completes the proof of \eqref{eq:ivstep2}.}
% First, consider the case $f\in\rm{PL}(2)$. Then,  $h\in\rm{PL}(3)$; thus, also $h\in\rm{PL}(4)$. Second, consider the case $f=\fl\in\Fl$. Then, we prove in Lemma \ref{lem:fL}  that $\fl\in\rm{PL}(3)$ i.e.~$\Fl\subset\rm{PL}(3)$. Thus, the desired holds in this case too.  

%In the remaining of the proof, \cts{we show that $h\in\rm{PL}(k)$} which is formalized below.

\vspace{5pt}
\subsubsection{Proof of (v):} Let $\psi:\R\rightarrow\R$ be any bounded Lipschitz function. The function $f(a,b,c) = \psi(a)$ is trivially $\rm{PL}$(2). Thus, by directly applying statement (iv) of the lemma, we find that
$$
\frac{1}{p}\sum_{i=1}^p{\psi(\sqrt{p}\btha_{n,i}(\g,\h))} \rP \E\left[\psi(X_{\kappa,\sigma^2})\right].
$$
Since this holds for any bounded Lipschitz function, we have shown that the empirical distribution of $\btha_n$ converges weakly to the distribution of $X_{\kappa,\sigma^2}$. It remains to prove boundedness of the $2$nd moment as advertised in \eqref{eq:k_AO}. Recall  from
\eqref{eq:betan} that
\begin{align}
\sqrt{p}\btha_{n,i}&=\frac{{\Sigmab_{i,i}^{-1/2}}}{1+(\ksi_n\Sigmab_{i,i})^{-1}}\sqrt{\kappa}\tau_n\h_i +\left(1-\frac{1}{1+\ksi_n\Sigmab_{i,i}}\right)(\sqrt{p}\betas_i).\nn
\end{align}
Using this, boundedness of $\Sigmab_{i,i}$ from Assumption \ref{ass:mu}, and the fact that $\tau_n\rP\tau_\star, \ksi_n\rP\ksi_\star$, there exists constant $C=C(\Sigma_{\max},\Sigma_{\min},k,\tau_\star,\ksi_\star)$ such that wpa 1,
\begin{align}
\frac{1}{p}\sum_{i=1}^p|\sqrt{p}\btha_{n,i}|^{2} \leq C\left(\frac{1}{p}\sum_{i=1}^p|\h_{i}|^{2}+\frac{1}{p}\sum_{i=1}^p|\sqrt{p}\betas_{i}|^{2}\right).\nn
\end{align}
But the two summands in the expression above are finite in the limit of $p\rightarrow\infty$. Specifically, (i) from Assumption \ref{ass:mu}, $\frac{1}{p}\sum_{i=1}^p|\sqrt{p}\betas_{i}|^{2}\rP\E[B^{2}]<\infty$; (ii) $\frac{1}{p}\sum_{i=1}^p|\h_{i}|^{2}\rP\E[H^{2}]=1$, using the facts that $\h_i\simiid\Nn(0,1)$ and $H\sim\Nn(0,1)$. This proves \eqref{eq:k_AO}, as desired.

\section{Asymptotic formulas on Magnitude- and Hessian- pruning}\label{sec prune risk}
Here, we use Theorem \ref{thm:master_W2} to characterize the risk of the magnitude- and Hessian- pruned solutions. This section supplements the discussion in Section \ref{sec:risk}. For completeness, first, we recall the magnitude-pruning results of Section \ref{sec:risk} and restate Corollary \ref{cor:mag} below. This corollary characterizes the performance of magnitude pruning. Following this, we shall further discuss Hessian pruning.

\subsection{Magnitude-based pruning}

We begin with the following necessary definitions.  Define the hard-thresholding function with fixed threshold $t\in\R_+$ as follows:
\begin{align}\label{eq:threshold_app}
\Tc_t(x) = \begin{cases} x & \text{if } |x|>t \\ 0 & \text{otherwise} \end{cases}.
\end{align}
Further, {given model sparsity target $1>\alpha>0$}, define the threshold $t^\star$ as follows:
\begin{align}\label{eq:tstar_app}
t^\star:=\sup\left\{t\in\R\,:\, \Pr(|X_{\kappa,\sigma^2}|\geq t) \geq \alpha \right\}.
\end{align}

\cormag*

The proof of the corollary above, is given in Section \ref{sec:risk}. Below, we extend the results to Hessian-based pruning.

\subsection{Hessian-based pruning}
Let $\bth$ be the min-norm solution in \eqref{eq:min_norm}. Recall that the Hessian-pruned model (via Optimal Brain Damage) $\betab_s^H$ at sparsity $s$ is given by
\begin{align}\label{eq:hp}
\betab_s^H = \hat\Sigmab^{-1/2}\Tb_s(\hat\Sigmab^{1/2}\betab),
\end{align}
where $\hat\Sigmab=\diag{\X^T\X}/n$ the diagonal of the empirical covariance matrix. 

We will argue that the following formula characterizes the asymptotic risk of the Hessian pruning solution. 
Recall the notation in \eqref{eq:threshold_app} and define
\begin{align}\label{eq:tdiam_app}
t^\diamond:=\sup\left\{t\in\R\,:\, \Pr(|\Lambda^{1/2}X_{\kappa,\sigma^2}|\geq t) \geq \alpha \right\}.
\end{align}
The risk of the Hessian-pruned model satisfies the following in the limit of $n,p,s\rightarrow\infty$ at rates $\kappa:=p/n>1$ and $\alpha:=s/p\in(0,1)$ (cf. Assumption \ref{ass:linear}):
\begin{align}\label{eq:hessian_risk}
\Lc(\bth^H_s) \rP \sigma^2 + \E\left[ \left(\Lambda^{1/2}B-\Tc_{t^\diamond}(\Lambda^{1/2}X_{\kappa,\sigma^2})\right)^2 \right],
\end{align}
where the expectation is over $(\Lambda,B,H)\sim\mu\otimes\Nn(0,1)$. In our LGP experiments, we used this formula \eqref{eq:hessian_risk} to accurately predict the Hessian-based pruning performance. 
%\begin{corollary}[Risk of Hessian-pruning]\label{cor:hes}
%Let the same assumptions and notation as in the statement of Theorem \ref{thm:master_W2} hold. Specifically, let $\hat\betab$ be the min-norm solution in \eqref{eq:min_norm} and $\hat\beta_s^H$ the Hessian-pruned model at sparsity $s$. Recall the notation in \eqref{eq:threshold_app} and define
%\begin{align}\label{eq:tdiam_app}
%t^\diamond:=\inf\left\{t\in\R\,:\, \Pr(|\Lambda^{1/2}X_{\kappa,\sigma^2}|\geq t) \geq \alpha \right\}.
%\end{align}
%The risk of the magnitude-pruned model satisfies the following in the limit of $n,p,s\rightarrow\infty$ at rates $\kappa:=p/n>1$ and $\alpha:=s/p\in(0,1)$ (cf. Assumption \ref{ass:linear}):
%\begin{align}
%\Lc(\bth^H_s) \rP \sigma^2 + \E\left[ \Lambda\left(B-\Lambda^{-1/2}\Tc_{t^\diamond}(\Lambda^{1/2}X_{\kappa,\sigma^2})\right) \right],\nn
%\end{align}
%where the expectation is over $(\Lambda,B,H)\sim\mu\otimes\Nn(0,1).$
%\end{corollary}
%\begin{proof}

Recall the definition of the hard-thresholding operator $\Tc_t(x)$. Similar to Section \ref{sec:risk}, we consider a threshold-based pruning vector $$\bth^{\Tc,H}_{t} := \hat\Sigmab^{-1/2}\Tc_{t/\sqrt{p}}(\hat\Sigmab^{1/2}\hat\bt),$$ where $\Tc_t$ acts component-wise. Further define 
$$\bth^{\Tc,H^\star}_{t} := \Sigmab^{-1/2}\Tc_{t/\sqrt{p}}(\Sigmab^{1/2}\hat\bt).$$
Note that $\bth^{\Tc,H^\star}_{t}$ uses the true (diagonal) covariance matrix $\Sigmab$ instead of its sample estimate $\hat\Sigmab$. For later reference, note here that $\hat\Sigmab$ concentrates (entry-wise) to $\Sigmab$. Using boundedness of $\Sigmab$ and standard concentration of sub-exponential random variables.

First, we compute the limiting risk of $\bth^{\Tc,H^\star}_{t}$. Then, we will use the fact that $\hat\Sigmab$ concentrates (entry-wise) to $\Sigmab$, to show that the risks of $\bth^{\Tc,H^\star}_{t}$ and $\bth^{\Tc,H}_{t}$ are arbitrarily close as $p\rightarrow\infty$.

%We start by computing the risk of $\bth^{\Tc,H^\star}_{t}$. 
Similar to \eqref{eq:loss_w2}, 
\begin{align}
\Lc(\bth^{\Tc,H^\star}_t) &= \sigma^2 + (\bt^\star-\bth^{\Tc,H^\star}_t)^T\Sigmab(\bt^\star-\bth^{\Tc,H^\star}_t) \nn \\
  &= \sigma^2 + \frac{1}{p}\sum_{i=1}^p\Sigmab_{i,i}\big(\sqrt{p}\betas_i-\lai^{-1/2}\Tc_{t}(\lai^{1/2}\sqrt{p}\hat\betab_i)\big)^2 \nn
    = \sigma^2 + \frac{1}{p}\sum_{i=1}^p\big(\lai^{1/2}\sqrt{p}\betas_i-\Tc_{t}(\lai^{1/2}\sqrt{p}\hat\betab_i)\big)^2 
      \\
    &= \sigma^2 + \frac{1}{p}\sum_{i=1}^p\big(\lai^{1/2}\sqrt{p}\betas_i-\Tc_{t}(\wh_i + \lai^{1/2}\sqrt{p}\betas_i\big)^2 \nn \\
        &=: \sigma^2 + \frac{1}{p}\sum_{i=1}^p\big(\sqrt{p}\blat_i-\Tc_{t}(\wh_i + \blat_i)\big)^2.\label{eq:hesa}
  %\nn\\
%  &\rP \sigma^2 + \E\left[ \Lambda\left(B-\Tc_t(X_{\kappa,\sigma^2})\right) \ri
\end{align}
In the second line above, we used that $\sqrt{p}\Tc_{t/\sqrt{p}}(x)=\Tc_{t}(\sqrt{p}x)$. In the third line, we recalled the change of variable in \eqref{eq:w}, i.e., $\wh$ is the solution to \eqref{eq:PO}. Finally, in the last line we defined $\blat:=\sqrt{\Sigmab}\betas$ (note that this is related to the saliency score $\bla$ defined in \eqref{saliency eq}).

To evaluate the limit of the empirical average in \eqref{eq:hesa}, we proceed as follows. First, we claim that the empirical distribution of $\sqrt{p}\blat$ converges weakly to the distribution of the random variable $B\sqrt{\Lambda}$, where $(\Lambda,B)\sim\mu$. \so{Note that this convergence is already implied by the proof of Theorem \ref{thm:master_W2} by setting the $g$ function to be zero in \eqref{eq:fdef}.}
%\somm{This part is repetitive, can be simplified.}
For an explicit proof, take any bounded Lipschitz test function $\psi:\R\rightarrow\R$ and call $\psi'(x,y):=\psi({\sqrt{x}}y)$. Then, 
\begin{align}
|\psi'(\lai,\sqrt{p}\betas_i)-\psi'(\lai',\sqrt{p}{\betas_i}')| &= |\psi(\sqrt{\lai}\sqrt{p}\betas_i)-\psi(\sqrt{\lai'}\sqrt{p}{\betas_i}')| \leq C |\sqrt{\lai}\sqrt{p}\betas_i - \sqrt{\lai'}\sqrt{p}{\betas_i}'| \nn
\\
&\leq
C |\sqrt{p}\betas_i - \sqrt{p}{\betas_i}'| + C'|\sqrt{p}{\betas_i}'| |\sqrt{\lai}- \sqrt{\lai'}|\nn  \\
&\leq
C (|\sqrt{p}\betas_i - \sqrt{p}{\betas_i}'| + |\sqrt{p}{\betas_i}'| |{\lai}- {\lai'}|)\nn  \\
&\leq
C (1 + \|[\sqrt{p}\betas_i,\lai]\|_2 +\|[\sqrt{p}{\betas_i}',\lai']\|_2 ) \sqrt{|\sqrt{p}\betas_i - \sqrt{p}{\betas_i}'|^2   + |{\lai}- {\lai'}|^2}\nn.
\end{align}
Thus, $\psi'$ is $\rm{PL(2)}$. Hence, from Assumption \ref{ass:mu},
\begin{align}
\frac{1}{p}\sum_{i=1}^p \psi(\sqrt{p}\blat_i) = \frac{1}{p}\sum_{i=1}^p \psi'(\lai,\sqrt{p}\betas_i) \rP  \E[\psi'(B,\sqrt{\Lambda})] =  \E[\psi(B\sqrt{\Lambda})]\label{eq:convvvv}
\end{align}
Besides, from Theorem \ref{thm:master_W2} applied for $f_\Lc(x,y,z)=zy^2$ (i.e., set $g$ the zero function in \eqref{eq:fdef}), we have that
$$
\frac{1}{p}\sum_{i=1}^p \sqrt{p}(\blat_i)^2 \rP  \E[B^2\Lambda].
$$
Therefore, convergence in \eqref{eq:convvvv} actually holds for any $\psi\in\rm{PL}(2)$. Next, observe that, $\wo$ of \eqref{eq:w_n} can be written in terms of $\blat$ via \begin{align}\label{eq:w_n2}
\w_n=\w'(\tau_n,u_n)& = -\left(\Iden+\frac{u_n}{\tau_n}\Sigmab\right)^{-1}\left(\Sigmab^{1/2}\betas-u_n\sqrt{\kappa}\Sigmab\hba\right)\\
&=-\left(\Iden+\frac{u_n}{\tau_n}\Sigmab\right)^{-1}\left(\blat-u_n\sqrt{\kappa}\Sigmab\hba\right).
\end{align}
After this observation, the convergence proof can be finalized by using a modified version of Assumption \ref{ass:mu} as follows.%in two ways. First approach is

{
\begin{assumption}[Empirical distribution for saliency] \label{ass:lambda} Set $\blat_i=\sqrt{\lai}\bts_i$. The joint empirical distribution of $\{(\lai,\blat_i)\}_{i\in[p]}$ converges in {Wasserstein-k} distance to a probability distribution $\mu=\mu(\Lambda,S)$ on $\R_{>0}\times\R$ for some \fx{$k\geq 4$}. That is
$
\frac{1}{p}\sum_{i\in[p]}\delta_{(\lai,\sqrt{p}\blat_i)} \stackrel{W_k}{\Longrightarrow} \mu.
$
%weakly to a probability distribution $\mu=\mu(\Lambda, S)$ on $\R_{>0}\times\R$ with bounded $4$th moment and assume that as $p\rightarrow\infty$, the $4$th moment of the empirical distribution converges to the $4$th moment of $\mu$, i.e.,
%$$
%\frac{1}{p}\sum_{i=1}^p\left(\sqrt{(\sqrt{p}\blat_i)^{2}+\lai^2}\right)^4 \rP \E_{(\Lambda,S)\sim\mu}\left[\left(\sqrt{\Lambda^2+S^2}\right)^4\right]<\infty.
%$$
\end{assumption}
}

Under this assumption, it can be verified that, the exact same proof strategy we used for magnitude-based pruning would apply to Hessian-based pruning by replacing $\bts$ with $\blat$. The reason is that the Hessian pruning risk is a $\text{PL}(4)$ function of $\h,\blat,\bSi$ (e.g.~can be shown in a similar fashion to Lemma \ref{lem:hPL}). Observe that Assumption \ref{ass:lambda} is a reasonable assumption given the naturalness of the saliency score. If we only wish to use the earlier Assumption \ref{ass:mu} rather than Assumption \ref{ass:lambda}, one can obtain the equivalent result by modifying \ref{ass:mu} to enforce a slightly higher order bounded moment and convergence condition. Finally, one needs to address the perturbation due to the finite sample estimation of the covariance. Note that, even if the empirical covariance doesn't converge to the population, its diagonal weakly converges to the population (as we assumed the population is diagonal). The (asymptotically vanishing) deviation due to the finite sample affects can be addressed in an identical fashion to the deviation analysis of $\tau_n,u_n$ at \eqref{eq:ivstep1} and \eqref{eq:betav}. While these arguments are reasonably straightforward and our Hessian pruning formula accurately predicts the empirical performance, the fully formal proof of the Hessian-based pruning is rather lengthy to write and does not provide additional insights. 
%Thus the associated formal theoretical statement is deferred to a future version.

\cmt{
if one assumes Assumption \ref{ass:mu} with the 
The overall function $f(\blat_i,$ becomes
\[
h(x,y,z)=f(g(y,z,x),y,z)=y(x-g(z))^2
\]}
%\cts{Using the assumption of the corollary that $\beta$}
%\end{proof}

%%%%%%%%%%%%%%%%%%%%%%%%%%%%%%%%%%%%
%%
%%
%%%%%%%%%%%%%%%%%%%%%%%%%%%%%%%%%%%%
\section{Useful results about pseudo-Lipschitz functions and CGMT}\label{SM useful fact}
For $k\geq 1$ we say a function $f:\R^m\rightarrow\R$ is pseudo-Lipschitz of order $k$ and denote it by $f\in \rm{PL}(k)$ if there exists a constant $L>0$ such that, for all $\x,\y\in\R^m$:
\begin{align}
|f(\x)-f(\y)|\leq L\left(1+\|\x\|_2^{k-1}+\|\y\|_2^{k-1}\right)\|\x-\y\|_2.\label{PL func}
\end{align}
In particular, when $f\in\rm{PL}(k)$, the following properties hold; see \cite{bayati2011dynamics}:
\begin{enumerate}
\item There exists a constant $L'$ such that for all $\x\in\R^n$: $|f(\x)|\leq L'(1+\|\x\|_2^k).$

\item $f$ is locally Lipschitz, that is for any $M>0$, there exists a constant $L_{M,m}<\infty$ such that for all $x,y\in[-M,M]^m$,
$
|f(\x)-f(\y)| \leq L_{M,m}\|\x-\y\|_2.
$
Further, $L_{M,m}\leq c(1+(M\sqrt{m})^{k-1})$ for some costant $c$.
\end{enumerate}

Using the above properties, we prove the following two technical lemmas used in the proof of Theorem \ref{thm:master_W2}

\begin{lemma}\label{lem:fL}
Let $g:\R\rightarrow\R$ be a Lipschitz function. Consider the function $f:\R^3\rightarrow\R$ defined as follows:
$$
f(\x) = x_3(x_2-g(x_1))^2.
$$
Then, $f\in\rm{PL}(3)$. If additionally, $f:\R^2\times \Zc\rightarrow\R$ for a bounded set $\Zc\subset\R$, then $f\in\Plt\subset\rm{PL}(3)$. Specifically, setting $\Zc=[\Sigma_{\min},\Sigma_{\max}]$ (as per Assumption \ref{ass:mu}), we find that $\Fl\subset\Plt$, where $\Fl$ is defined in \eqref{eq:fdef}.
\end{lemma}

\begin{proof}
We first prove that $f\in\rm{PL}(3)$.

Let $h:\R^2\rightarrow\R$ defined as $h(\ub)=(\ub_2-g(\ub_1))^2$. The function $(\ub_1,\ub_2)\mapsto\ub_2-g(\ub_1)$ is clearly Lipschitz. Thus, $h\in\rm{PL}(2)$, i.e., 
\begin{align}
|h(\ub)-h(\vb)| \leq C(1+\|\ub\|_2+\|\vb\|_2)\|\ub-\vb\|_2\quad\text{and}\quad |h(\vb)|\leq C'(1+\|\vb\|_2^2).\label{eq:h_pl}
\end{align} 
Therefore, letting $\x=(\ub,x_3)\in\R^3$ and $\y=(\vb,y_3)\in\R^3$, we have that 
\begin{align}
|f(\x)-f(\y)| &= |x_3h(\ub) - y_3h(\vb)| \leq |x_3||h(\ub)-h(\vb)| + |h(\vb)| |x_3-y_3|\nn\\
%&\leq |x_3|(1+\|\ub\|_2+\|\vb\|_2)\|\ub-\vb\|_2 + |h(\vb)| |x_3-y_3| \nn\\
&\leq C|x_3|(1+\|\ub\|_2+\|\vb\|_2)\|\ub-\vb\|_2 + C'(1+\|\vb\|_2^2)|x_3-y_3| \nn\\
&\leq C(|x_3|^2+(1+\|\ub\|_2+\|\vb\|_2)^2)\|\ub-\vb\|_2 + C'(1+\|\vb\|_2^2)|x_3-y_3| \nn\\
&\leq C(1+\|\x\|_2^2+\|\y\|_2^2)\|\ub-\vb\|_2 + C'(1+\|\x\|_2^2+\|\y\|_2^2)|x_3-y_3| \nn\\
&\leq C(1+\|\x\|_2^2+\|\y\|_2^2)\|\x-\y\|_2.
\end{align}
In the second line, we used \eqref{eq:h_pl}. In the third line, we used $2xy\leq x^2+y^2$. In the fourth line, we used Cauchy-Schwarz inequality. $C,C'>0$ are absolute constants that may change from line to line.

\fy{Now, we shall prove that $f\in\Plt$. To accomplish this, we simply need to show that for all $z\in \Zc$, $f(\cdot,\cdot,z)$ is $\rm{PL}(2)$ (for some uniform PL constant). This can be shown as follows. Let $C=\sup_{z\in\Zc}|z|$. Then
\begin{align}
|f(x,y,z) - f(x',y',z)| &=|z(y-g(x))^2-z(y'-g(x'))^2|\\
%&\leq |z||(y-g(x))^2-(y'-g(x'))^2|\\
&\leq C|(y-g(x))^2-(y'-g(x'))^2|\\
&\leq C(1+\tn{\ub}+\tn{\vb})\tn{\ub-\vb},
 \end{align}
%\begin{align}
%|F_n(\btha_n(\g,\h),\betas,\Sigmab) - F_n(\bt,\betas,\Sigmab)| &=
% \frac{1}{p}\sum_{i=1}^p|\lai|\left| (\sqrt{p}\betas_i-g(\sqrt{p}\btha_{n,i}))^2-(\sqrt{p}\betas_i-g(\sqrt{p}\bt_{i}))^2 \right|\nn\\
% &\leq
%\Sigma_{\max} \frac{1}{p}\sum_{i=1}^p \left| (\sqrt{p}\betas_i-g(\sqrt{p}\btha_{n,i}))^2-(\sqrt{p}\betas_i-g(\sqrt{p}\bt_{i}))^2 \right|\nn\\
% &\leq
%{C} \Sigma_{\max} \frac{1}{p}\sum_{i=1}^p (1+ \|\sqrt{p}[\betas_i,\btha_{n,i}]\|_2 +  \|\sqrt{p}[\betas_i,\bt_{i}]\|_2) \sqrt{p}|\btha_{n,i}-\bt_{i}|\nn\\
% &\leq
%{C} \Sigma_{\max}  \Big(1+ \frac{1}{\sqrt{p}}\big(\sum_{i=1}^{p}\|\sqrt{p}[\betas_i,\btha_{n,i}]\|_2^2\big)^{1/2} + \frac{1}{\sqrt{p}}\big(\sum_{i=1}^p\|\sqrt{p}[\betas_i,\bt_{i}]\|_2^2\big)^{1/2}\Big) \|\btha_n-\bt\|_2\nn\\
% &\leq
%%C  \left(1+ \|\betas\|_2^2+ \|\btha_n\|_2^2 + \|\bt\|_2^2\right) \|\btha_n-\bt\|_2.
%C  \left(1+ \max\{\|\betas\|_2^2,\|\btha_n\|_2^2,\|\bt\|_2^2\}^{1/2} \right) \|\btha_n-\bt\|_2.\label{eq:fcase1}
% \end{align}
where $\ub=[x,y]$ and $\vb=[x',y']$. In the second line above, we used boundedness of $\Zc$. In the third line, we used the fact that the function $\psi(x,y) = (y-g(x))^2$ is $\rm{PL}(2)$ as it is a quadratic of a Lipschitz function.
}

This completes the proof of the lemma.
\end{proof}

{
\begin{lemma}[PL with Bounded Variables]\label{lem:PLbdd}%with bounded derivatives 
Let $f:\R^{d_1}\rightarrow \R$ be a $PL(k)$ function, $\Mc\subset \R^{d_2}$ be a compact set and $g$ be a continuously differentiable function over $\Mc$. Then $h(\x,\y)=f(\x)g(\y)$ is $PL(k+1)$ over $\R^{d_1}\times \Mc$.
\end{lemma}
\begin{proof} First observe that $g$ has continuous derivatives and is continuous over a compact set. Thus $g$ and its gradient are bounded and $g$ is Lipschitz over $\Mc$. Let $B=\sup_{\x\in \Mc}\max |g(x)|,\tn{\nabla g(\x)}$. To proceed, given pairs $(\x,\y)$ and $(\x',\y')$ over $\R\times \Mc$, we have that
\begin{align}
|h(\x,\y)-h(\x',y')|&\leq |h(\x,\y)-h(\x',\y)|+ |h(\x',\y)-h(\x',\y')|\\
&\leq |f(\x)-f(\x')||g(\y)|+ |f(\x')||g(\y)-g(\y')|\\
&\leq B|f(\x)-f(\x')|+ B\tn{\y-\y'}|f(\x')|\\
&\leq B(1+\tn{\x'}^{k-1}+\tn{\x}^{k-1})\tn{\x-\x'}+ B\tn{\y-\y'}(1+\tn{\x'}^k)\\
&\leq C (1+\tn{\z}^k+\tn{\z'}^{k})\tn{\z-\z'},
\end{align}
where $\z=[\x~\y]^T$ and $C$ an absolute constant. This shows the desired $\text{PL}(k+1)$ guarantee.
\end{proof}
The following lemma is in similar spirit to Lemma \ref{lem:PLbdd} and essentially follows from similar lines of arguments (i.e.~using Lipschitzness induced by boundedness).
}
\begin{lemma}\label{lem:hPL}
Let functions $f,g:\R^3\rightarrow\R$ such that $f\in\rm{PL}(k)$ and 
$$
g(x,y,z) := \frac{y^{-1/2}}{1+(\ksi_* y)^{-1}}\sqrt{\kappa}\tau_* z + (1-(1+\ksi_*y)^{-1})x.
$$
Here, $\ksi_*,\tau_*,\kappa$ are positive constants. 
%bounded, say $x_2\in[m,M]\subset(0,\infty)$.
Further define 
\begin{align}
h(x,y,z) := f\left(g(y,z,x),y,z\right),
\end{align}
and assume that $y$ take values on a fixed bounded compact set $\Mc{\subset\R^+}$.
Then, it holds that $h\in\rm{PL}(k+1)$.
% \fx{and $h(x,y,z)$ is continuously differentiable over $y\in \Mc$}
\end{lemma}

\begin{proof}
Since $f$ is $\rm{PL}(k)$, for some $L>0$, \eqref{PL func} holds. Fix $x,x'\in\R$, $\ab=[y,z]\in\R^{2}$ and $\ab'=[y',z']\in\R^{2}$. Let $\bb=[\g,\ab]=[\g,y,z]\in\R^3$ where $\g=g(y,z,x)\in\R$ and define accordingly $\bb'=[\g',\ab']$ and $\g'=g(y',z',x')$. We have that
\begin{align}
|h([x,\ab])-h([x',\ab'])|&= |f(\bb)-f(\bb')|\nn\\
&\leq L\left(1+\|\bb\|_2^{k-1}+\|\bb'\|_2^{k-1}\right)\|\bb-\bb'\|_2\nn\\
&\leq C\left(1+\|\ab\|_2^{k-1}+\|\ab'\|_2^{k-1}+|\g|^{k-1}+|\g'|^{k-1}\right)(\|\ab-\ab'\|_2+|\g-\g'|),\label{eq:hlip}
%~~~~~~~~&L\left(1+\|\g(\y,\z,\x)\|_2^{k-1}+\|g(\y',\z',\x')\|_2^{k-1}\right)\|g(\y,\z,\x)-g(\y',\z',\x')\|_2.
\end{align}
for some constant $C>0$. In the last inequality we have repeatedly used the inequality
$
\left(\sum_{i=1}^m \|\vb_i\|_2^2\right)^{\frac{d}{2}} \leq  C(m)\cdot\sum_{i=1}^m\|\vb_i\|_2^{d}.
$
Next, we need to bound the $\g$ term in terms of $(x,\ab)$. This is accomplished as follows
\begin{align}
|\g|^{k-1} &= \left|\frac{y^{-1/2}}{1+(\ksi y)^{-1}}\sqrt{\kappa}\tau_* z + (1-(1+\ksi_*y)^{-1})x\right|^{k-1}\nn\\
%\|\g\|_2^{k-1}&\leq C \frac{1}{p}\sum_{i=1}^p|g_i|^{k-1}\\
%&\lesssim \frac{1}{p}\sum_{i=1}^p\left|\frac{y_i^{-1/2}}{1+(\ksi y_i)^{-1}}\sqrt{\kappa}\tau_* z_i + (1-(1+\ksi_*y_i)^{-1})x_i\right|\\
&\leq C (|z|+|x|)^{k-1} \nn\\
&\leq C \left(|x|^{k-1}+|z|^{k-1}\right) \leq C (|x|^{k-1} + \|\ab\|_2^{k-1}).\label{eq:gb}
\end{align}
Here, the value of the constant $C>0$ may change from line to line. 
Secondly and similarly, we have the following perturbation bound on the $\g-\g'$. Recall that $y,y'\subset \Mc$ are bounded. Additionally, since $\Mc\subset\R^+$ and is compact, $\Mc$ is strictly bounded away from $0$. Let 
\[
g_1(y)=\sqrt{\kappa}\tau_*\frac{y^{-1/2}}{1+(\ksi_* y)^{-1}}\quad\text{and}\quad g_2(y)=1-(1+\ksi_*y)^{-1}.
\]
It can be seen that $g_1,g_2$ are continuously differentiable functions over $\Mc$. Thus $g_1,g_2$ are bounded and have bounded derivatives over $\Mc$. We will prove the following sequence of inequalities
%\CT{Not sure why this is red. Looks ok to me} 
% \so{Suppose the triples $(x_i,y_i,z_i)$ takes values in a fixed bounded compact set $\Mc$. We will prove the following sequence of inequalities} 
\begin{align}
%|\|\g\|_2-\|\g'\|_2|&\leq 
|{\g-\g'}|&= |g(y,z,x)-g(y',z',x')| \nn\\
&\leq \left|g_1(y) x - g_1(y') x'\right|
 + \left|g_2(y)z-g_2(y')z'\right|\nn\\
 &\leq
\left|g_1(y) x - g_1(y) x'\right| +\left|g_1(y) x'- g_1(y') x'\right|\nn\\
 &\qquad+ \left|g_2(y)z-g_2(y)z'\right| + \left|g_2(y)z'-g_2(y')z'\right| \nn\\
&\leq C_1|x-x'| + C_2|x'||y-y'| + C_3|z-z'| + C_4|z'||y-y'|\nn\\
&\leq C(1+|x'| + |z'|)(|x-x'| + |z-z'| + |y-y'|)\\
&\leq C\sqrt{3}(1+|x'| + |z'|)\|[\ab,x]-[\ab',x']\|_2
.\label{eq:gd}
%C^2_{\Mc} \sum_{i=1}^p(|x_i-x'_i|^2+|y_i-y'_i|^2+|z_i-z'_i|^2)\label{sec line eq}\\
%&\leq C^2_{\Mc} (\tn{\x-\x'}^2+\tn{\y-\y'}^2+\tn{\z-\z'}^2)\\
%&\lesssim \tn{\ab-\ab'}^2.
\end{align}
In the fourth inequality above, we used the fact that $|g_i(y)|,|g_i'(y)|$ are bounded. In the last line, we used Cauchy-Scwhartz.

Substituting \eqref{eq:gb} and \eqref{eq:gd} in \eqref{eq:hlip} gives:
\begin{align}
|h(x,y,z) - h(x',y',z')|  &\leq C\left(1+\|\ab\|_2^{k-1}+\|\ab'\|_2^{k-1}+|x|^{k-1}+|x'|^{k-1}\right)(1+|x'| + |z'|)\|[\ab,x]-[\ab',x']\|_2\nn \\
&\leq C\left(1+\|[\ab,x]\|_2^{k}+\|[\ab',x']\|_2^{k}\right)\|[\ab,x]-[\ab',x']\|_2.
\end{align}
Thus, $h\in\rm{PL}(k+1)$, as desired.
%In what follows, we prove the second line \eqref{sec line eq} i.e.~the fact that for any triples $a=(x,y,z),a'=(x',y',z')$ (with $a,a'\in\R^3$), we have that 
%\[
%|g(a)-g(a')|\leq C_{\Mc}\tn{a-a'}.
%\]
%Set $C_\nabla=\sup_{a\in \Mc}\tn{\nabla g(a)}$. By definition of gradient, the inequality above holds with $C_\Mc=C_\nabla$. Thus, all that remains is proving that $C_\nabla$ is upper bounded by a constant. \so{However, this automatically holds because from the definition of $g(\dots)$ function, it is clear that $C_\nabla$ is defined everywhere and continuous thus it has a finite maximum over a compact set.}
\end{proof}

%\section{Supporting Results on CGMT}\label{SM cgmt res}
The following theorem replaces the compactness constraint with  closedness in the CGMT and is borrowed from \cite{li2020exploring}. For related statements see also \cite[App.~A]{deng2019model}.
\begin{theorem} [CGMT with Closedness Constrains]\label{thm closed} Let $\psi$ be a convex function obeying $\lim_{\tn{\w}\rightarrow\infty}\psi(\w)=\infty$. Given a closed set $\Sc$, define
\begin{align}
\Phi_\la(\X)&=\min_{\w\in\Sc}\la\tn{\X\w}+\psi(\w)\\
\phi_\la(\g,\h)&=\min_{\w\in\Sc}\la(\tn{\w}\tn{\g}-\h^T\w)_++\psi(\w),
\end{align}
and
\begin{align}
&\Phi_\infty(\X)=\min_{\w\in\Sc,\X\w=0}\psi(\w)\\
&\phi_\infty(\g,\h)=\min_{\w\in\Sc,\tn{\w}\tn{\g}\leq \h^T\w}\psi(\w).
\end{align}
For all $\la\in[0,\infty)\cup\{\infty\}$, we have that
\begin{itemize}
\item $\Pro(\Phi_\la(\X)<t)\leq2\Pro(\phi_\la(\X)\leq t)$.
\item If $\Sc$ is additionally convex, we additionally have that $\Pro(\Phi_\la(\X)>t)\leq2\Pro(\phi_\la(\X)\geq t)$. Combining with the first statement, this implies that for any $\mu,t>0$
\[
\Pro(|\Phi_\la(\X)-\mu|>t)\leq2\Pro(|\phi_\la(\X)-\mu|\geq t)
\]
\end{itemize}
\end{theorem}

\section{Underparameterized analysis}\label{SM overdet}

This section provides our results for the asymptotic DC in the underparameterized regime. This results establish direct counterparts of the overparameterized results Definition \ref{def:Xi} and Theorem \ref{thm:master_W2}. However, underparameterized DC is substantially less involved compared to overparameterized. A key reason is that underparameterized least-squares returns an unbiased estimate of the ground-truth parameter. Similar to Section \ref{sec proof thm 1}, for simplicity, we assume diagonal covariance however results can be translated to arbitrary covariance via eigen-rotation trick (e.g.~recall Def.~\ref{aux_def}). Throughout, we solve the following problem% as observed by \cite{}
\begin{align}
\bth=\X^\dagger\y=\arg\min_{\bt} \tn{\y-\X\bt}^2\label{bth up}
\end{align}
where $\y=\X\bts+\sigma\z$ and $\X=\Xb\sqrt{\bSi}$. Now, set $\wo=\sqrt{\bSi}(\bt-\bts)$ as previously. We can rewrite
\begin{align}
\bth=\X^\dagger\y=\bts+\bSi^{-1/2}\w^\st\quad\text{where}\quad \w^\st=\arg\min_{\wo} \tn{\sigma\z-\Xb\wo}^2.\label{wst up}
\end{align}
We will prove the following DC for the underparameterized problem with $n<p$ and $p/n=\kappa<1$.
\begin{definition}[Asymptotic DC -- Underparameterized regime]\label{def:Xi_under}
Let random variables $(B,\Lambda)\sim \mu$ (where $\mu$ is defined in Assumption \ref{ass:mu}) and fix $\kappa<1$. Let $H\sim\Nn(0,1)$ and define the random variable
\begin{align}
X_{\kappa,\sigma^2}(B,H) :=  B + \sigma\frac{\Lambda^{-1/2}H}{\sqrt{\kappa^{-1}-1}}, \label{eq:X}%\sqrt{\kappa}\frac{\sqrt{\gamma}\,\Lambda^{-1/2}}{1+(\xi\Lambda)^{-1}} 
\end{align}
and let $\Pi_{\kappa,\sigma^2}$ be its distribution.
\end{definition}

We are now ready to state our main theoretical result.
\cmt{
\begin{theorem}[Asymptotic DC -- Underparameterized LGP]\label{thm:master_W2_under} %Fix $\kappa=p/n>1$.
Let Assumption \ref{ass:linear} hold with $\kappa<1$ and  further let Assumptions \ref{ass:inv} and \ref{ass:mu} hold. Consider $\hat{\bt}$ as in \eqref{bth up} and $\hat\Pi_n(\y,\X,\betas,\Sigmab):=\frac{1}{p}\sum_{i=1}^{p}\delta_{\sqrt{p}\hat{\bt}_i,\sqrt{p}\betas_i,\Sigmab_{i,i}}$, the joint empirical distribution of $(\sqrt{p}\hat\betab,\sqrt{p}\betas,\Sigmab)$. Recall the definition of the measure $\Pi_{\kappa,\sigma^2}$ in Def.~\ref{def:Xi_under}. Then, $\hat\Pi_n(\y,\X,\betas,\Sigmab)$ converges in Wasserstein-k distance to $\Pi_{\kappa,\sigma^2}\otimes\mu$.
Specifically, for any function $f:\R^3\rightarrow\R$, $f\in\rm{PL}(k)$ with $k\geq 3$, it holds that
\begin{align}\label{eq:thm_up}
\hspace{-0.1in}p^{-1} \sum_{i=1}^{p} f(\sqrt{p}\hat\betab_i,\sqrt{p}\betas_i,\Sigmab_{i,i}) \rP \E\left[f(X_{\kappa,\sigma^2},B,\Lambda) \right],
\end{align}
where the expectation is over $(\Lambda,B,H)\sim\mu\otimes\Nn(0,1)$. Specifically, the asymptotic test risk is given by $\frac{\sigma^2}{1-\kappa}$.
\end{theorem}}
%any of the following two: (a) $f\in\rm{PL}(2)$, or, (b) $f=f_\Lc$
\begin{theorem}[Asymptotic DC -- Underparameterized LGP]\label{thm:master_W2_under} %Fix $\kappa=p/n>1$.
Fix $\kappa<1$. Let Assumptions \ref{ass:inv} and \ref{ass:mu} hold. Consider $\hat{\bt}$ as in \eqref{bth up} and $\hat\Pi_n(\y,\X,\betas,\Sigmab):=\frac{1}{p}\sum_{i=1}^{p}\delta_{\sqrt{p}\hat{\bt}_i,\sqrt{p}\betas_i,\Sigmab_{i,i}}$, the joint empirical distribution of $(\sqrt{p}\hat\betab,\sqrt{p}\betas,\Sigmab)$. Recall the definition of the measure $\Pi_{\kappa,\sigma^2}$ in Def.~\ref{def:Xi_under}. {Let $f:\R^3\rightarrow\R$ be a function in $\Plt$ where $\Plt$ is defined in \eqref{eq:pdef}.} We have that
\begin{align}\label{eq:thm_up}
\frac{1}{p} \sum_{i=1}^{p} f(\sqrt{p}\hat\betab_i,\sqrt{p}\betas_i,\Sigmab_{i,i}) \rP \E_{(\Lambda,B,H)\sim\mu\otimes\Nn(0,1)}\left[f(X_{\kappa,\sigma^2},B,\Lambda) \right].
\end{align}
Specifically, the asymptotic test risk of $\bth$ is given by $\frac{\sigma^2}{1-\kappa}$.
\end{theorem}
\begin{proof} \so{To avoid repetition, we will not provide the full proof as the technical details of the proofs for over/under-parameterized overlap to a significant extent. Instead, we will provide the part of the proof that deviates from the overparameterized.} 

Since $\kappa<1$, the problem has a unique solution. Set $\wo=\sqrt{\bSi}(\bt-\bts)$. Define $\X_\z=[\Xb~\z]$ and $\wo_\sigma=[\wo~\sigma]$. This leads to the optimization problem
\[
\hat{\wo}=\arg\min_{\wo} \tn{\Xb\wo+\sigma \z}=\arg\min_{\wo} \tn{\X_\z\wo_\sigma}.
\]
Fix $\g\sim\Nn(0,\Iden_p),\h\sim\Nn(0,\Iden_n),g\sim\Nn(0,1)$. Applying CGMT leads to the following Auxiliary Optimization%and noticing optimal $\wo$ has form $\wo=w\bar{\g}$ where $\bar{\g}=\g/\tn{\g}$, we find
\begin{align}
\phi(\g,\h)&=\min_{\wo} \max_{\tn{\ab}\leq 1} \h^T\ab \tn{\wo_{\sigma}}-\tn{\ab}\g^T\wo+g\sigma\\
&=\min_{\wo}  \tn{\h} \tn{\wo_{\sigma}}-\g^T\wo+g\sigma.
\end{align}
Solving for optimal $\wo$ leads to the solution
\[
\wo^{\text{AO}}=\arg\min_{\wo}\tn{\h} \tn{\wo_{\sigma}}-\g^T\wo\implies \tn{\h}\frac{\wo^{\text{AO}}}{\sqrt{\tn{\wo^{\text{AO}}}^2+\sigma^2}}-\g\implies \wo^{\text{AO}}=\frac{\sigma\g}{\sqrt{\tn{\h}^2-\tn{\g}^2}}.
\]
Observing $\tn{\wo^{\text{AO}}}^2+\sigma^2=\frac{\sigma^2\tn{\h}^2}{\tn{\h}^2-\tn{\g}^2}$ and plugging $\wo^{\text{AO}}$ in, we find
\[
\sigma^{-1}\phi(\g,\h)=\frac{\tn{\h}^2-\tn{\g}^2}{\sqrt{\tn{\h}^2-\tn{\g}^2}}+g=\sqrt{\tn{\h}^2-\tn{\g}^2}+g.
\]
Thus, in the asymptotic regime $\phi(\g,\h)$ converges to the objective% following scalarized problem asymptotically.
\[
\phi(\g,\h)\rP \bar{\phi}=\sigma\sqrt{n-p}.
\]
The remaining arguments are same as in Lemma \ref{lem:AO}. First, the problem is strongly convex with $\sigma^2_{\min}(\X)$, which satisfies $\sigma^2_{\min}(\X)/p\gtrsim 1$ wpa.~1. Thus, the solution $\w^\st$ of the primary problem \eqref{wst up} will not deviate from $\wo^{\text{AO}}$. Secondly, the empirical distribution of 
\[
\sqrt{p}\wo^\st=\frac{\sigma\sqrt{p}\g}{\sqrt{\tn{\h}^2-\tn{\g}^2}}\rP \frac{\sigma\sqrt{p}\g}{\sqrt{n-p}}= \frac{\sigma\g}{\sqrt{n/p-1}}=\frac{\sigma\g}{\sqrt{\kappa^{-1}-1}},
\]
converges to $\sigma H/\sqrt{\kappa^{-1}-1}$. By Assumption \ref{ass:mu}, the empirical distribution of $\sqrt{p}\bth=\sqrt{p}(\bt^\st+\bSi^{-1/2}\wo^\st)$ converges to $B+\sigma \Lambda^{-1/2}H/\sqrt{\kappa^{-1}-1}\sim \Pi_{\kappa,\sigma^2}$. Finally, again by Assumption \ref{ass:mu}, for any $f\in \Plt$, we obtain the advertised result \eqref{eq:thm_up}. The asymptotic test risk is given by
\[
\Lc(\bth)=\E[\tn{\g^T\sqrt{\bSi}(\bth-\bts)+\sigma z}^2]=\sigma^2+\sum_{i=1}^p (\bth_i-\bts)^2\bSi_{i,i}\rP\sigma^2+\frac{\sigma^2}{\kappa^{-1}-1}=\frac{\sigma^2}{1-\kappa}.
\]
\end{proof}

In the main body of the paper, we claim that the optimal $s$ features to use in the underparameterized regime is given by the features with the maximum saliency score. This is proven below.
\begin{lemma}[Optimal $s$ features to use]\label{lem best s} Fix a sequence of sets $\Delta_p\subset[p]$ of size $s$ such that $\sum_{i\in \Delta_p} {\bts_i}^2\bSi_{i,i}\rP B(\Delta)$. Set $\kappa=s/n$. Under same assumptions as in Thm~\ref{thm:master_W2_under}, the asymptotic test risk of $\bth(\Delta)$ is given by
\[
\Lc(\bth(\Delta))\rP \frac{B-B(\Delta)+\sigma^2}{1-\kappa}.
\]
Thus, the optimal feature set $\Delta$ chooses the indices with maximum Saliency Score \eqref{saliency eq} which maximizes $B(\Delta)$.
\end{lemma}
\begin{proof} The key idea is the fact that we can treat the missing features as uncorrelated noise. First, due to diagonal covariance, observe that, over the feature set $\Delta$, the optimal population model (i.e.~infinite sample) is $\bts_{\Delta}$. Thus, the $s$ feature problem minimized by $\bth(\Delta)$ can be written as the dataset model
\[
y=\x_{\Delta}^T\bts_{\Delta}+\sigma_{\Delta}^2,
\]
where the noise level is given by
\[
\E[(y-\x_{\Delta}\bts_{\Delta})^2]=\sigma^2+\E[(\x_{\bar{\Delta}}\bts_{\bar{\Delta}})^2]=\sigma^2+\sum_{i\not\in\Delta}\bSi_{i,i}{\bts_{i}}^2.
\]
The latter quantity converges to $B-B(\Delta)$ wpa.~1. Thus, applying Theorem \ref{thm:master_W2_under}, $\bth(\Delta)$ achieves the advertised asymptotic risk. 
%$\bth(\Delta)$ is the solution of a regression problem with $s$ features and noise level
\end{proof}

\section{Proof of Lemma \ref{lem rank one}}\label{SM lem proof}
 
\begin{lemma} [Lemma \ref{lem rank one} restated]Suppose $\Sc$ is drawn from an LGP$(\sigma,\bSi,\bt_\st)$ as in Def.~\ref{def LGP} where $\text{rank}(\bSi)=1$ with $\bSi=\bla\bla^T$ for $\bla\in\R^p$. Define $\zeta=\Tb_s(\bla)^2/\tn{\bla}^2$. For magnitude and Hessian pruning ($P\in\{M,H\}$) and the associated retraining, we have the following excess risks with respect to $\bt^\st$%population minima
\begin{align}%\underbrace{\frac{\zeta^2\sigma^2}{p-2}}_{\text{Variance Term}}
&\E_{\Sc}[\Lc(\bth_s^P)]-\Lc(\bt^\st)={\frac{\zeta^2\sigma^2}{n-2}}+\underbrace{(1-\zeta)^2(\bla^T\bt^\st)^2}_{\text{Error due to bias}}\\
&\E_{\Sc}[\Lc(\bth_s^{RT})]-\Lc(\bt^\st)={\sigma^2}/({n-2}).
\end{align}
\end{lemma}
\begin{proof}
\noindent \textbf{Retraining analysis:} We claim that for any feature set $\Delta$ with $\bla_{\Delta}\neq 0$, the test risk of $\bth(\Delta)$ is exactly identical. Secondly, pruning is guaranteed to pick a  nonzero support satisfying $\bla_{\Delta}\neq 0$ \footnote{This is because $\bth=c\bla$ for some scalar $c\neq 0$ as $\bth$ lies in the row space of $\X$. Then, Hessian/Magnitude-pruning would pick a nonzero support of $\bth$ which corresponds to the nonzero support of $\bla$.}. Thus, as described next, retraining always achieves a fixed risk. Set $c^\st=\bla^T\bt^\st$. By definition, each input example $\x_i$ has the form $\x_i=g_i\bla$ and $y_i=g_ic^\st+\sigma z_i$. Set $\g=[g_1~\dots~g_n]^T$ and $\bar{\g}=\g/\tn{\g}$. Thus, we have $\X=\g\bla^T$ and $\y=\g\bla^T\bt^\st+\sigma\z$. Decompose $\z=\bar{\z}+\bar{\g}^T\z\bar{\g}$ where $\bar{\z}$ is orthogonal to $\g$. When solving the regression of $\Delta$, we have that 
\[
\X_\Delta=\g\blad^T,~\y=c^\st\g+\sigma(\bar{\z}+\bar{\g}^T\z\bar{\g})
\]
The least-squares solution has the form $\bth=\bth(\Delta)=\hat{c}\blad/\tn{\blad}^2$ where
\begin{align}
&\hat{c}=\arg\min_{c}\tn{(c^\st-c)\g+\sigma\z}\implies\hat{c}=c^\st+\sigma\gamma\label{beta}.
\end{align}
where $\gamma=\frac{\bar{\g}^T\z}{\tn{\g}}$. Observe that $\sqrt{p}\gamma$ has Student's t-distribution with $p$ degrees of freedom. Set $h$, $\epsilon\distas \Nn(0,1)$. Now, observe that a fresh test sample with $y=\x^T\bt^\st+\sigma \epsilon$ with $\x=g\bla$, the test error obeys
\begin{align}
\Lc(\bth(\Delta))&=\E(y-\x_{\Delta}^T\bth(\Delta))^2\\
&=\mathbb{E}[((c^\st-\hat{c})g+\sigma\epsilon)^2]\\
&=(\gamma^2+1)\sigma^2
\end{align}
Now, observe that the minimum risk is obviously $\Lc(\bt^\st)=\sigma^2$. Thus, the excess retraining risk becomes
\[
\Lc(\bth(\Delta))-\Lc(\bt^\st)=\gamma^2\sigma^2.
\]
regardless of choice of $\Delta$. Finally, averaging this risk over $\Sc$ returns $\E[\sigma^2\gamma^2]=\sigma^2/(n-2)$. Thus retraining risk has the fixed excess risk same as the one advertised in Lemma \ref{lem rank one}.

\noindent \textbf{Pruning analysis:} For pruning setting $\Delta=[p]$ above, we have that $\bth=\hat{c}\bla/\tn{\bla}^2$. This means that, for both Magnitude and Hessian pruning\footnote{They yield the same result since diagonal covariance is proportional to $\bla$ in magnitude.}, pruned vector takes the form $\bth_s=\hat{c}\Tb_s(\bla)/\tn{\bla}^2$. Using the fact that $\bla^T\Tb_s(\bla)=\tn{\Tb_s(\bla)}^2=\zeta\tn{\bla}^2$, we find
\begin{align}
\Lc(\bth_s)-\Lc(\bt^\st)&=\mathbb{E}[(c^\st- \hat{c}\frac{\bla^T\Tb_s(\bla)}{\tn{\bla}^2})g+\sigma\epsilon)^2]-\sigma^2\\
&=\mathbb{E}[(c^\st- \hat{c}\zeta)g+\sigma\epsilon)^2]-\sigma^2\\
&=\E[(((1-\zeta)c^\st-\zeta\sigma\gamma)g+\sigma\epsilon)^2]-\sigma^2\\
&=((1-\zeta)c^\st-\zeta\sigma\gamma)^2.
\end{align}
Finally, using zero-mean $\gamma$, we find
\[
\E_{\Sc}[\Lc(\bth_s)]-\Lc(\bt^\st)=\E_{\Sc}[((1-\zeta)c^\st-\zeta\sigma\gamma)^2]=(1-\zeta)^2{c^\st}^2+\frac{\zeta^2\sigma^2}{n-2},
\]
which concludes the proof after observing ${c^\st}^2=(\bla^T\bt^\st)^2$. Here, we call $(1-\zeta)^2{c^\st}^2$ ``the error due to bias''. The reason is that the predictable signal in the data is the noiseless component $\x^T\bt^\st$. Pruning leads to an error in this predictable component by resulting in a biased estimate of the label (when conditioned on the random variable $g$ which controls the signal).
\end{proof}

\end{document}